\documentclass[preprint]{elsarticle}

\usepackage{lineno,hyperref}

\usepackage{epsfig}
\usepackage{epstopdf}
\usepackage{verbatim}
\usepackage{amsmath,amssymb} 
\usepackage{graphicx}
\usepackage[lofdepth,lotdepth]{subfig}
\usepackage{colortbl}
\usepackage{multirow}
\usepackage{rotating}
\usepackage{setspace}
\usepackage{caption}
\usepackage{bchart}
\usepackage{algpseudocode}
\usepackage{algorithm}
\usepackage{adjustbox}
\usepackage{tikz}
\usepackage{soul}
\usetikzlibrary{arrows,matrix,positioning}
\modulolinenumbers[5]

\journal{Image and Vision Computing}

\newcommand {\bH}{\mathbf{H}}               
\newcommand {\bR}{\mathbf{R}}               
\newcommand {\bK}{\mathbf{K}}               
\newcommand {\bI}{\mathbf{I}}               
\newcommand {\bS}{\mathbf{S}}               
\newcommand {\bKt}{\tilde{\mathbf{K}}}      
\newcommand {\bX}{\mathbf{X}}              
\newcommand {\ba}{\mathbf{a}}               
\newcommand {\bb}{\mathbf{b}}              
\newcommand {\bp}{\mathbf{p}}               
\newcommand {\bx}{\mathbf{x}}              
\newcommand {\by}{\mathbf{y}}              
\newcommand {\bv}{\mathbf{v}}              
\newcommand {\bu}{\mathbf{u}}              
\newcommand {\R}{\mathbb {R}}              

\begin{document}

\begin{frontmatter}

\title{Enhancing Feature Tracking With Gyro Regularization}


\author{Bryan Poling}
\ead{poli0048@math.umn.edu}

\author{Gilad Lerman\corref{mycorrespondingauthor}}
\cortext[mycorrespondingauthor]{Corresponding author}
\ead{lerman@math.umn.edu}

\address{School of Mathematics\\127 Vincent Hall, 206 Church St. SE, Minneapolis, MN 55455}

\begin{abstract}
We present a deeply integrated method of exploiting low-cost gyroscopes to improve general purpose feature tracking. Most previous methods use gyroscopes to initialize and bound the search for features. In contrast, we use them to regularize the tracking energy function so that they can directly assist in the tracking of ambiguous and poor-quality features. We demonstrate that our simple technique offers significant improvements in performance over conventional template-based tracking methods, and is in fact competitive with more complex and computationally expensive state-of-the-art trackers, but at a fraction of the computational cost. Additionally, we show that the practice of initializing template-based feature trackers like KLT (Kanade-Lucas-Tomasi) using gyro-predicted optical flow offers no advantage over using a careful optical-only initialization method, suggesting that some deeper level of integration, like the method we propose, is needed in order to realize a genuine improvement in tracking performance from these inertial sensors.
\end{abstract}

\begin{keyword}
feature tracking\sep optical flow\sep inertial sensors\sep gyroscopes\\
\textit{Supp. Webpage:} \url{http://www-users.math.umn.edu/~lerman/GyroTracking/}
\end{keyword}
\end{frontmatter}


\section{Introduction}
Feature tracking is the task of determining and maintaining the location of one or more visually interesting points as they move about in motion video. This task is crucial in computer vision, where it is frequently used as a first step in solutions to important problems like simultaneous localization and mapping (SLAM) and structure from motion (SFM). A common solution is to characterize each feature using a \emph{template} image, which is a small image centered on the feature, extracted from a recent frame. This template is updated periodically, and the feature's location in each new frame is determined by searching through the new imagery for the region that is most common to the template. The celebrated Kanade-Lucas-Tomasi (KLT) feature tracker~\cite{lucas1981iterative,Tomasi91detectionand,shi_tomasi94,baker2004lucas} achieves this, for instance, using Gauss-Newton optimization to find the location in a new image that minimizes the mean-squared difference between the image and the template.

For a given camera, the location of a feature in the image plane is a function of the location of the corresponding 3D world point relative to the coordinate frame of the camera. If we imagine a ``world frame'' fixed to the environment, then the motion of a feature in the image plane (which we will refer to as a feature's \emph{flow}) can be decomposed into two quantities: motion of the corresponding 3D point relative to the world frame, and movement of the camera frame relative to the world frame. This decomposition is useful because motion of world points is often much slower when measured in the coordinate frame of their environment than when measured in the coordinate frame of the camera. If the motion of the camera relative to the environment can be measured independently, then the component of a feature's flow due to camera egomotion can be predicted a-priori, leaving only the smaller component due to motion through the environment to be determined. This is especially helpful in hand-held camera applications, where camera rotation can completely dominate other sources of flow.

Independently measuring camera motion can be quite challenging. Position and orientation can often be determined by measuring external signals like GNSS (Global Navigation Satellite Signals). Such external signals are not always detectable or reliable however, and exploiting them can require additional undesirable hardware (like GNSS antennas). Additionally, for the purpose of predicting a feature's flow we actually only need to measure relative motion (from one frame to the next), instead of knowing our pose relative to an absolute reference frame. In this setting, inertial sensors (which measure accelerations due to specific forces and rotation rates) are a desirable alternative because they can offer excellent relative motion measurements over short time intervals without relying on external signals. Additionally, these sensors have become very low cost in recent years and have made their way into all kinds of consumer electronics (like smart phones and tablet computers). It is now the case that many devices one might use to collect video already happen to have inertial sensors built into them. This makes them a natural choice for integration with feature trackers.

Inertial sensors can be used to estimate relative changes in camera orientation and/or position between two frames and predict the component of the optical flow field due to camera egomotion. The simplest way to exploit this information is to use the predicted flow field to initialize and bound the search for features in the new frame. This has been found to be a successful strategy by several authors (see \S~\ref{sec:Review of Previous Work}). While this strategy can reduce the number of candidate locations for a feature in a new frame of video, it still relies entirely on the imagery for selecting the best candidate. We propose an additional level of integration where we regularize the tracking energy function to gently penalize deviation from our prior estimate of flow. Thus, in addition to reducing the number of candidate locations for a feature, our method can help differentiate between them when the imagery is not distinctive enough to reveal the location on its own.

The rest of the paper is organized as follows. We 
review previous work in \S\ref{sec:Review of Previous Work} and detail the contributions of this work in \S\ref{sec:Original Contributions of This Work}. Template-based feature tracking is reviewed in \S\ref{sec:A Review of Template-Based Feature Tracking}. In \S\ref{sec:Using Gyroscopes to Predict Optical Flow} we summarize how a prior estimate of optical flow can be computed using a gyroscope. In \S\ref{sec:Exploiting A Prior Estimate Of Flow} we introduce our method of exploiting this prior flow estimate for feature tracking. In \S\ref{sec:Experiment Results} we present results which demonstrate that our technique offers significant improvements in performance over conventional template-based tracking methods, and is in fact competitive with much more computationally expensive state-of-the-art trackers, but at a fraction of the cost. \S\ref{sec:Future Work} explores possible areas for future research. Finally, \S\ref{sec:Conclusions} concludes this work.

\subsection{Review of Previous Work}
\label{sec:Review of Previous Work}
Gyroscopes have been used in conjunction with optical systems for multiple purposes in recent years. Perhaps the most ubiquitous application is image stabilization. This broadly refers to two different problems. The first is to reduce blur in individual images during long-exposure acquisitions (e.g., low-light photography). This can be done by detecting camera rotation during the acquisition period and driving actuators connected to the imager to mechanically compensate for the motion \cite{sachs2006image} (this is called optical image stabilization, or OIS), or it can be done by using the detected camera motion to deblur the image in post-processing (as in \cite{joshi2010image}). The second problem is to reduce the shakiness of video by measuring rotation during video acquisition and translating frames in the sequence to compensate for short, rapid motions, resulting in video that only reflects the smoother camera motion \cite{karpenko2011digital}.

Gyro-optical integration has also been used to improve feature tracking for computer vision applications. You et al.~\cite{you1999hybrid} used gyroscopes to predict feature motion in the image plane of a camera and restrict the search space for feature tracking. Feature trajectories were then used in turn to yield driftless attitude estimates for use in augmented reality systems. Hwangbo et al.~\cite{hwangbo2011gyro} used gyroscopes to estimate relative changes in orientation to predict feature motion for initializing a KLT feature tracker (and to pre-warp templates), in order to provide more robust feature tracking for vision applications such as 3D reconstruction.

Perhaps the most studied application of integrated optical and inertial sensors has been in aiding inertial navigation systems when conventional sources of navigation information (like GNSS) are unavailable or untrustworthy. Pachter and Mutlu~\cite{5531219} proposed tracking and geo-localizing unknown terrain features and using them as landmarks for self-localization when conventional aides are unavailable. Mourikis et al.~\cite{mourikis2009vision} and Roumeliotis et al.~\cite{roumeliotis2002augmenting} exploit feature trackers to generate measurements to aid inertial navigation systems for space travel and space vehicle descent. Jones and Soatto~\cite{jones2011visual} use trajectories from a KLT feature tracker with an inertial navigation system in a modern framework for robust SLAM. While these methods employ loose integration between optical and inertial systems, other authors have proposed tighter integration (where the inertial system plays an important role in feature tracking). Hol et al.~\cite{Hol_Robust_real_time_tracking} used measured changes in both position and orientation to predict the motion of features in the image plane and bound the search space for those features. They located features by minimizing the sum-of-absolute-differences between imagery in the search region and feature template images. They used trajectories in turn to provide corrections to the navigation system via a Kalman filter. Veth and Raquet~\cite{veth2006fusion,veth2007fusion} implemented a similar optical-inertial system, while matching SIFT descriptors to locate features in the search region. Gray~\cite{gray2009deeply} extended this work to achieve deeper integration between the inertial and optical systems by predicting perturbations in feature descriptors that result from changes in system pose and directly compensating feature descriptors to achieve better matching performance. Predicting feature motion due to sensor translation is considerably more challenging than predicting motion due to rotation, as it requires estimating the range from the camera center to each tracked feature. To deeply integrate feature tracking with a full inertial navigation system one must employ some method of estimating feature range (\cite{Hol_Robust_real_time_tracking} assumes a 3D scene model is known, \cite{veth2006fusion} and \cite{veth2007fusion} use a binocular camera, and \cite{gray2009deeply} presents results using binocular systems and monocular systems with range estimated online from motion). An alternative is to neglect the effect of camera translation on feature motion (when rotation is dominant). Diel et al.~\cite{diel2005epipolar} used gyroscopes to measure relative changes in camera orientation and pre-warped imagery to compensate, thereby making feature tracking simpler. They used tracked features to derive epipolar constraints and applied them to an inertial navigation system through a Kalman filter.

\subsection{Original Contributions of This Work}
\label{sec:Original Contributions of This Work}
Our main contribution is a novel method for deeply integrating inertial sensors with template-based feature tracking. We directly modify the tracking energy function to exploit inertial measurements as a prior estimate of feature position. This modification aids the tracker in localizing features in directions where the imagery is ambiguous (directions where the unmodified energy function is relatively flat). Unlike our proposed method, previous works in integrating inertial sensors with feature trackers generally achieve lower levels of sensor integration. Many methods are very loosely integrated, where feature tracking is nearly independent of the inertial system~\cite{mourikis2009vision,roumeliotis2002augmenting} and realizes no benefit from these other sensors. Other methods exploit the inertial sensors (and sometimes additional sources of navigation information) to initialize and bound the search for features in the image plane~\cite{gray2009deeply,Hol_Robust_real_time_tracking,hwangbo2011gyro,veth2007fusion,veth2006fusion,you1999hybrid}. Some techniques also use this information to pre-warp feature template images~\cite{hwangbo2011gyro}, or to correct feature descriptors to better match new imagery~\cite{gray2009deeply}. However, to our knowledge, ours is the first method to directly regularize the tracking energy function using gyro-predicted optical flow.

Our proposed solution competes with state-of-the-art feature trackers in performance, but has only a fraction of the computational cost of some competing methods. In addition, we show that using gyro-predicted optical flow to merely initialize a feature tracker (like KLT) offers no advantage over a careful optical-only initialization method (see \eqref{eqn:Average flow initialization}). This suggests that in some of the advances reported by other authors the gyroscopes were not really needed to achieve the gains from better initialization. A deeper level of integration, like the one proposed here, may be necessary to realize a genuine tracking performance improvement from these inertial sensors.

\section{A Review of Template-Based Feature Tracking}
\label{sec:A Review of Template-Based Feature Tracking}
The goal of feature tracking is to determine the location of a small point or feature as it moves about in motion video. This work focuses on template-based feature tracking, where a feature is characterized by a small (usually square) \emph{template image} that describes the expected appearance of the feature. This fundamentally distinguishes feature tracking from \emph{object tracking}, where the target is potentially larger and can change significantly in appearance between frames, requiring more sophisticated techniques for target characterization. In template-based feature tracking, the goal is usually to locate a feature by minimizing the \emph{single-feature energy function}:
\begin{equation}
\label{eqn:Single feature energy function}
c(\bx) = {1 \over n^2} \sum_{\bu \in \Omega} \psi \left( T(\bu) - I(\bu + \bx) \right),
\end{equation}
where:
\begin{align*}
T =& \text{ Template Image characterizing the feature}\\
n =& \text{ Width and height of template image}\\
I =& \text{ Frame of video in which we are trying to locate the feature}\\
\psi =& \text{ Loss function. Typically } \psi(\by)=|\by| \text{ or } \psi(\by)=|\by|^2\\
\bx =& \text{ The candidate location of the feature in the new frame}\\
\Omega =& \text{ The set of locations where } T \text{ is defined: }
\text{\{} (i,j) \text{ where } i,j \in \{1,2,...,n\} \text{\}}
\end{align*}

Intuitively, we are overlying our template image $T$ on the video frame $I$ in a location governed by the input $\bx$. Then, for each pixel in the template, we compute the difference in intensity between $T$ and $I$ and take the square or absolute value of the result. For each pixel in the template, this produces a measure of dissimilarity between the template and the image which is $0$ when they are equal and positive otherwise. Our energy is the average of these values over all pixels in the template. Thus, when we minimize \eqref{eqn:Single feature energy function}, we are effectively sliding our template around over the video frame, looking for the location where the image looks most similar to the template.

Generally, there is a maximum distance which a feature is expected to move between two consecutive frames. Thus, the energy \eqref{eqn:Single feature energy function} is minimized in a local neighborhood of its previous position. The minimization can be performed through exhaustive search, using $1^\text{st}$-order methods like gradient descent, or using higher-order schemes like Gauss-Newton minimization (as in the KLT tracker). Additionally, most feature trackers are implemented in a coarse-to-fine framework, where tracking is done in stages: first on lower-resolution imagery and then subsequently refined on higher-resolution imagery~\cite{bouguet2001pyramidal}. This is to increase the likelihood that in any given stage the initial estimate of a feature's position is within the region of convergence of the true minimum of \eqref{eqn:Single feature energy function}.

As an alternative to tracking each feature independently, some modern feature trackers have as their state variable the joint positions of a collection of many (or all) features in the scene. Hence, they iteratively refine the position estimates of each feature simultaneously, as opposed to one after another. This allows them to impose constraints on how features can move relative to one other. An example of this is \cite{PolingLermanSzlam2014}, where features are encouraged, through a penalty term, to maintain trajectories over a short temporal window which lie in a low-dimensional subspace. We refer to such tracking methods as \emph{multi-feature trackers} or in short \emph{multi-trackers}.

In the multi-feature tracking framework the state variable encodes the joint positions of a collection of many (or all) features in the scene. The main energy function for multi-feature tracking, which we call the \emph{multi-feature energy function}, is formed by simply combining all of the single-feature energy functions of a collection of $F$ tracked features:
\begin{equation}
\label{eqn:Multi-feature energy function}
{1 \over n^2} \sum_{f=1}^F \sum_{\bu \in \Omega} \psi \left( T_f(\bu) - I(\bu + \bS_f \bx) \right), \text{ where } \bS_f=
\begin{tikzpicture}[baseline={([yshift=-2ex]current bounding box.center)},scale=0.76, every node/.style={scale=0.76}]
\matrix [matrix of math nodes,left delimiter=(,right delimiter=)] (m)
{
	0 & ... & 0 & 1 & 0 & 0 & ... & 0 \\
	0 & ... & 0 & 0 & 1 & 0 & ... & 0 \\
};
\draw[color=red] (m-1-4.north west) -- (m-1-5.north east) -- (m-2-5.south east) -- (m-2-4.south west) -- (m-1-4.north west);
\draw[color=red,double,implies-](m-1-4.north) -- +(0,0.3) -- +(-0.3,0.3);
\node (lab1) [color=red,above=1.5ex of m-1-2.north] {col 2f-1};
\draw[color=red,double,implies-](m-1-5.north) -- +(0,0.3) -- +(0.3,0.3);
\node (lab1) [color=red,above=1.5ex of m-1-7.north] {col 2f};
\end{tikzpicture}.
\end{equation}
The state variable $\bx$ is a $2F$-element vector, where elements $(2f-1)$ and $2f$ form the position of feature $f$. The matrix $\bS_f$ extracts these coordinates from $\bx$. The inner sum in \eqref{eqn:Multi-feature energy function} is nothing more than the standard single-feature energy function \eqref{eqn:Single feature energy function} for feature $f$. The outer sum adds up these energy functions for each feature being tracked. Notice though that each single-feature energy function depends on a different pair of coordinates in the state vector, $\bx$. Thus, minimizing \eqref{eqn:Multi-feature energy function} is equivalent to minimizing each single-feature energy function separately. It only becomes different if we impose constraints or add additional terms that introduce interactions between the locations of different features. As with single-feature trackers, there are many ways this energy can be minimized, and it is often implemented in a multi-resolution, coarse-to-fine framework.

It is important to understand the limitations inherent in both single-feature and multi-feature tracking. In single-feature tracking, unless one pulls in additional sources of information, it is not generally possible to track arbitrary features. For instance, if you are trying to track a feature in a completely textureless, uniform portion of an image, then it can be readily verified that the energy function \eqref{eqn:Single feature energy function} becomes completely flat. When this happens, there is no unique minimizer and the flow of the feature is not observable in the imagery. A less extreme example would be tracking a feature along an edge in an image. In this case, sliding the template transverse to the edge results in an energy increase, but sliding the template along the edge does not change the energy. In this situation it is theoretically possible to partially resolve the flow of the feature (i.e. in the direction orthogonal to the edge). However, template-based trackers are not designed to exploit such partial information and will instead try to fully resolve the location of the feature. They will select what they see as the best minimum, even when there are several, nearly identical candidates. The result is that such features can wander and jump around along the edges to which they belong. Without additional sources of data or additional assumptions about how features interact, the multi-feature tracking framework suffers from the same problem because it essentially minimizes the single-feature energy functions for a set of features.

One straightforward way to address this issue is to simply not attempt to track features which are less than ideal. For instance, \cite{Tomasi91detectionand} defines a criterion for a template that ensures that the corresponding feature is distinctive enough to make tracking possible and suggests discarding features that don't meet the criterion. Many other criteria have been proposed in the literature, but all effectively try to select corner-like features and have the shared goal of screening out features that are likely to cause problems in tracking. This is a reasonable approach to the problem but it is not universally appropriate. For instance, when using trajectories for the purpose of 3D reconstruction, one wants long feature trajectories that survive long enough to allow individual features to be viewed from many different poses. Replacing features when they become difficult to track has the contrary effect of building a larger set of shorter-lived feature trajectories. Navigation applications also tend to prefer longer feature trajectories. Additionally, there are phenomena that can negatively impact many or all features in a scene together, such as poor lighting conditions. The ability to track through even short-duration events like this can be quite valuable in these applications.

When trying to track a collection of less-than-ideal features the flow of each feature may not be independently observable in the imagery. Thus, some additional assumptions or sources of information must be brought in to make this possible. In \cite{PolingLermanSzlam2014} a low rank constraint is used as a way of sharing information between features to make low-quality features more trackable. In this work, we bring in an outside estimate of the flow of each feature, in particular, the (often dominant) component due to camera rotation. We rely on this to help locate features in directions where the imagery is not distinctive enough for a features location to be discerned from the imagery alone. Before we discuss how we use this outside estimate, we develop the equations used to produce the estimate using gyroscopes rigidly mounted to our camera.

\section{Using Gyroscopes to Predict Optical Flow}
\label{sec:Using Gyroscopes to Predict Optical Flow}
In our application we estimate the component of optical flow that is due to changes in camera orientation between frames to use as our prior estimate of flow in feature tracking. In many situations this is the dominant source of flow simply because cameras are often free to rotate much faster than they can move through their environments (or relative to other visible objects). For the purpose of illustration, consider a common cellular phone camera with $50^\circ$ diagonal angle of view. In hand-held applications it is not uncommon for the camera to achieve rotation rates (in part due to unintentional camera ``shaking'') of $350^\circ$/s. If images are being collected at $30$ fps, this camera rotation induces motion in the image plane of up to $23\%$ of the image diameter between consecutive frames\footnote{The magnitude of the flow in pixels depends on the sensor resolution. If the sensor has a resolution of $600 \text{x} 800$ ($1000$ pixels across the diagonal), this would correspond to a maximum displacement of $230$ pixels.}. For a non-rotating camera viewing an object $50$ feet away, in order to generate a comparable flow in the image plane the object would need to be moving at least $199$ mph ($320$ km/h). In many situations objects are not moving nearly this fast, and camera rotation is therefore the dominant source of flow.

A typical camera can be modeled using the perspective model~\cite{Hartley2000}. With this model, there is a \emph{calibration matrix} $\bK$ associated with the camera, depending only on the internal characteristics of the image sensor and optics, such that the image of a 3D point with coordinates $\bX$ in the coordinate frame of the camera is $\bx = \bK \left[ \bI, \mathbf{0} \right] \bX$. Both $\bX$ and $\bx$ are expressed in homogeneous coordinates. The matrix $\bK$ has the following form:
\begin{equation}
\bK = \left[
\begin{array}{c c c}
f_x & 0 & a_0\\
0 & f_y & b_0\\
0 & 0 & 1
\end{array}
\right],
\end{equation}
where $f_x$, $f_y$, $a_0$, and $b_0$ are constants associated with the imaging sensor and optics. This model is relatively general since most camera non-idealities (e.g., radial distortion) can be compensated for, when necessary, explicitly in pre-processing.

Next, we must identify what information we are able to extract from the gyroscopes. Ideally, they would allow us to measure the rotation of the camera between arbitrary instants in time. Unfortunately, however, gyro measurements are taken in the coordinate frame of the gyro sensor package, which is not generally aligned with the camera frame. We will assume that there is some fixed rotation matrix that takes vectors in the gyro sensors coordinate frame to the cameras internal coordinate frame (this assumption requires that the gyro be physically mounted to the camera so that they experience the same rotations).

We will be considering the camera at two instants in time corresponding to acquisition times of two consecutive frames of video. We will refer to these instants as time $0$ and time $1$, respectively. We will let $\bX^0$ represent the position of a feature at time $0$ (in homogeneous coordinates), in a coordinate frame which we will refer to as the \emph{gyro frame}. This frame has its origin at the camera center and has its axes parallel to those of the coordinate frame of the gyro sensor package. Similarly, we will let $\bX^1$ represent the position of the feature at time $1$ in the gyro frame. Let $\bR_\text{Gyro}^\text{Cam}$ be the fixed rotation matrix that takes vectors in the gyro frame to the cameras internal coordinate frame (these coordinate frames share the same origin in space so there is no translational component to this transformation). Then, the images of a feature in the focal plane at times $0$ and $1$ are given by:
\begin{equation}
\bx^0 = \bK \left[ \bI, \mathbf{0} \right] \left[
\begin{array}{c c}
\bR_\text{Gyro}^\text{Cam} & \mathbf{0} \\
\mathbf{0} & 1\\
\end{array}
\right] \bX^0, \text{\;\; and \;\;}
\bx^1 = \bK \left[ \bI, \mathbf{0} \right] \left[
\begin{array}{c c}
\bR_\text{Gyro}^\text{Cam} & \mathbf{0} \\
\mathbf{0} & 1\\
\end{array}
\right] \bX^1.
\end{equation}
We can rewrite these equations as:
\begin{equation}
\bx^0 = \bK \bR_\text{Gyro}^\text{Cam} \left[ \bI, \mathbf{0} \right] \bX^0,
\text{\;\;\; and \;\;\;}
\bx^1 = \bK \bR_\text{Gyro}^\text{Cam} \left[ \bI, \mathbf{0} \right] \bX^1.
\end{equation}
If we define $\tilde{\bK} = \bK \bR_\text{Gyro}^\text{Cam}$, we get:
\begin{equation}
\label{eqn:Projections}
\bx^0 = \tilde{\bK} \left[ \bI, \mathbf{0} \right] \bX^0,
\text{\;\;\;and\;\;\;}
\bx^1 = \tilde{\bK} \left[ \bI, \mathbf{0} \right] \bX^1.
\end{equation}

Next, we will collect measurements from our $3$-axis gyroscope between time $0$ and time $1$. What comes out of a gyro are measurements of the rotation rates of the gyro frame about its X, Y, and Z axes relative to inertial space, coordinatized in the gyro frame. We will denote these rates $r_x$, $r_y$, and $r_z$, respectively. From these measurements we want to compute the $3\text{x}3$ matrix describing the rotation of the gyro frame between times $0$ and $1$, which we will call $\bR$. This is a well-studied problem, and we use a common quaternion representation for rotations to help solve it. It can be shown (e.g., \cite{Quaternions_and_Rotations_in_3-Space_Vernon_Chi}) that the quaternion representing the orientation of the gyro frame obeys the ODE:
$\dot{Q} = \left( {1 \over 2} \right) Q P,$
where $P = 0 + r_x \mathbf{i} + r_y \mathbf{j} + r_z \mathbf{k}$. We initialize our quaternion $Q$ at time $0$ as $Q = 1 + 0 \mathbf{i} + 0 \mathbf{j} + 0 \mathbf{k}$, which represents the trivial rotation (by angle $0$), and we integrate this ODE numerically from time $0$ to time $1$ and convert the final quaternion to a rotation matrix, $\bR$, as described in \cite{Quaternions_and_Rotations_in_3-Space_Vernon_Chi}.

Since $\bX^0$ and $\bX^1$ are defined relative to the gyro frame, they are related by:
\begin{equation}
\bX^1 = \left[
\begin{array}{c c}
\bR & \mathbf{0} \\
\mathbf{0} & 1\\
\end{array}
\right] \bX^0.
\end{equation}
Combining this with \eqref{eqn:Projections}, we can write:
\begin{equation}
\label{eqn:Relation between x^1 and X^0}
\bx^1 = \tilde{\bK} \left[ \bI, \mathbf{0} \right] \bX^1 = \tilde{\bK} \left[ \bI, \mathbf{0} \right] \left[
\begin{array}{c c}
\bR & \mathbf{0} \\
\mathbf{0} & 1\\
\end{array}
\right] \bX^0 = \tilde{\bK} \left[ \bR, \mathbf{0} \right] \bX^0 = \tilde{\bK} \bR \left[ \bI, \mathbf{0} \right] \bX^0.
\end{equation}

Next, observe that $\tilde{\bK}$ is invertible because it is the product of a rotation matrix and an invertible camera calibration matrix. We will insert $\tilde{\bK}^{-1}\tilde{\bK}=\bI$ in the right spot on the right hand side of \eqref{eqn:Relation between x^1 and X^0} to get:
\begin{equation}
\bx^1 = \left( \tilde{\bK} \bR \tilde{\bK}^{-1} \right) \tilde{\bK} \left[ \bI, \mathbf{0} \right] \bX^0 = \tilde{\bK} \bR\tilde{\bK}^{-1} \bx^0.
\end{equation}
Hence, the images of the feature at times $0$ and $1$ in the image plane are related by:
\begin{equation}
\label{eqn:Homogrophy}
\bx^1 = \bH \bx^0, \text{\; where \;} \bH = \tilde{\bK} \bR \tilde{\bK}^{-1}.
\end{equation}

\begin{figure}[t]
\captionsetup{margin=10pt,labelfont=bf}
\centering
\subfloat[Rotation mostly about cameras horizontal axis]{
	\includegraphics[width=0.49\linewidth, clip=true, trim=0mm 0mm 0mm 0mm]{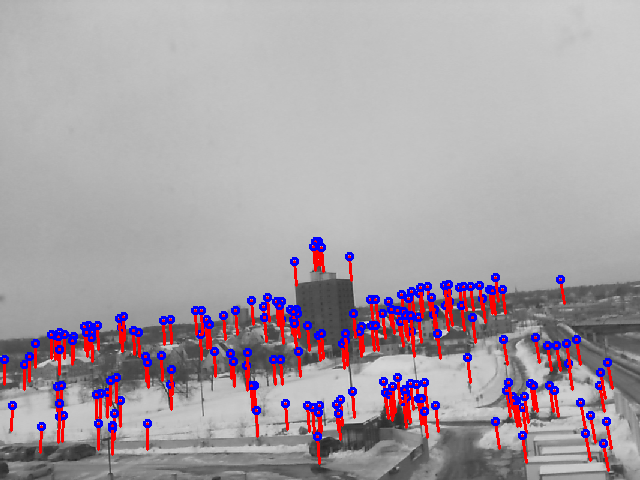}
}
\subfloat[Rotation mostly about cameras optical axis]{
	\includegraphics[width=0.49\linewidth, clip=true, trim=0mm 0mm 0mm 0mm]{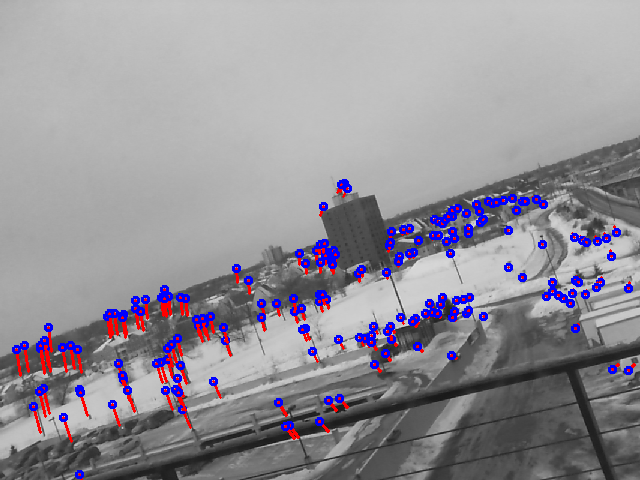}
}
\caption{Examples of optical flow prediction using a 3-axis gyroscope. Circles (blue) indicate feature locations in the previous frame; lines (red) connect them to the predicted locations of the same points in the current frame.}
\label{fig:Flow Prediction}
\end{figure}

Thus, if we are able to estimate the rotation of the gyro frame between two consecutive frames, $\bR$, and the constant $3 \text{x} 3$ matrix $\tilde{\bK}$ is known, then we can compute the matrix $\bH$ (which is a homography between the two frames) and predict where each point observed in the first frame will appear in the second. Figure \ref{fig:Flow Prediction} illustrates the output of this process. It is important to observe that the range to a feature is immaterial in this development. Range only effects the component of flow due to a points translational motion relative to the camera from time $0$ to time $1$, which is assumed to be negligible in this development.

The constant matrix $\tilde{\bK}$ can be determined by collecting a video sequence of a static scene where the camera undergoes rotations about all $3$ axes (but only negligible translations). Between each pair of consecutive frames the rotation $\bR$ is computed from the gyroscope and $\bH$ is estimated via optical image registration. Each consecutive pair of frames yields a set of $9$ equations from \eqref{eqn:Homogrophy}, where the unknowns are the elements of $\tilde{\bK}$. The system can be solved in a least-squares sense to yield $\tilde{\bK}$.

It should also be mentioned that using gyroscopes to predict optical flow requires some amount of calibration and integration between the gyros and the camera. For instance, the data from the gyros must be synchronized with the camera data. Also, gyros have slowly varying biases that must be compensated for. See \ref{Appendix: Data Collection Hardware} for a discussion of these details.

\section{Exploiting a Prior Estimate of Flow}
\label{sec:Exploiting A Prior Estimate Of Flow}
The idea behind our proposed method is to regularize the registration energy function by adding a term that penalizes feature positions that deviate from our prior estimate of flow (in our case derived from a gyroscope). The term we add will be small enough to not interfere with the well-behaved energy functions of strong features, but it will be significant enough to dominate the energy functions of weak features in ambiguous directions. This allows us to extract all of the information that is present in the imagery\footnote{For instance, we may be able to determine the horizontal position of a feature on a vertical edge very precisely from the imagery, but we may be unable to determine the same features vertical position along the edge. In this case there is still valuable information in the imagery that we do not want to ignore.}, but rely on our prior estimate of position when localizing in ambiguous directions. That is, we want the ability to ``ride'' our gyro measurements through poor imagery, until a feature can be re-acquired, but our implementation should only effect weak features and only in directions where the feature cannot be localized based purely on the imagery.

\begin{figure}[t]
\captionsetup{margin=10pt,labelfont=bf}
\centering
\subfloat[Corner-like feature template]{
	\includegraphics[width=0.23\linewidth, clip=true, trim=0mm 0mm 0mm 0mm]{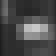}
}
\subfloat[Energy surface for (a)]{
	\includegraphics[width=0.23\linewidth, clip=true, trim=0mm 0mm 0mm 0mm]{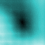}
}
\subfloat[Edge-like feature template]{
	\includegraphics[width=0.23\linewidth, clip=true, trim=0mm 0mm 0mm 0mm]{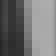}
}
\subfloat[Energy surface for (c)]{
	\includegraphics[width=0.23\linewidth, clip=true, trim=0mm 0mm 0mm 0mm]{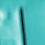}
}
\caption{Examples of a corner-like and edge-like feature template images from a real-world video sequence, along with their corresponding energies (see \eqref{eqn:Single feature energy function}). The energy for the edge-like feature does not have a unique minimum. The energy images are colorized to help distinguish them from the template images. Black is low energy, blue is medium energy, and white is high energy.}
\label{fig:corner vs edge energy surfaces}
\end{figure}

When tracking corner-like features, the single-feature energy surface~\eqref{eqn:Single feature energy function} is generally well-behaved in a small neighborhood of the global minimum. If we start minimization with an initial guess that is sufficiently close to the minimum we generally have no problems converging to it. However, when tracking features that are not corner-like, the energy surface is often not so well-behaved. In fact there may not even be a unique global minimum. In this situation, it is not possible to completely identify the correct location of a feature using template registration alone (this has nothing to do with the optimization scheme employed; it is a problem with the energy function itself). In Figure \ref{fig:corner vs edge energy surfaces} we give sample template images and energy functions for corner-like and edge-like features. Notice that there is a line of global minimums for the energy function corresponding to the edge-like feature (we will call this a line of ambiguity). This is because sliding the template image along the edge does not result in a better or worse match between the image and the template. Of course, because the energy function is derived from real-world data there will be a single global minimum somewhere on the line of ambiguity, but the location of the minimizer on this line will be unstable and unpredictable. This is why edge-like features tend to wander (and sometime jump) along their lines of ambiguity during tracking. It is important to note that simply initializing (or even bounding) the search of such a feature using the prior estimate of flow (as in \cite{hwangbo2011gyro}) will have little effect here since the issue is the instability of the minimizer of the energy function, not the inadequacy of a given optimization scheme at finding it.

In this work, we combine information about the flow of a feature from two very different sources. On one hand, we have imagery, which promises to yield an extremely accurate estimate of flow, provided the imagery is distinctive enough to reliably align with the feature's template. On the other hand, we have a flow estimate based on measurements of camera rotation. This estimate is more crude because it does not reflect flow due to translational motion of the feature relative to the camera, but it is not effected by ambiguity in the imagery. In summary, we have one source of information (the imagery) that can accurately capture all sources of flow (both due to camera rotation and translational motion of the feature), but which is susceptible to ambiguity in the imagery. Then we have another source of information (gyroscopes) that can only observe the often-dominant portion of flow, but which is not susceptible to ambiguity in the imagery. These properties make the two sources complimentary for the purposes of tracking.

When bringing together information from complimentary sources such as these, there is a temptation to produce estimates of the flow from both sources of information and combine them in a filter, exploiting known error characteristics of both estimates. This approach is impractical in this situation, however, because the error of the gyro-derived flow estimate is difficult, if not impossible to accurately characterize because it is blind to translational motion of features. Depending on the subject matter of the video, it may not even be reasonable to assume that the gyro-derived flow estimate is unbiased (imagine video of traffic, where most features are moving in one direction, in no part due to camera motion). What can surely be said is this: When the imagery is distinctive enough to localize a feature in a given direction, that estimate should be preferred over any other estimate. When the imagery is not distinctive enough to localize a feature in some direction, then it makes sense to fall back on the gyro-based estimate of flow. This is precisely what is accomplished by our proposed method. By weakly regularizing the image-based registration energy function using the gyro-derived estimate of flow, we use the imagery as our primary instrument for localizing a feature. However, when the imagery is not distinct enough to localize a feature in some direction, our gyro-based estimate takes over to fill in the missing information.

\subsection{Single-Feature Tracker Implementation}
\label{sec:Single-Feature Tracker Implementation}
\begin{figure}[h]
\centering
\includegraphics[width=.95\linewidth, clip=true, trim=0mm 0mm 0mm 0mm]{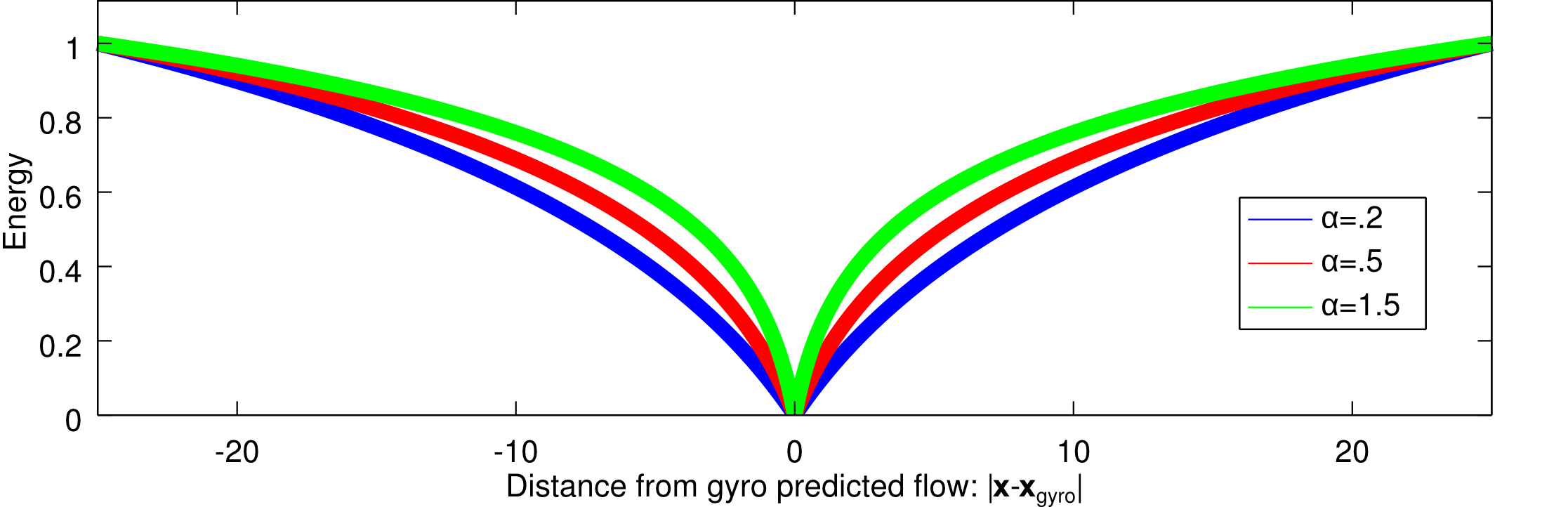}
\caption{The penalty \eqref{eqn:Regularization Function} with $x_\text{max}=25$ pixels, $\lambda=1$ and $\alpha= 0.2, 0.5, 1.5$.
It levels off not far from the gyro prediction and thus allows a strong registration energy function to dominate, even if the global min is far from the gyro prediction.}
\label{fig:Loss function}
\end{figure}

In \S\ref{sec:Using Gyroscopes to Predict Optical Flow} we covered how we can take a features location in one frame and predict its location in the next using gyro measurements. We will let $\bx_\text{gyro}$ denote this predicted position for a feature (2D position, in units of pixels). To achieve our goal in a single-feature tracking framework, we will add a term to \eqref{eqn:Single feature energy function} to penalize locations that differ from this prior prediction. The regularization term we add should be small enough to be completely dominated by a strong registration energy function. This must still be the case if the minimum of the registration energy function is relatively far from the prior estimate because there may be valid reasons for a significant discrepancy between true flow and predicted flow. In particular, since our prior is estimated using a gyroscope and only accounts for camera rotation, features on objects that are moving relative to the environment can have a significant component of flow that will not be reflected in the prior estimate. Thus, our regularization term must not grow too quickly as we move away from the prior estimate. It is also desirable, of course, for the regularization function to not contain local extrema. The penalty function we use is:
\begin{equation}
\label{eqn:Regularization Function}
P(\bx) = \lambda \; \ln(\alpha |\bx - \bx_\text{gyro}| + 1)/\ln(\alpha x_\text{max} + 1),
\end{equation}
where $x_\text{max}$ is a constant (in units of pixels) reflecting the greatest expected deviation from the gyro prediction, the constant $\alpha$ controls the ``pointiness'' of the curve, and the constant $\lambda$ controls the overall strength of the penalty (see Figure \ref{fig:Loss function}). In this entire work, we keep $\alpha$ fixed at $0.5$ and $x_\text{max}$ fixed at $25$ pixels. We discuss in \S\ref{sec:Experiment Results} how the parameter $\lambda$ can be learned.

After incorporating the above regularization term, we see that to track a feature when we have a prior estimate of flow we need to minimize the energy function:
\begin{equation}
\label{eqn:Regularized Single feature energy function}
c_{\text{single}}(\bx) = {1 \over n^2} \sum_{\bu \in \Omega} \psi \left( T(\bu) - I(\bu + \bx) \right) + \lambda \; {{\ln(\alpha |\bx - \bx_\text{gyro}| + 1)} \over {\ln(\alpha x_\text{max} + 1)}}.
\end{equation}

In this work, we use the absolute-value loss function for $\psi$. Other variables, such as $n$, $\Omega$, $T$, and $I$ have the same meanings as in \S\ref{sec:A Review of Template-Based Feature Tracking}. In the above energy function, the sum on the left measures the dissimilarity between the template image for the feature and the equally-sized region of the video frame, centered about the point $\bx$. The term on the right measures the dissimilarity between the point $\bx$ and the gyro-derived predicted location for the feature. The value of $\lambda$ will be small, so as to put greater emphasis on the objective of matching the template with the energy. We caution against trying to impose a strict statistical interpretation on the combination of terms in \eqref{eqn:Regularized Single feature energy function}. The sum on the left and the term on the right are not estimates of the location of the feature. Instead, they are energy functions which are, under suitable assumptions, minimized by the location of the feature. While our term on the right is derived from an estimate of the features location, little can be said about the error characteristics of that estimate. In general, as a result of translational motion of world points relative to the camera, the estimate may not even be unbiased. We suggest viewing \eqref{eqn:Regularized Single feature energy function} as a modification of the classical feature tracking energy function, where we add some additional shape to the energy function using a prior estimate of flow. This addition is very slight, and is only consequential when the classical energy function has an ambiguous minimizer (i.e. there are multiple candidate locations in the frame that appear to match the template).

As was the case when tracking without a prior flow estimate, this energy function can be minimized in a number of ways. In our implementation we use gradient descent with fast line search in a coarse-to-fine framework. This is much faster than exhaustive search, while still providing reliable, consistent behavior, even with poorly conditioned energy functions. This is an iterative algorithm where in each step, the gradient of the energy function, $\nabla E$, at the current location is computed and $- \nabla E$ is taken as the search direction. Then, the energy is approximately minimized on a line segment starting at the current location and continuing in the search direction. These steps are repeated for a minimum number of steps and until a maximum number of steps is reached or a stop condition is met. In this work we have two stop conditions: (a) The gradient of the energy has sufficiently small magnitude ($0.00001$), or (b) the gradient magnitude is not sufficiently smaller than in the previous iteration (greater than $0.9999$ times the magnitude from the previous iteration). This algorithm is detailed in Algorithm \ref{alg:Gradient Descent With Fast Line Search}. The gradient of the energy function is differentiated numerically using the centered derivative approximation (we perturb the position of the feature by $0.25$ pixels in each direction). We initialize the tracker on a given feature to the gyro-predicted location for that feature:
\begin{equation}
\label{eqn:Gyro Initialization}
\bx = \bx_{\text{gyro}}.
\end{equation}
The source code for this tracker will be available on our supplemental web page.

\begin{spacing}{1.45}
\begin{algorithm}[H]
\caption{Gradient Descent With Fast Line Search}\label{alg:Gradient Descent With Fast Line Search}
\begin{algorithmic}
\Require $E:\R^D \rightarrow \R$: Energy function. $\bp_0 \in \R^D$: Initial guess of the minimizer of $E$ (\eqref{eqn:Gyro Initialization} in our method, or \eqref{eqn:Average flow initialization} for optical-only methods). $\text{minSteps}$ (default = $40$ for lowest resolution level, $3$ for all other levels), $\text{maxSteps}$ (default = $40$), $\text{stepSize}$ (default = $2.0$ pixels), $\text{maxRefinements}$ (default = $10$).
\Ensure Minimizer of energy function (at least a local minimizer).
\State $\bp \gets \bp_0$.
\For {$n=1:\text{maxSteps}$}
	\State $\bv \gets - \nabla E(\bp)$\Comment(Compute search direction)
	\If {$n > \text{minSteps}$ and ($|\bv| \approx 0$ or $|\bv| > 0.9999 |\bv_\text{prev}|$)}
		\State Break from loop
	\EndIf
	\State $\bv_\text{prev} \gets \bv$\Comment(Save $\bv$ for future convergence tests)
	\State $\bv \gets \text{stepSize} \left( \bv/|\bv| \right)$\Comment(Compute step vector)
	\State $\bx_1 \gets \bp$,\;\; $a \gets E(\bx_1)$
	\State $\bx_2 \gets \bp + \bv$,\;\; $b \gets E(\bx_2)$
	\State $\bx_3 \gets \bp + 2 \bv$,\;\; $c \gets E(\bx_3)$
	\State $R \gets 1$\Comment(Initialize refinement counter)
	\While {$R < \text{maxRefinements}$}
		\If {$a > b > c$}\Comment(Step forward in search direction)
			\State $\bx_1 \gets \bx_2$, \;\; $a \gets b$
			\State $\bx_2 \gets \bx_3$, \;\; $b \gets c$
			\State $\bx_3 \gets \bx_3 + \bv$, \;\; $c \gets E(\bx_3)$
		\Else\Comment(Shorten step size)
			\State $\bv \gets \bv/2$			
			\State $\bx_3 \gets \bx_2$,\;\; $c \gets b$			
			\State $\bx_2 \gets \bx_1 + \bv$,\;\; $b \gets E(\bx_2)$
			\State $R \gets R+1$
		\EndIf
	\EndWhile
\EndFor
\State Return$(\bx_1)$
\end{algorithmic}
\end{algorithm}
\end{spacing}

\subsection{Multi-Feature Tracker Implementation}
In addition to the single-feature tracker of \S\ref{sec:Single-Feature Tracker Implementation}, our proposed method can also be used in a multi-feature tracking framework (see~\S\ref{sec:A Review of Template-Based Feature Tracking}). To incorporate our gyro prior, we add a collection of terms to \eqref{eqn:Multi-feature energy function} which penalize each feature's deviation from its predicted position. Thus, we minimize the following regularized multi-feature energy function:
\begin{align}
\label{eqn:Multi-feature energy function - regularized}
c_\text{multi}(\bx) = {1 \over n^2} &\sum_{f=1}^F \sum_{\bu \in \Omega} \psi \left( T_f(\bu) - I(\bu + \bS_f \bx) \right) +\notag\\
\lambda &\sum_{f=1}^F {{\ln(\alpha |\bS_f \bx - \bx_{f,\text{gyro}}| + 1)} \over {\ln(\alpha x_\text{max} + 1)}},
\end{align}
where $\bx_{f,\text{gyro}}$ denotes the prior estimate of position for feature $f$ (2D position, in units of pixels). As with the single-feature implementation, we use the absolute-value loss function for $\psi$. This energy function can be minimized by whatever means would be used to minimize \eqref{eqn:Multi-feature energy function}. We minimize this energy in a coarse-to-fine scheme using $4$ pyramid levels, and we use a slightly modified version of the gradient descent method described in Algorithm \ref{alg:Gradient Descent With Fast Line Search}. The only difference is that the search direction is modified to prevent strong features from completely controlling the search direction, causing weak features to be lost. The search direction computation is detailed in Algorithm \ref{alg:search direction computation for multi-tracker}, and is very similar to the method used in \cite{PolingLermanSzlam2014}.

\begin{algorithm}[h!]
\caption{Search direction for multi-feature tracker with gyro prior}\label{alg:search direction computation for multi-tracker}
\begin{algorithmic}
\Require $Dc_\text{multi}$: Gradient of energy function, $F$: number of features
\Ensure $\bv$: Final search direction
\State $\ba \leftarrow -Dc_\text{multi}$
\State $\ba \leftarrow \ba/|\ba|$
\For {$f=1:F$}
\State $\by_f \leftarrow [\ba(2f-1),\ba(2f)]^T$
\State $\bb_f \leftarrow \by_f/|\by_f|$
\EndFor
\State $\bb \leftarrow [\bb_1^T,\bb_2^T,...,\bb_F^T]^T$
\State $\bv \leftarrow 0.5\ba + 0.5\bb$
\end{algorithmic}
\end{algorithm}

Like the single-feature tracker, we compute the gradients of the image-template fit terms numerically, using the centered gradient approximation and sampling $0.25$ pixels in each direction. However, rather than grouping together regularization terms with image-template fit terms, we evaluate the gradients for these explicitly. We will denote the sum of all of the gyro-prior terms in \eqref{eqn:Multi-feature energy function - regularized} using $P$:
\begin{equation}
P = \lambda \sum_{f=1}^F {{\ln(\alpha |\bS_f \bx - \bx_{f,\text{gyro}}| + 1)} \over {\ln(\alpha x_\text{max} + 1)}}.
\end{equation}
Now, let $x_i$ denote the $i$'th entry of $\bx$, and let $x_{i,\text{gyro}}$ denote the gyro-derived prior estimate of $x_i$. The gradient of $P$ is given by:
\begin{equation}
{{dP} \over {dx_i}} = {{\lambda \alpha (x_i - x_{i,\text{gyro}})} \over {\ln(\alpha x_\text{max} + 1)(\alpha N_i + 1)N_i}},
\end{equation}
where:
\begin{align*}
N_i &= \sqrt{(x_i - x_{i,\text{gyro}})^2 + (x_{i+1} - x_{i+1,\text{gyro}})^2} \text{\;\; if $i$ is odd, and}\\
N_i &= \sqrt{(x_{i-1} - x_{i-1,\text{gyro}})^2 + (x_{i} - x_{i,\text{gyro}})^2} \text{\;\; if $i$ is even.}
\end{align*}

We also present results for our multi-feature tracker with gyro prior with an additional rank penalty from \cite{PolingLermanSzlam2014}. We use the rank penalty based on empirical dimension and the centered trackpoint matrix and we compute the gradient of the rank term using the method described in \cite{PolingLermanSzlam2014}. When minimizing the energy function with an additional rank penalty term, we use the same algorithm (including the search direction rule) as we use when there is no additional rank term. The source code for the multi-feature tracker will be available on our supplemental web page.

\section{Experiments}
\label{sec:Experiment Results}
\begin{figure}[H]
\captionsetup{margin=10pt,labelfont=bf}
\centering
\subfloat[Video 1]{
	\includegraphics[width=0.49\linewidth, clip=true, trim=5mm 20mm 5mm 20mm]{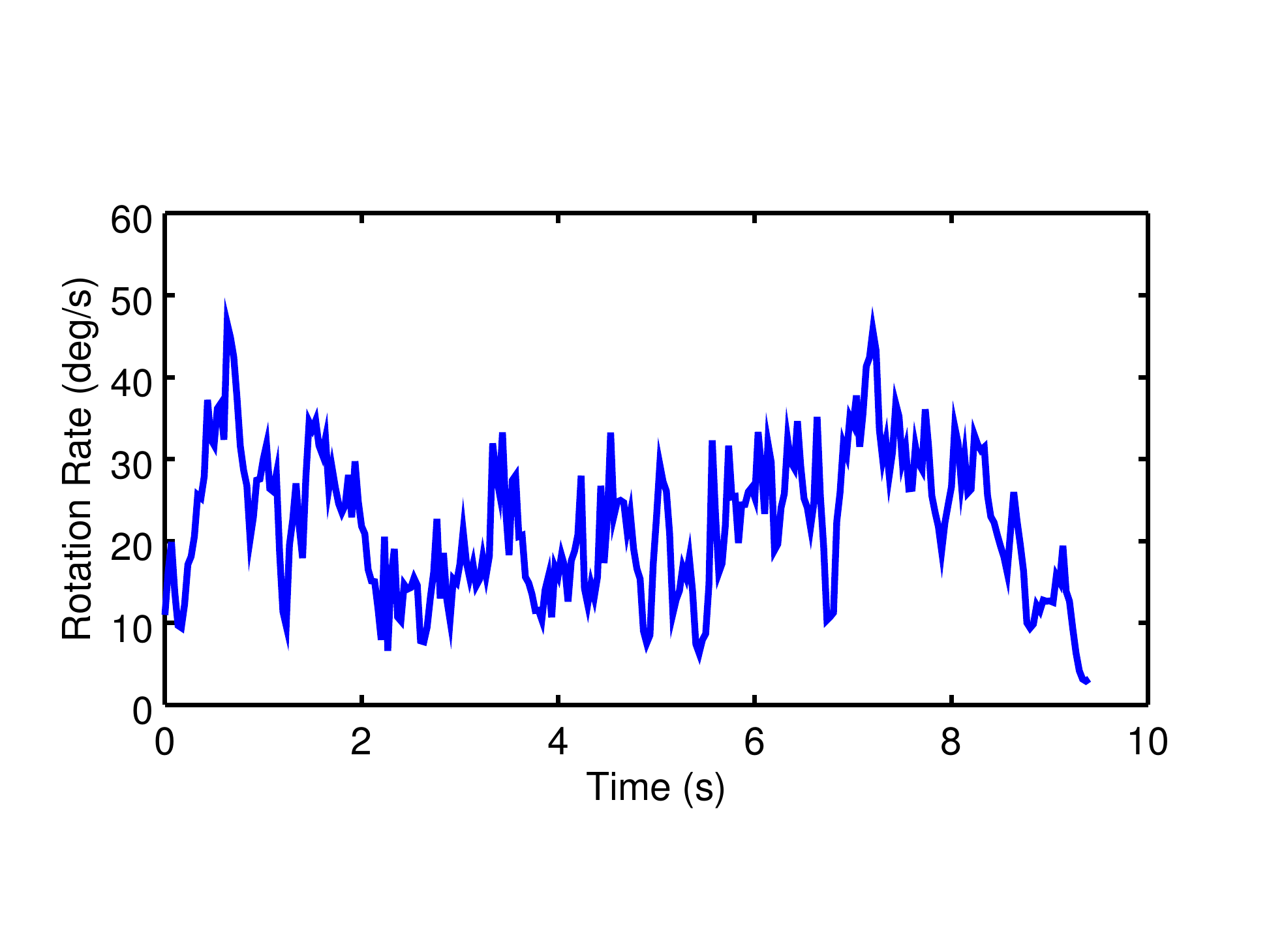}
}
\subfloat[Video 2]{
	\includegraphics[width=0.49\linewidth, clip=true, trim=5mm 20mm 5mm 20mm]{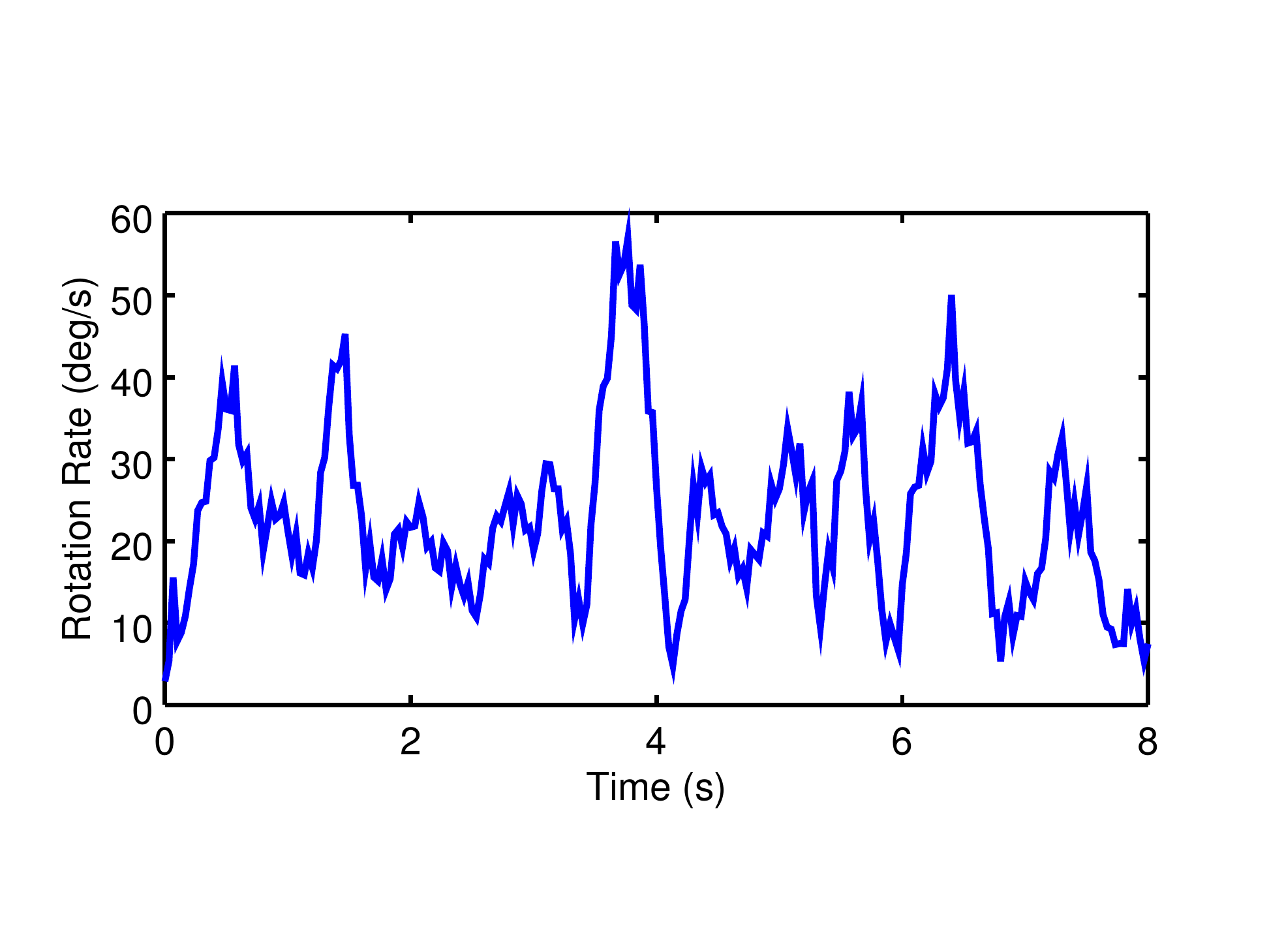}
}\\[-0.3cm]
\subfloat[Video 3]{
	\includegraphics[width=0.49\linewidth, clip=true, trim=5mm 20mm 5mm 20mm]{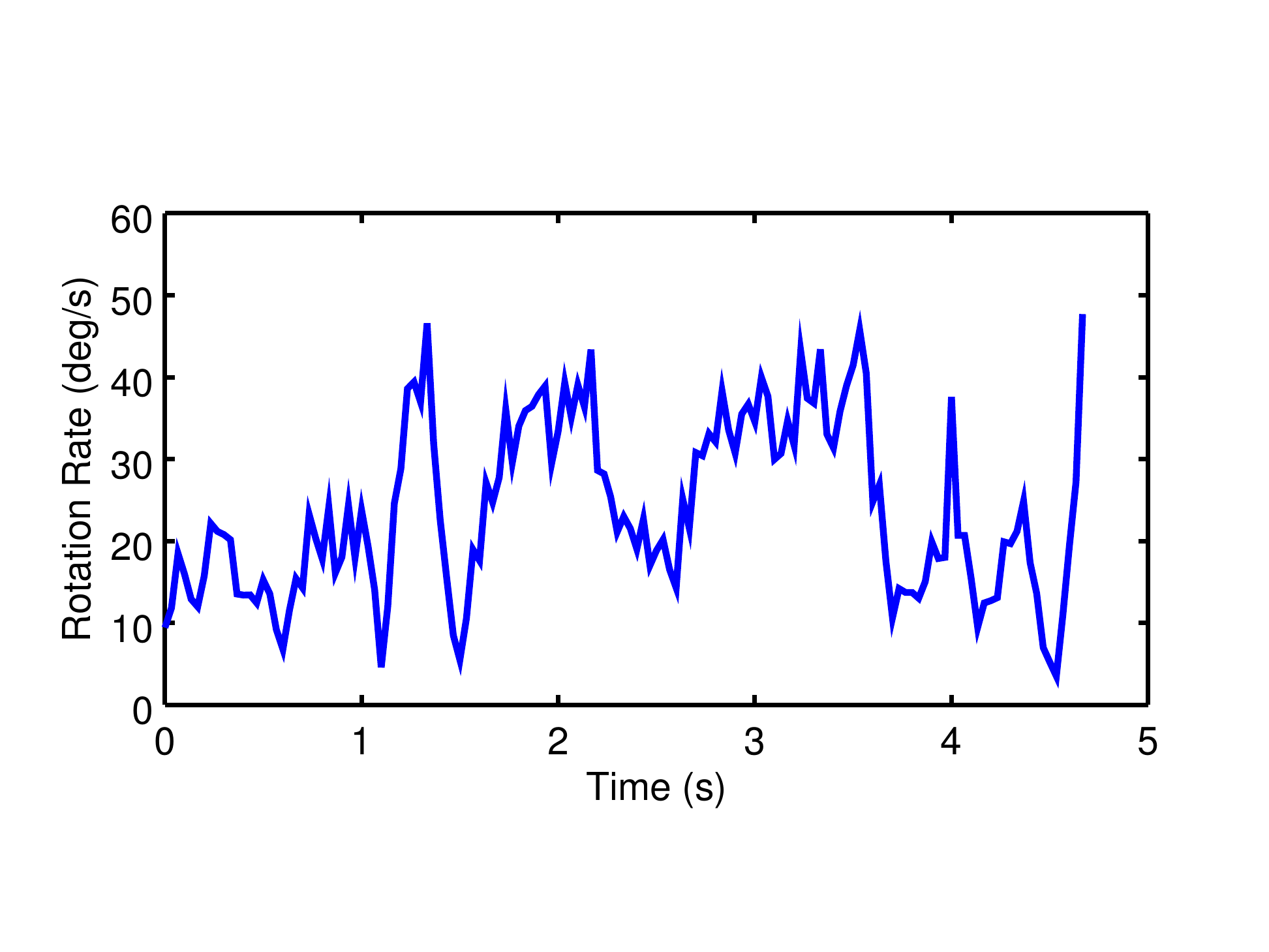}
}
\subfloat[Video 4]{
	\includegraphics[width=0.49\linewidth, clip=true, trim=5mm 20mm 5mm 20mm]{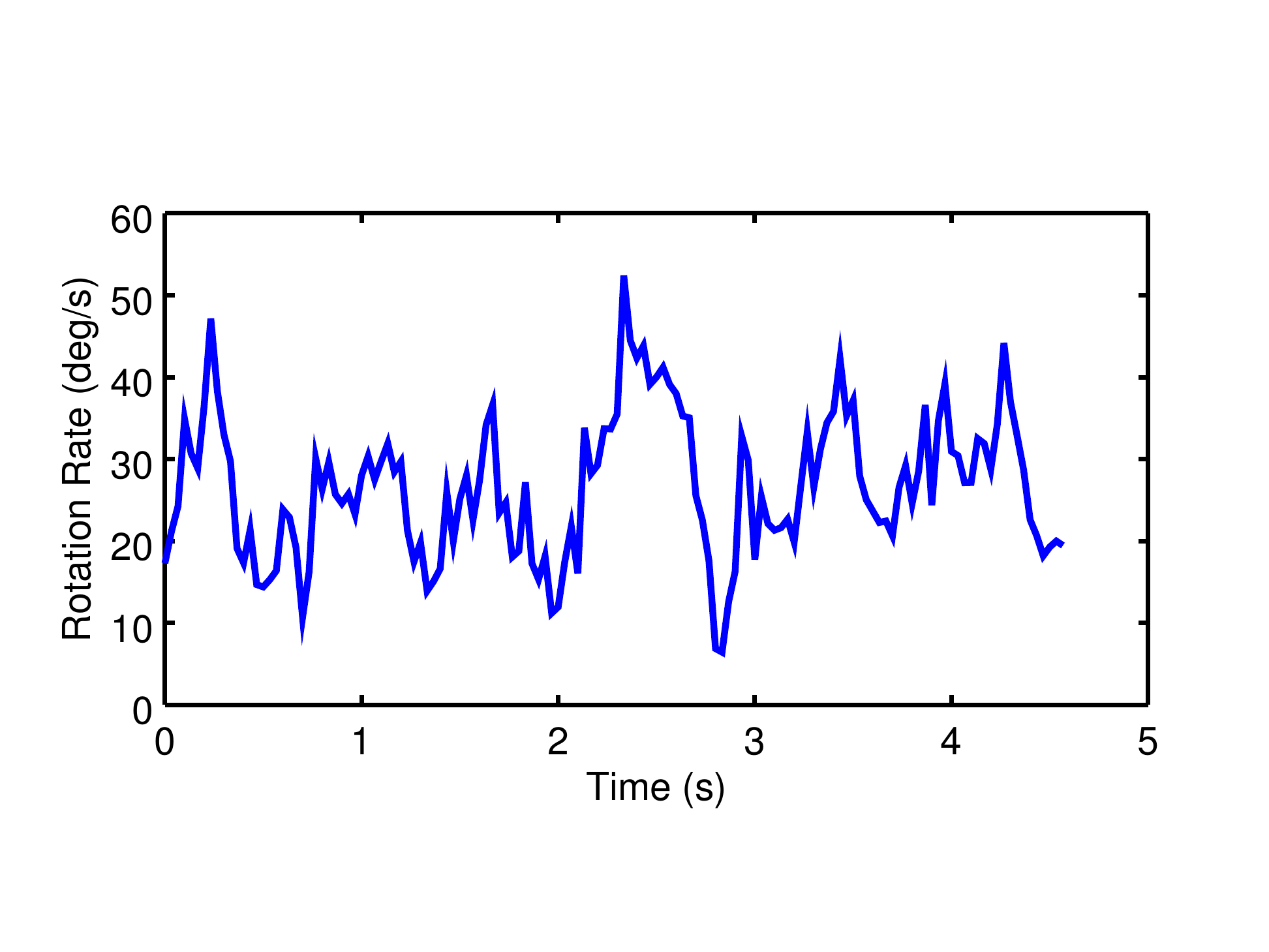}
}\\[-0.3cm]
\subfloat[Video 5]{
	\includegraphics[width=0.49\linewidth, clip=true, trim=5mm 20mm 5mm 20mm]{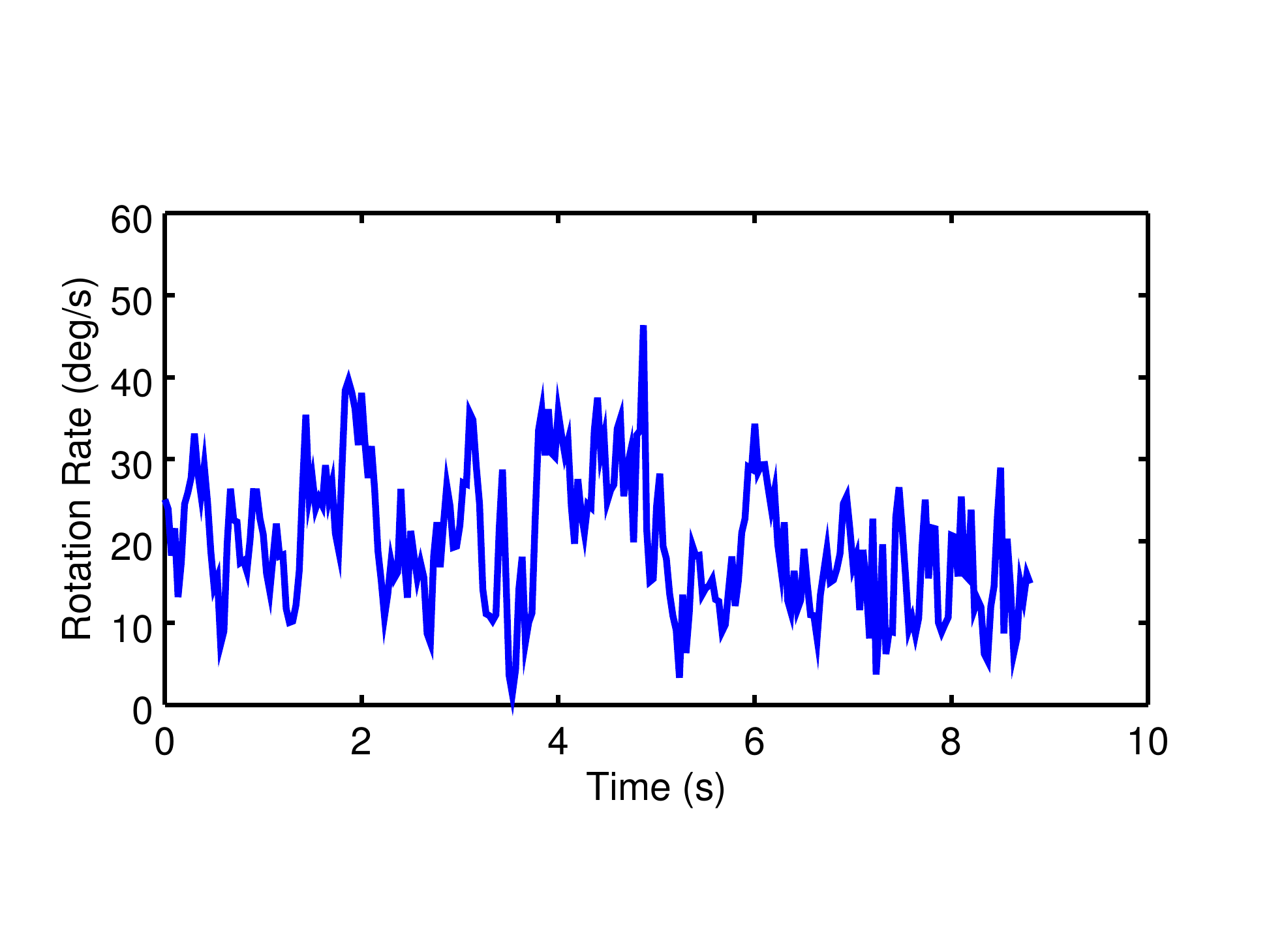}
}
\subfloat[Video 6]{
	\includegraphics[width=0.49\linewidth, clip=true, trim=5mm 20mm 5mm 20mm]{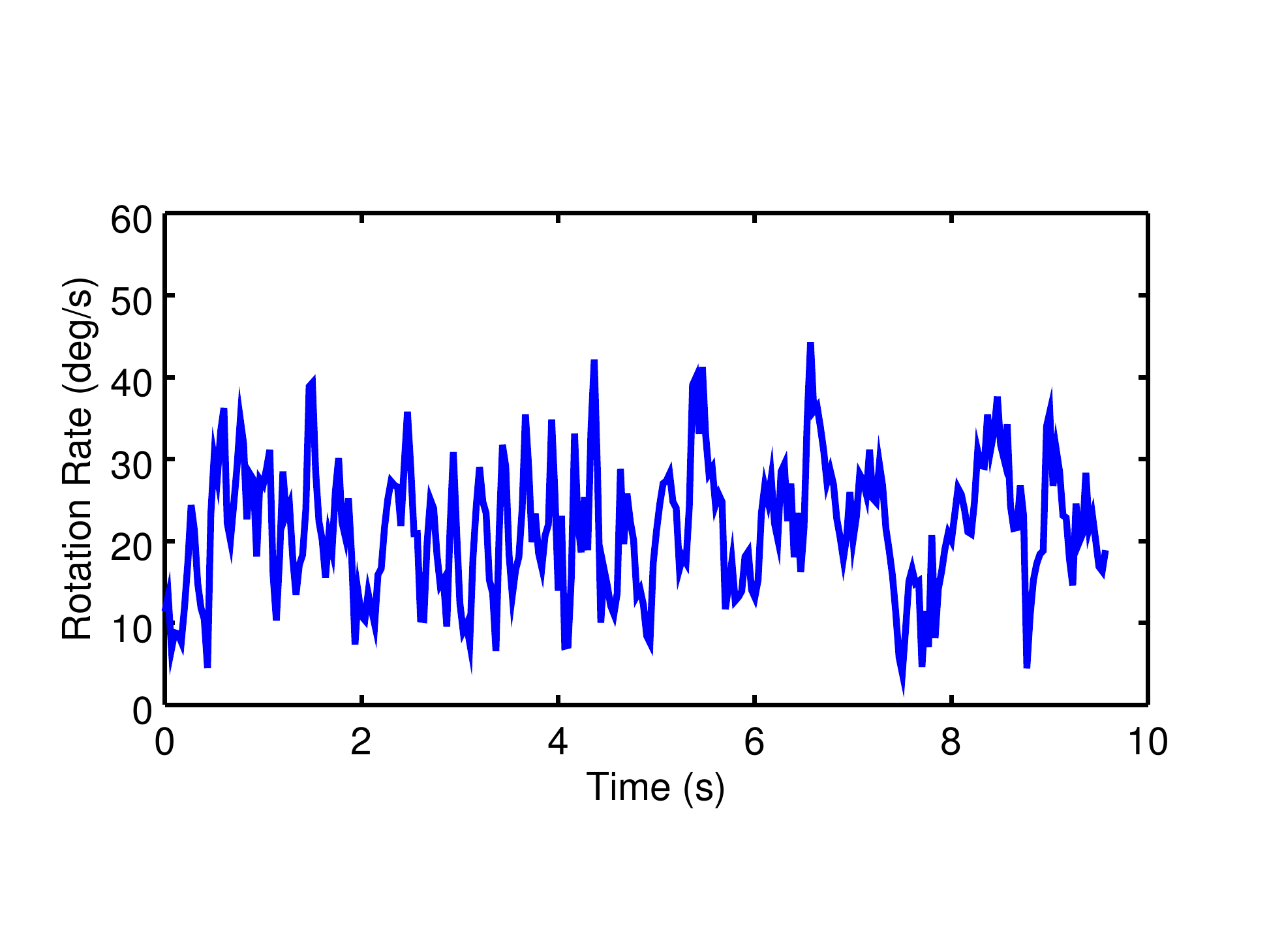}
}\\[-0.3cm]
\subfloat[Video 7]{
	\includegraphics[width=0.49\linewidth, clip=true, trim=5mm 20mm 5mm 20mm]{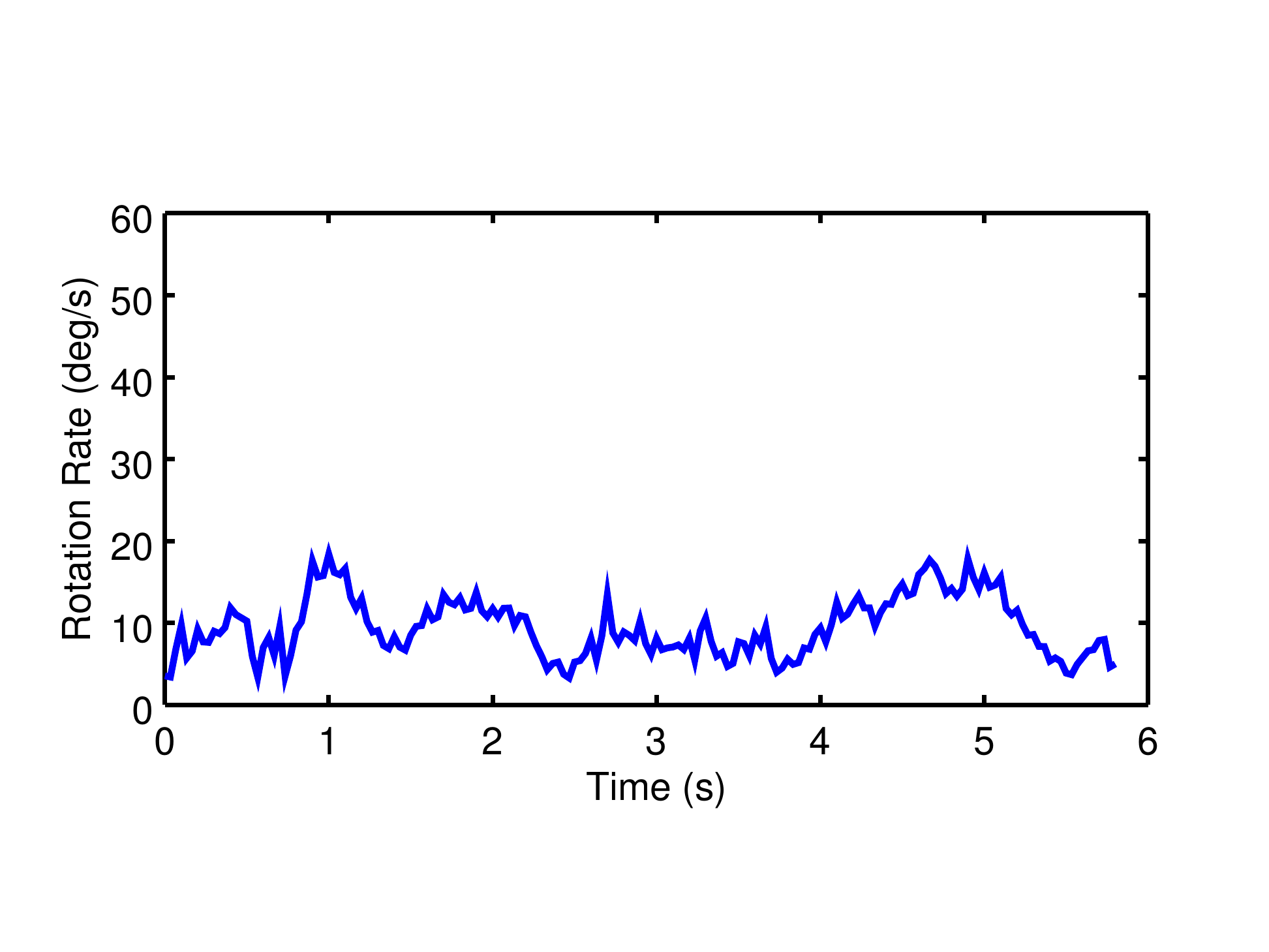}
}
\subfloat[Video 8]{
	\includegraphics[width=0.49\linewidth, clip=true, trim=5mm 20mm 5mm 20mm]{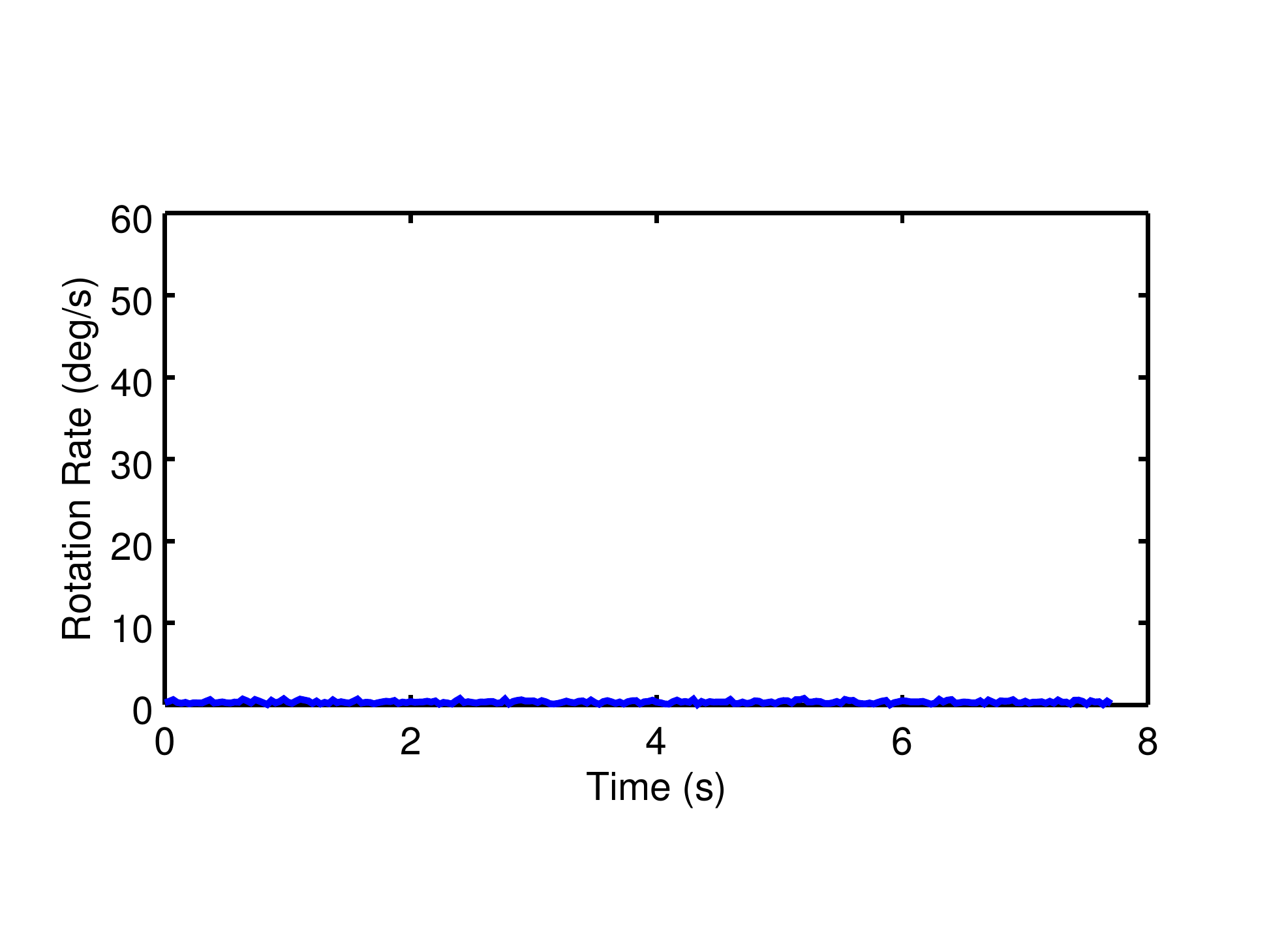}
}
\caption{Rotation Rates as functions of time in Quantitative Test Videos.}
\label{fig:Rotation Rates in Quantitative Test Videos}
\end{figure}

\begin{figure}[H]
\captionsetup{margin=10pt,labelfont=bf}
\centering
\subfloat[Low Quality 1]{
	\includegraphics[width=0.49\linewidth, clip=true, trim=5mm 20mm 5mm 20mm]{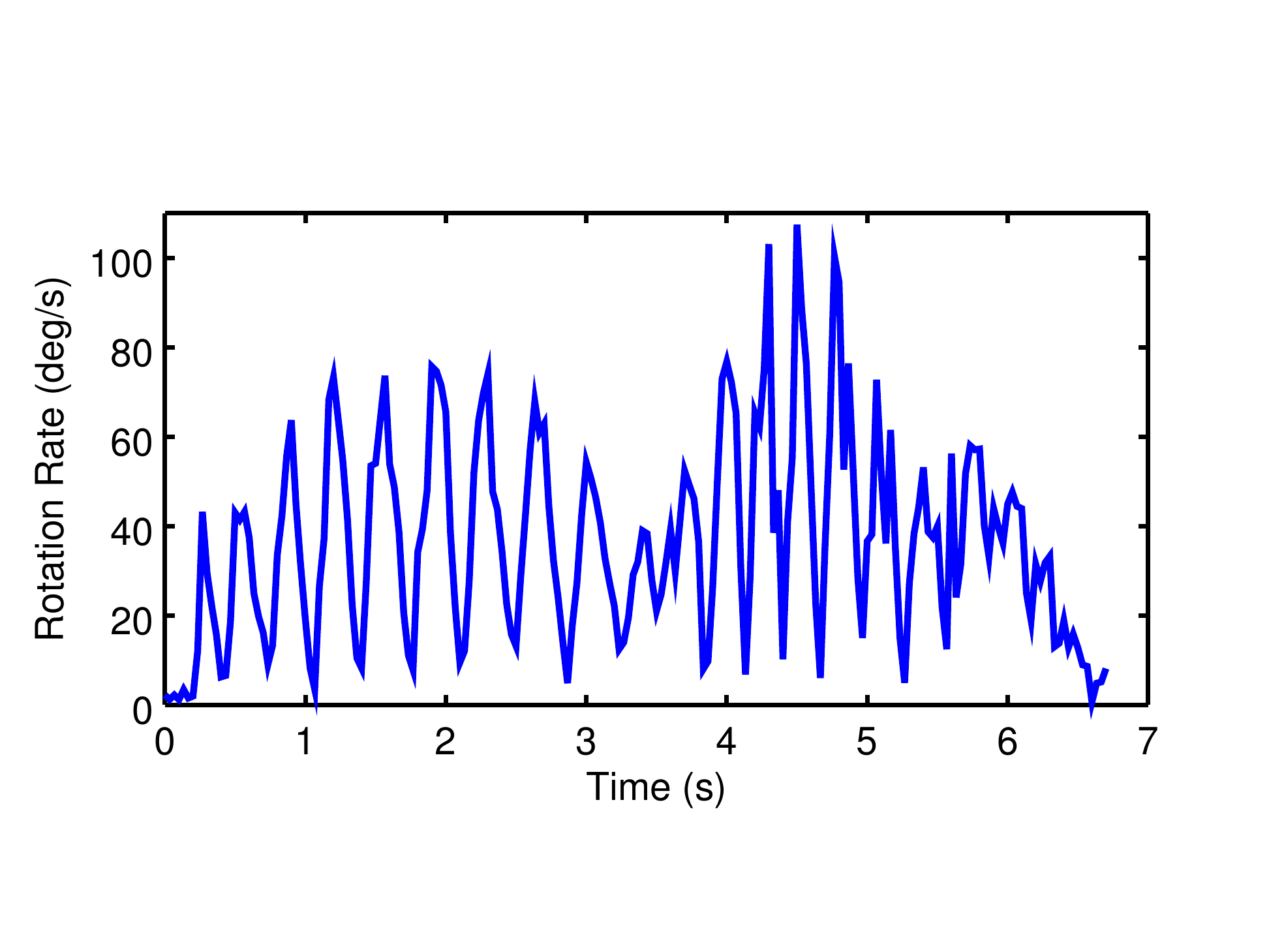}
}
\subfloat[Low Quality 2]{
	\includegraphics[width=0.49\linewidth, clip=true, trim=5mm 20mm 5mm 20mm]{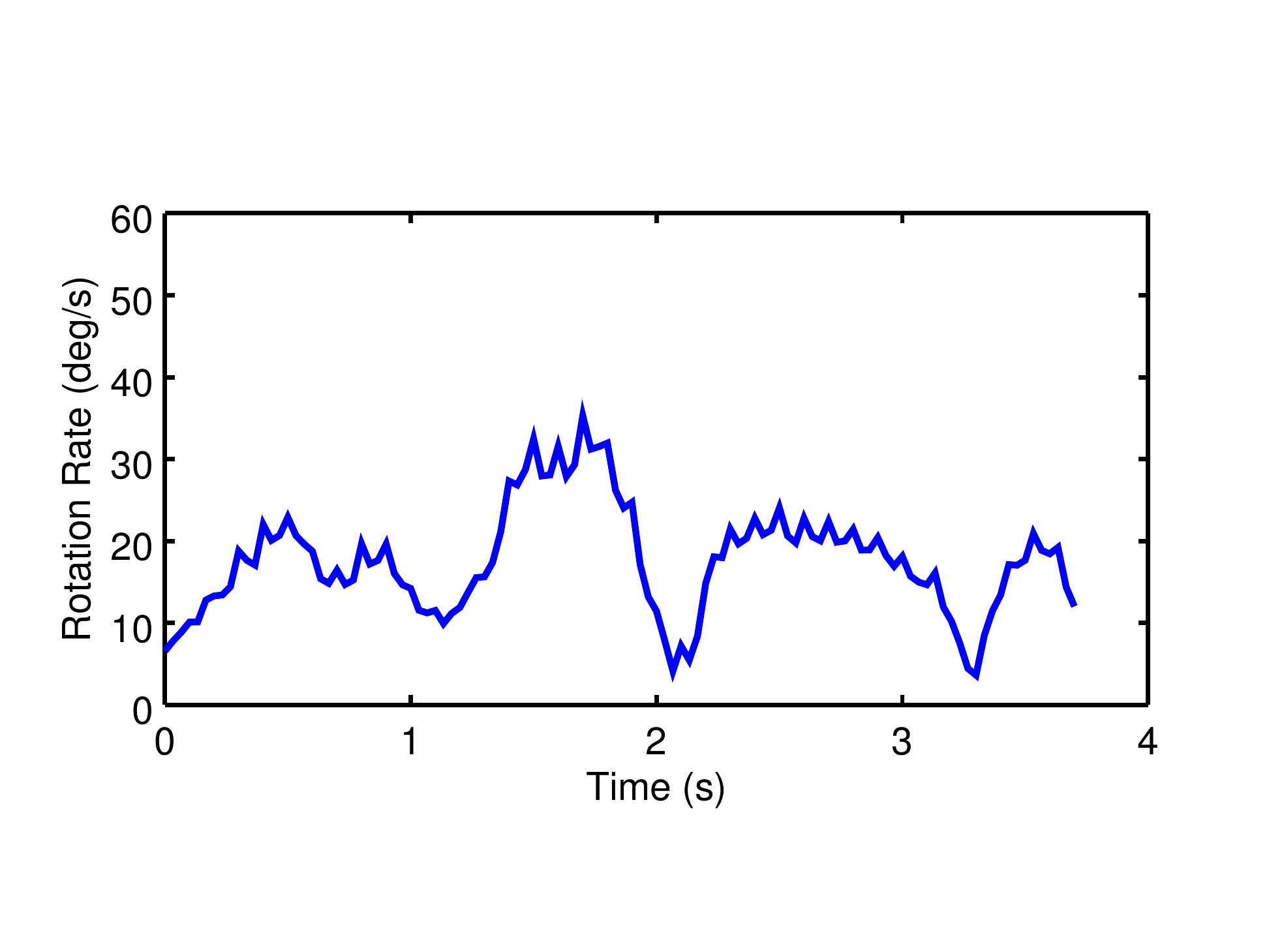}
}\\[-0.3cm]
\subfloat[Low Quality 3]{
	\includegraphics[width=0.49\linewidth, clip=true, trim=5mm 20mm 5mm 20mm]{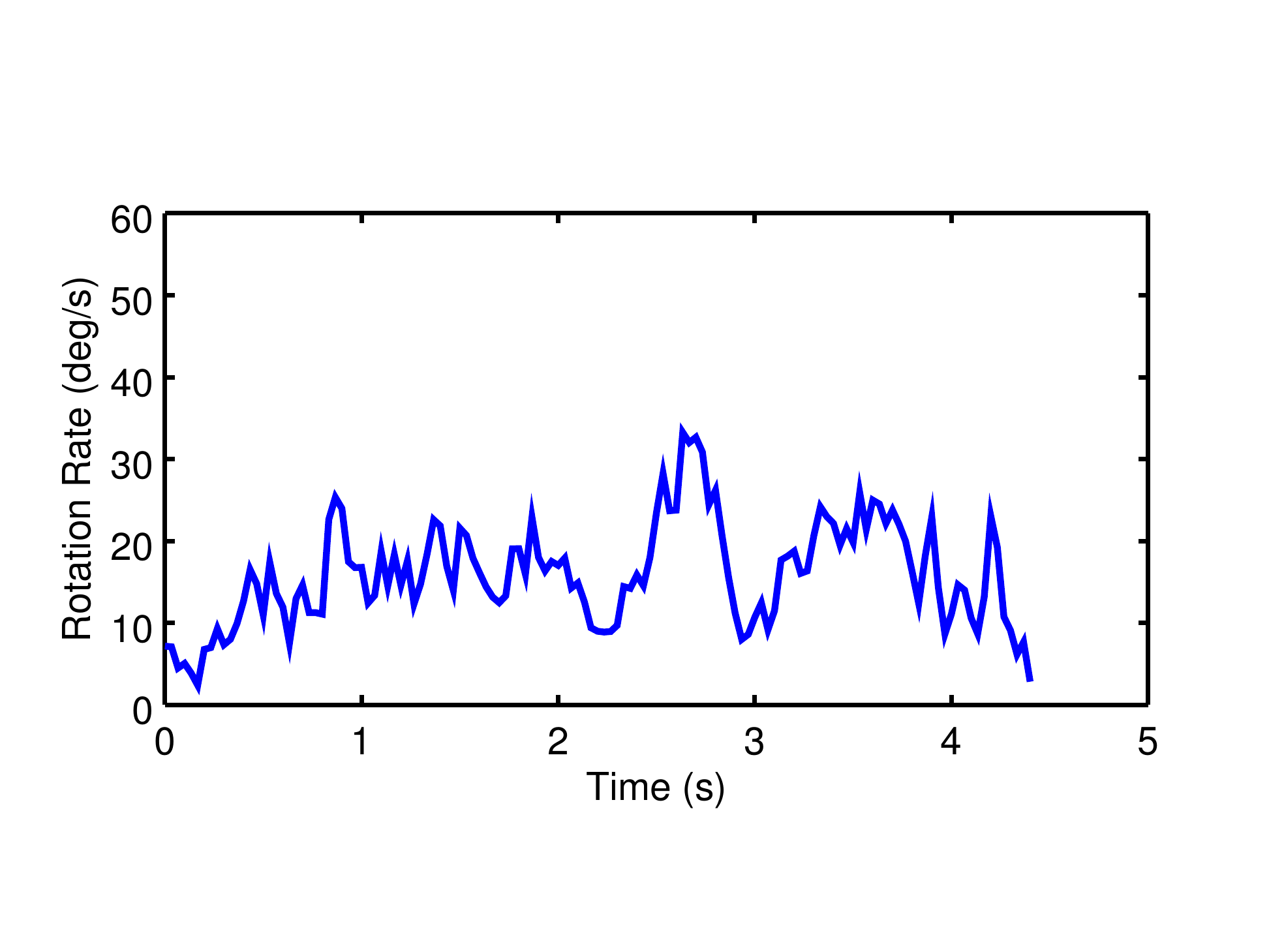}
}
\subfloat[Low Quality 4]{
	\includegraphics[width=0.49\linewidth, clip=true, trim=5mm 20mm 5mm 20mm]{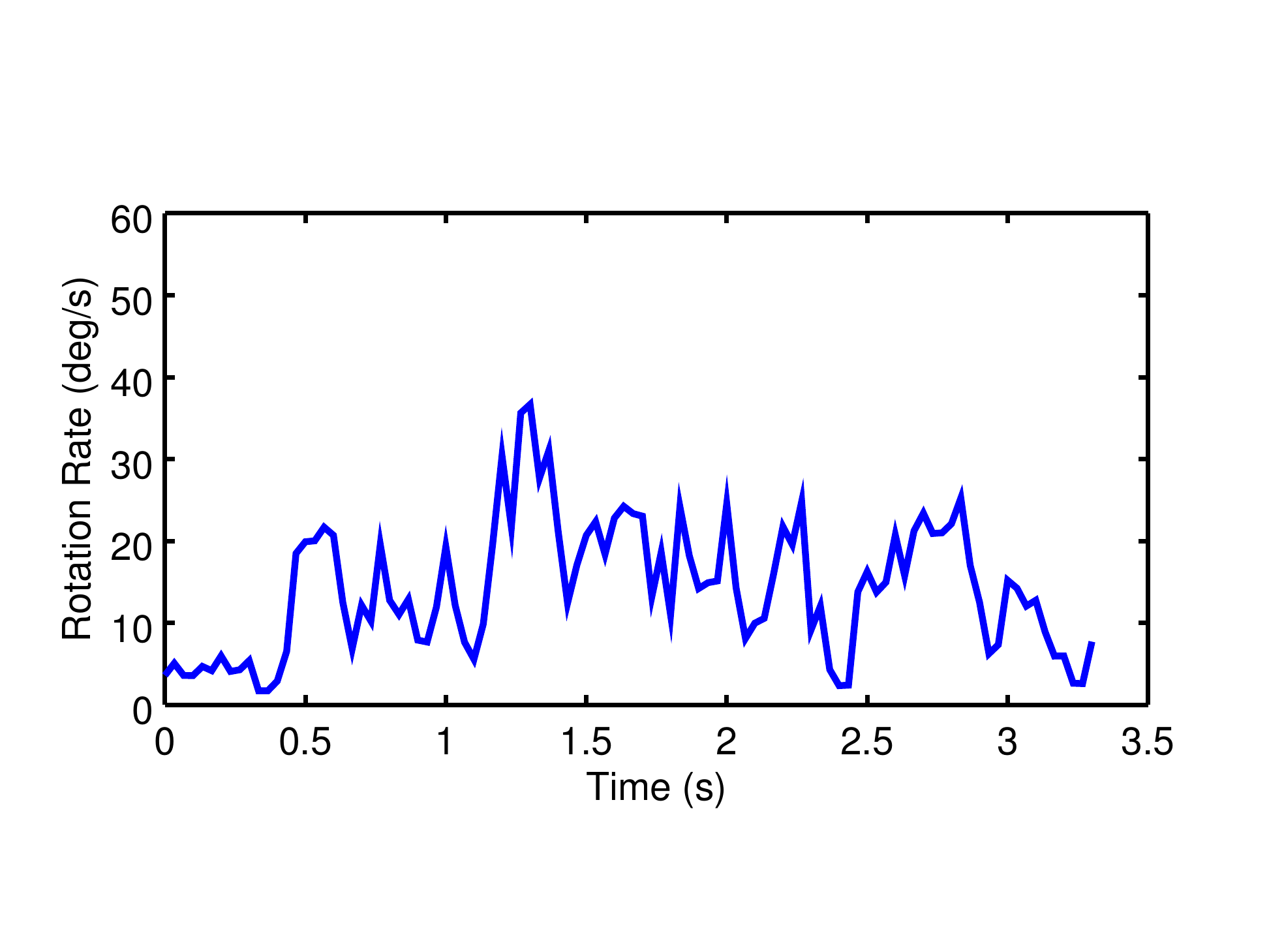}
}\\[-0.3cm]
\subfloat[High Quality 1]{
	\includegraphics[width=0.49\linewidth, clip=true, trim=5mm 20mm 5mm 20mm]{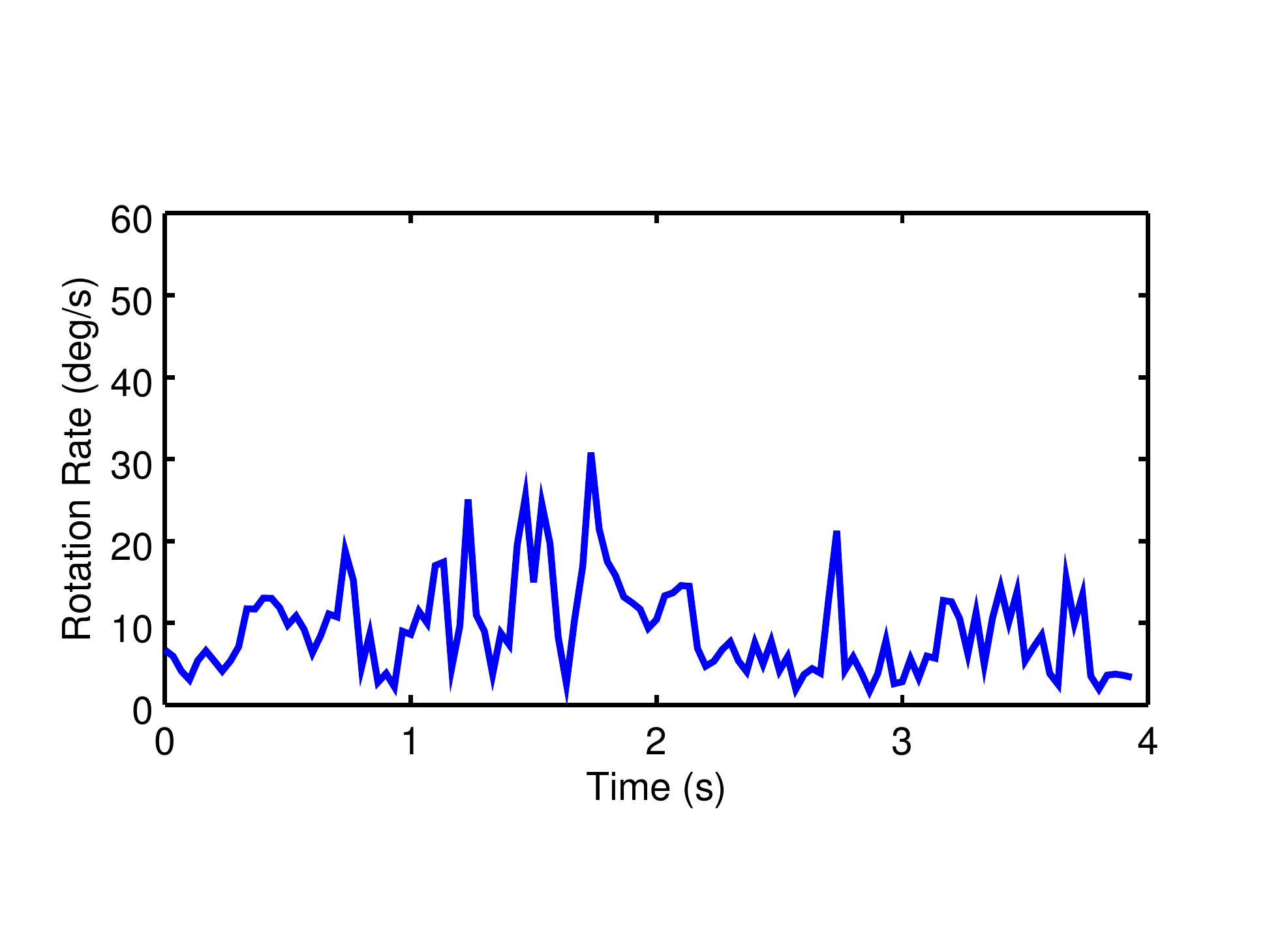}
}
\subfloat[High Quality 2]{
	\includegraphics[width=0.49\linewidth, clip=true, trim=5mm 20mm 5mm 20mm]{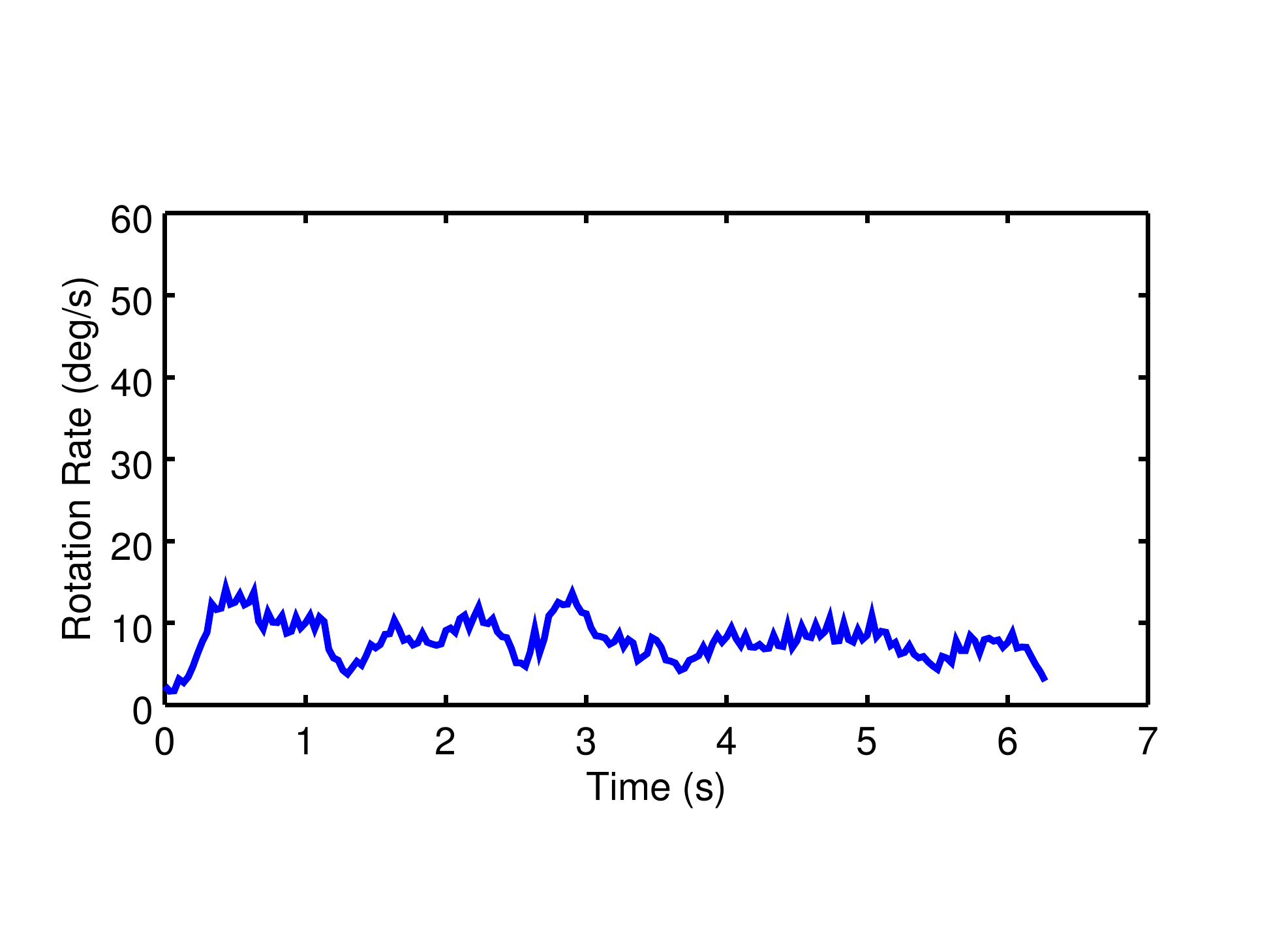}
}\\
\subfloat[High Quality 3]{
	\includegraphics[width=0.49\linewidth, clip=true, trim=5mm 20mm 5mm 20mm]{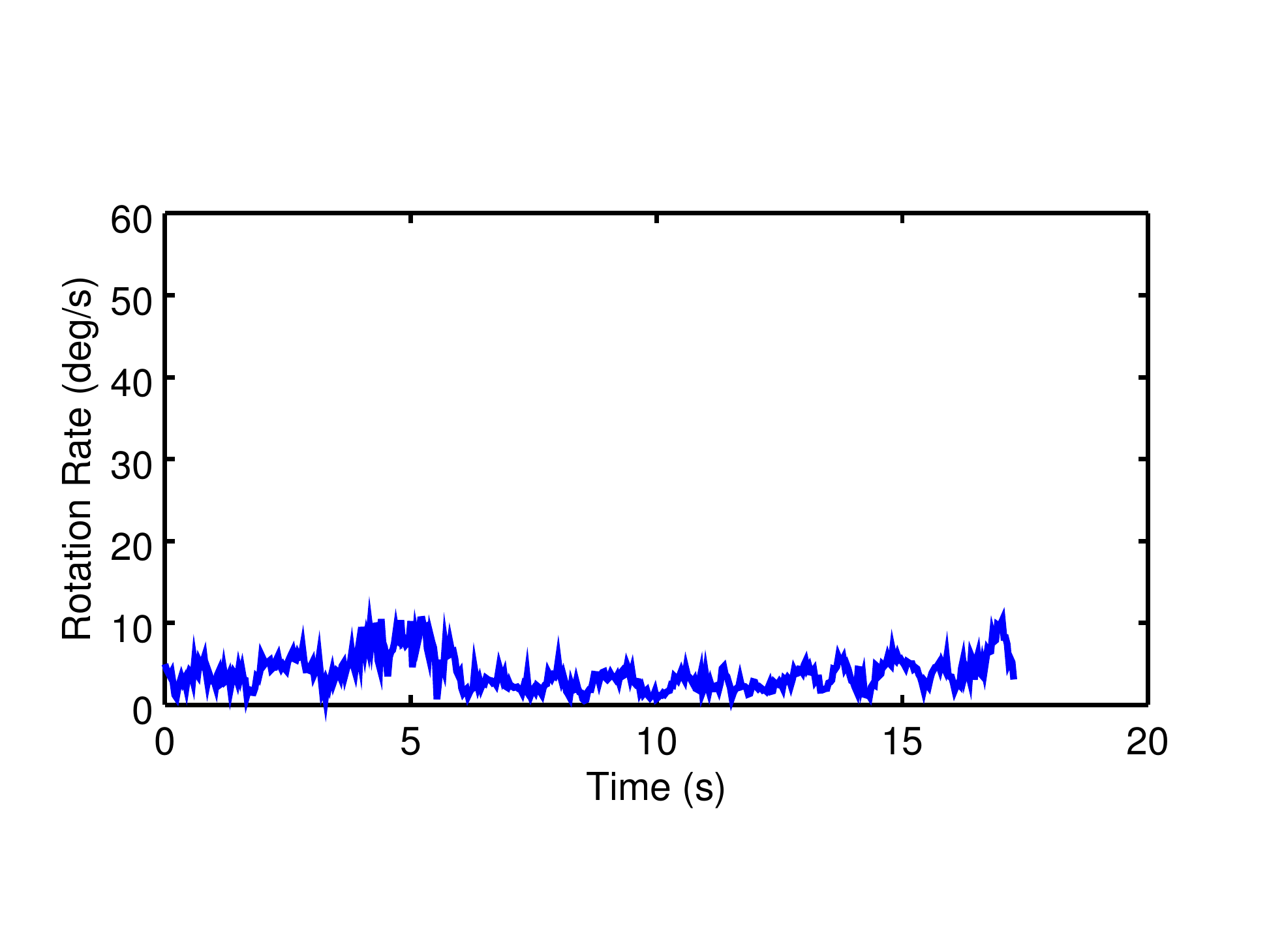}
}
\caption{Rotation Rates as functions of time in Qualitative Test Videos.}
\label{fig:Rotation Rates in Qualitative Test Videos}
\end{figure}

\begin{figure}[H]
\captionsetup{margin=0pt,labelfont=bf}
\centering
\subfloat[KLT + gyro init.]{
	\includegraphics[width=0.49\linewidth, clip=true, trim=0mm 0mm 0mm 0mm]{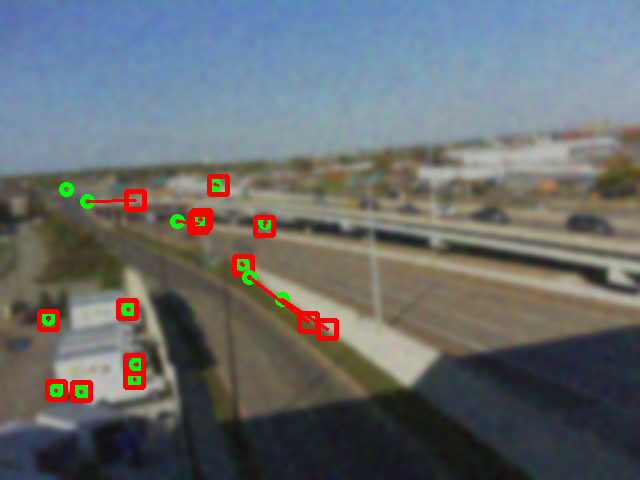}
}
\subfloat[Our method]{
	\includegraphics[width=0.49\linewidth, clip=true, trim=0mm 0mm 0mm 0mm]{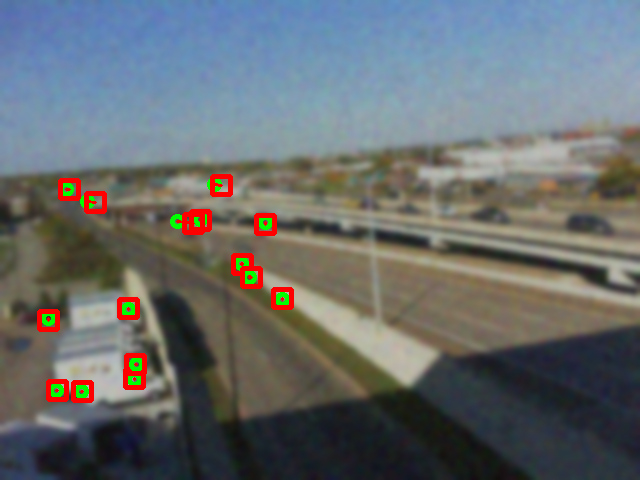}
}\\
\subfloat[KLT + gyro init.]{
	\includegraphics[width=0.49\linewidth, clip=true, trim=0mm 0mm 0mm 0mm]{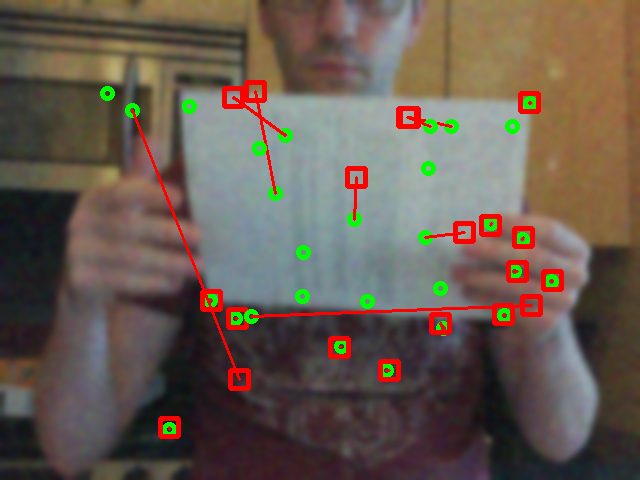}
}
\subfloat[Our method]{
	\includegraphics[width=0.49\linewidth, clip=true, trim=0mm 0mm 0mm 0mm]{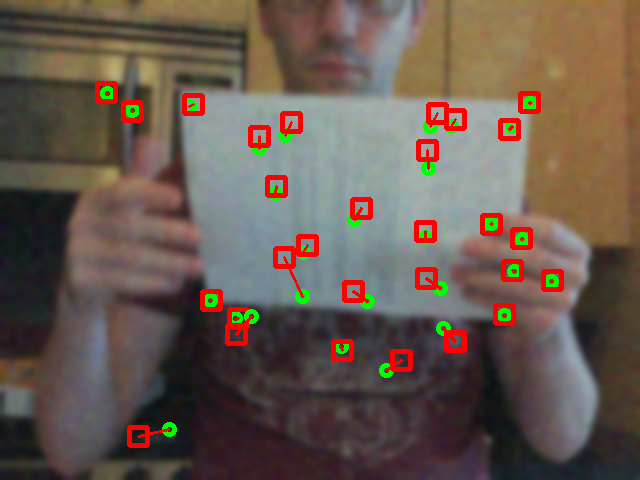}
}
\caption{Characteristic results for KLT with gyro initialization and $1^\text{st}$-order descent tracker with gyro initialization and regularization (denoted ``Our method'') on high-degradation videos when all trackers are initialized with the same set of features and run without being corrected. Tracker output locations (red squares) are connected by red lines to ground-truth locations (green circles). Green circles without squares attached are lost features (drifted off-screen).}
\label{fig:Example Output}
\end{figure}

\begin{figure}[htb]
\captionsetup{margin=0pt,labelfont=bf}
\centering
\tabcolsep=0.0cm
\begin{tabular}{ccccc}
\subfloat{
\begin{sideways}
{\setstretch{1.0}{\scriptsize KLT + Gyro Init}}
\end{sideways}
} &
\subfloat[Frame 0]{
	\includegraphics[width=0.23\linewidth, clip=true, trim=0mm 0mm 0mm 0mm]{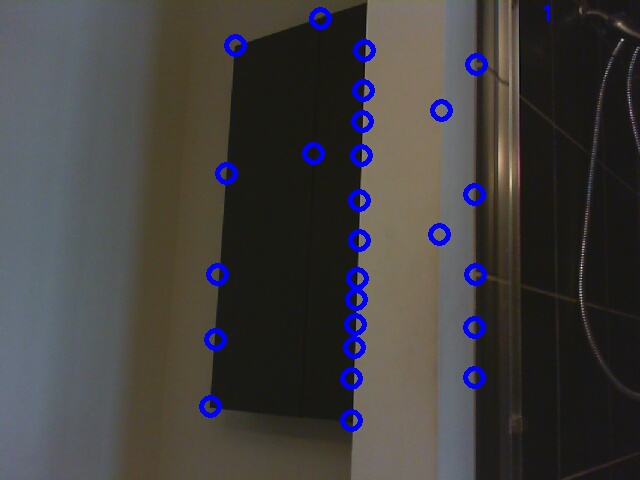}
} &
\subfloat[Frame 10]{
	\includegraphics[width=0.23\linewidth, clip=true, trim=0mm 0mm 0mm 0mm]{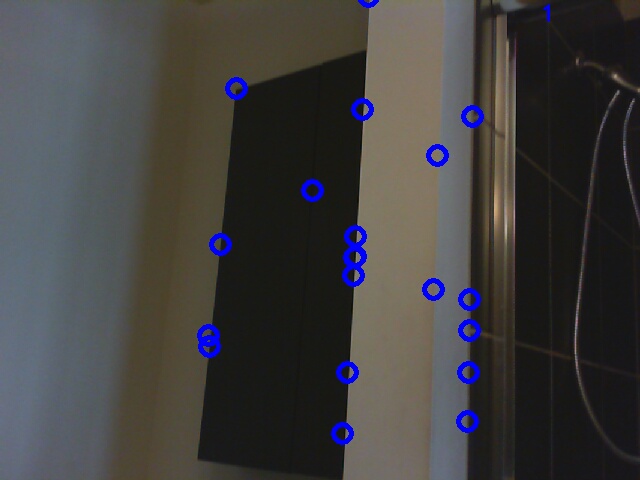}
} &
\subfloat[Frame 20]{
	\includegraphics[width=0.23\linewidth, clip=true, trim=0mm 0mm 0mm 0mm]{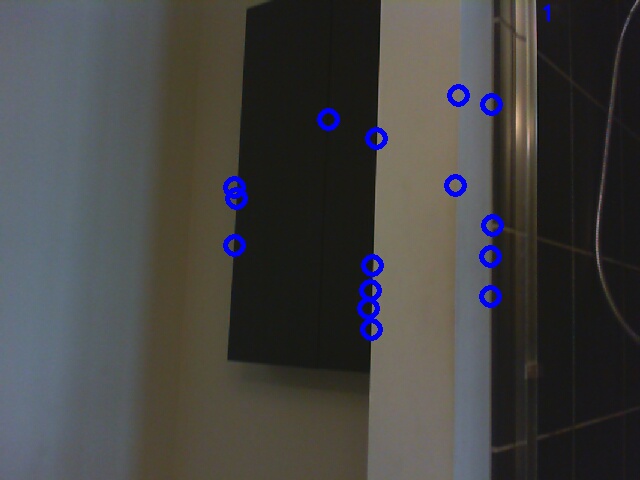}
} &
\subfloat[Frame 30]{
	\includegraphics[width=0.23\linewidth, clip=true, trim=0mm 0mm 0mm 0mm]{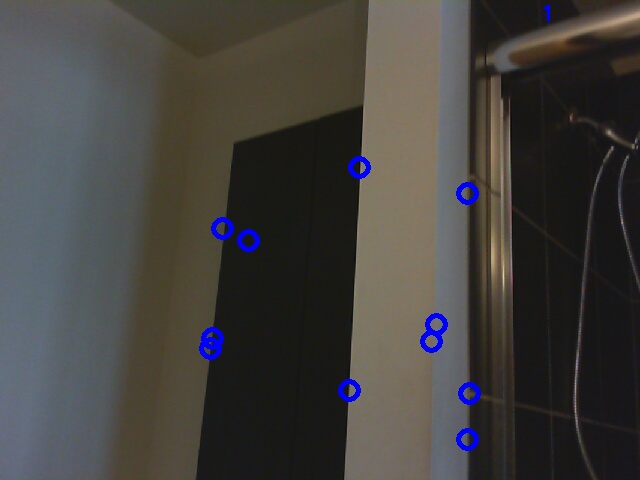}
}\\
\subfloat{
\begin{sideways}
{\setstretch{1.0}{\scriptsize \;\; Our Method}}
\end{sideways}
} &
\subfloat[Frame 0]{
	\includegraphics[width=0.23\linewidth, clip=true, trim=0mm 0mm 0mm 0mm]{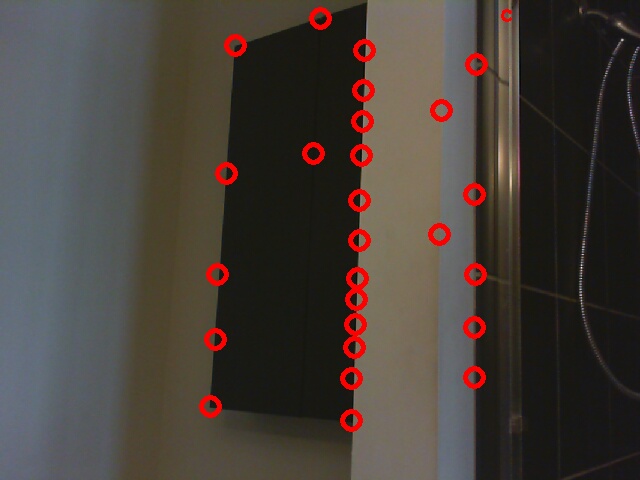}
} &
\subfloat[Frame 10]{
	\includegraphics[width=0.23\linewidth, clip=true, trim=0mm 0mm 0mm 0mm]{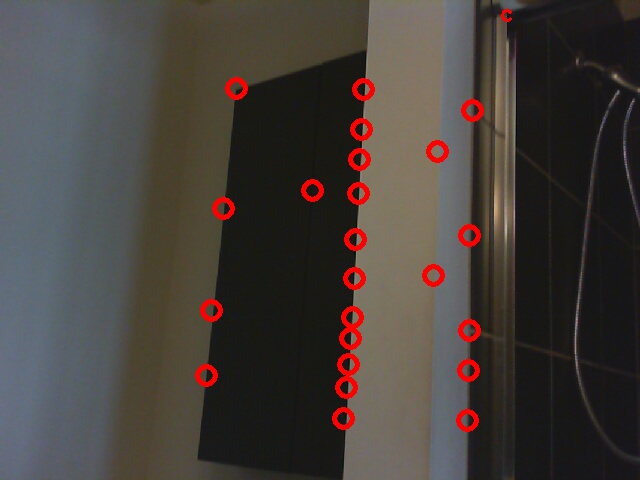}
} &
\subfloat[Frame 20]{
	\includegraphics[width=0.23\linewidth, clip=true, trim=0mm 0mm 0mm 0mm]{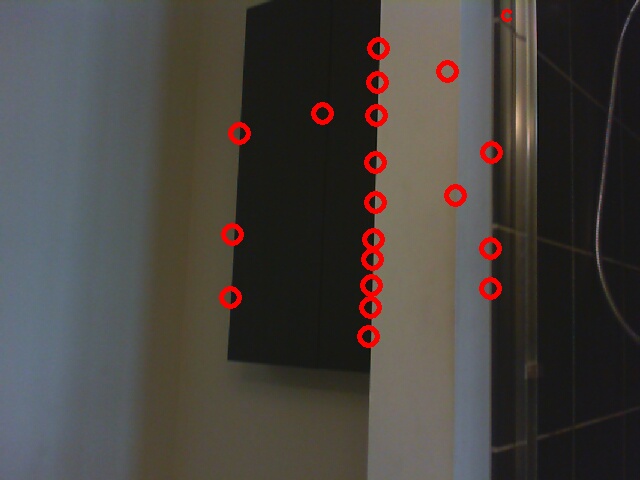}
} &
\subfloat[Frame 30]{
	\includegraphics[width=0.23\linewidth, clip=true, trim=0mm 0mm 0mm 0mm]{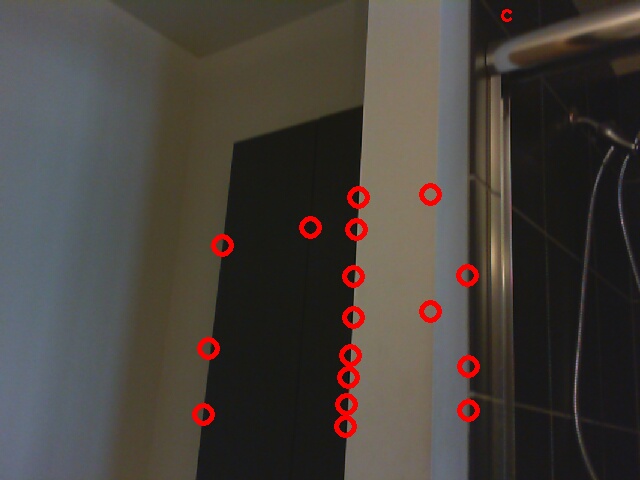}
}
\end{tabular}
\caption{\label{fig:Real Video Output 1}Sample frames showing the output of KLT + Gyro Initialization (blue circles) and our method (red circles) on a real-world low-light video (``Low Quality 1'') with many ambiguous features. Notice the number of features that are lost by KLT + Gyro Init.}
\end{figure}

\begin{figure}[h!]
\captionsetup{margin=0pt,labelfont=bf}
\centering
\tabcolsep=0.0cm
\begin{tabular}{ccccc}
\subfloat{
\begin{sideways}
{\setstretch{1.0}{\scriptsize KLT + Gyro Init}}
\end{sideways}
} &
\subfloat[Frame 0]{
	\includegraphics[width=0.23\linewidth, clip=true, trim=0mm 0mm 0mm 0mm]{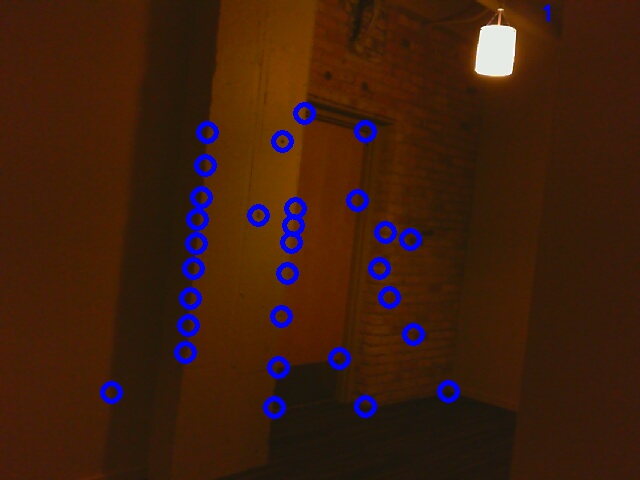}
} &
\subfloat[Frame 20]{
	\includegraphics[width=0.23\linewidth, clip=true, trim=0mm 0mm 0mm 0mm]{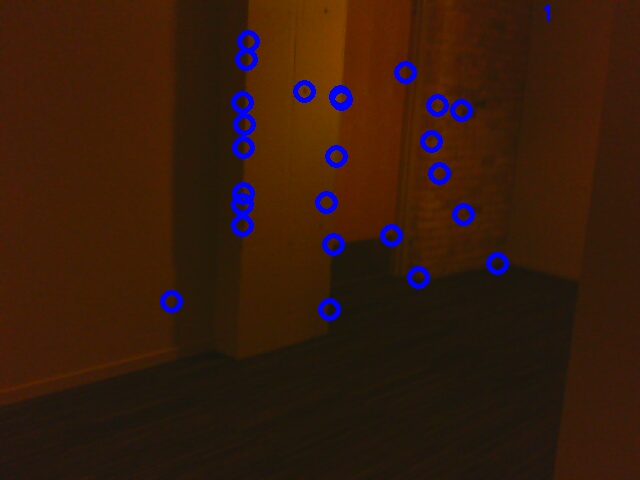}
} &
\subfloat[Frame 40]{
	\includegraphics[width=0.23\linewidth, clip=true, trim=0mm 0mm 0mm 0mm]{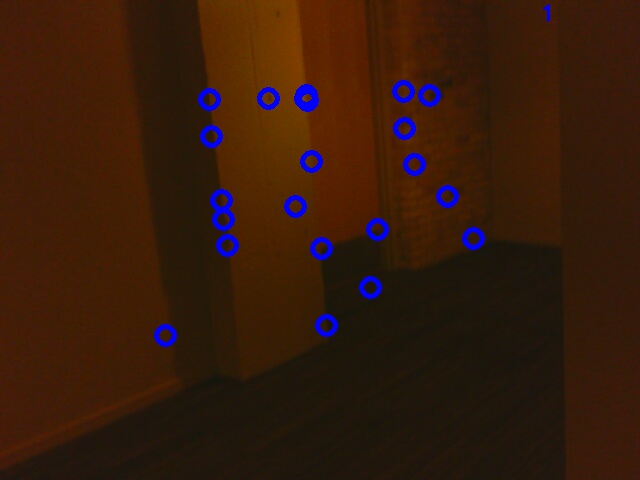}
} &
\subfloat[Frame 60]{
	\includegraphics[width=0.23\linewidth, clip=true, trim=0mm 0mm 0mm 0mm]{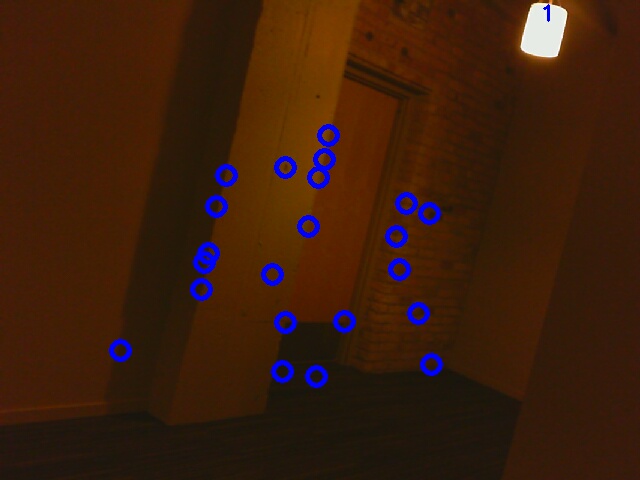}
}\\
\subfloat{
\begin{sideways}
{\setstretch{1.0}{\scriptsize \;\; Our Method}}
\end{sideways}
} &
\subfloat[Frame 0]{
	\includegraphics[width=0.23\linewidth, clip=true, trim=0mm 0mm 0mm 0mm]{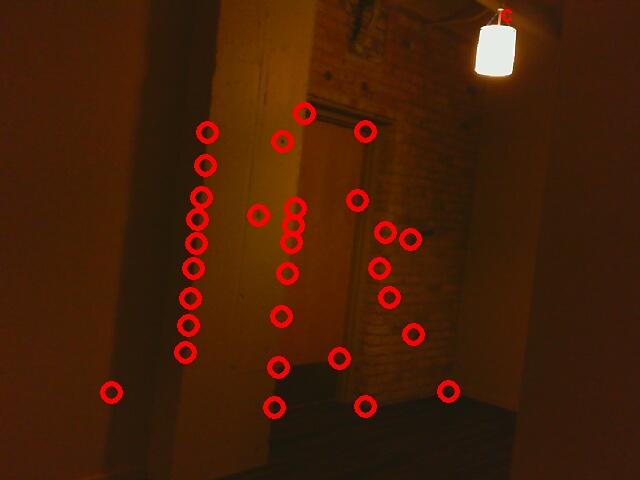}
} &
\subfloat[Frame 20]{
	\includegraphics[width=0.23\linewidth, clip=true, trim=0mm 0mm 0mm 0mm]{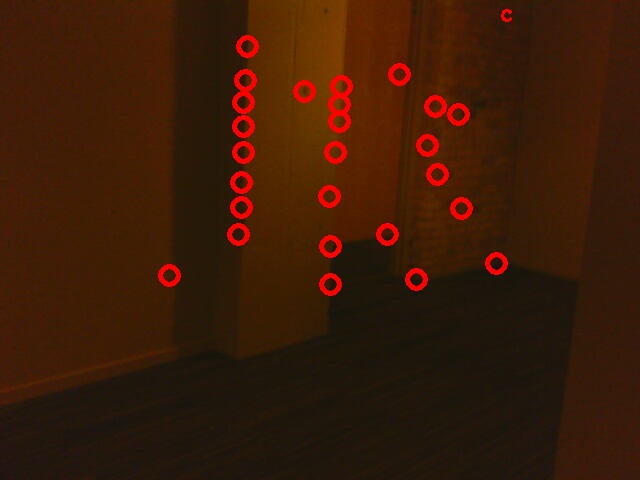}
} &
\subfloat[Frame 40]{
	\includegraphics[width=0.23\linewidth, clip=true, trim=0mm 0mm 0mm 0mm]{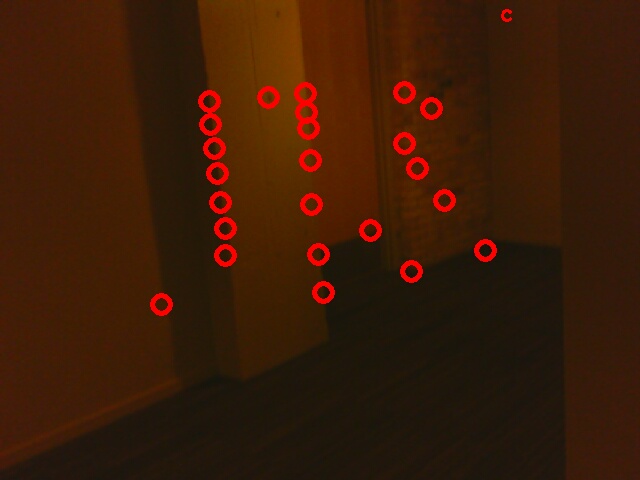}
} &
\subfloat[Frame 60]{
	\includegraphics[width=0.23\linewidth, clip=true, trim=0mm 0mm 0mm 0mm]{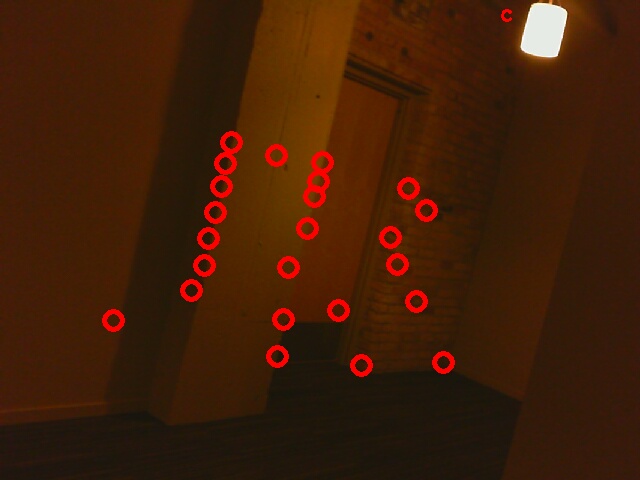}
}
\end{tabular}
\caption{\label{fig:Real Video Output 2}Sample frames showing the output of KLT + Gyro Initialization (blue circles) and our method (red circles) on a real-world low-light video (``Low Quality 2'') with many ambiguous features. Notice the bunching of features along the column edges with KLT + Gyro Init.}
\end{figure}

\begin{figure}[h!]
\captionsetup{margin=0pt,labelfont=bf}
\centering
\tabcolsep=0.0cm
\begin{tabular}{ccccc}
\subfloat{
\begin{sideways}
{\setstretch{1.0}{\scriptsize KLT + Gyro Init}}
\end{sideways}
} &
\subfloat[Frame 0]{
	\includegraphics[width=0.23\linewidth, clip=true, trim=0mm 0mm 0mm 0mm]{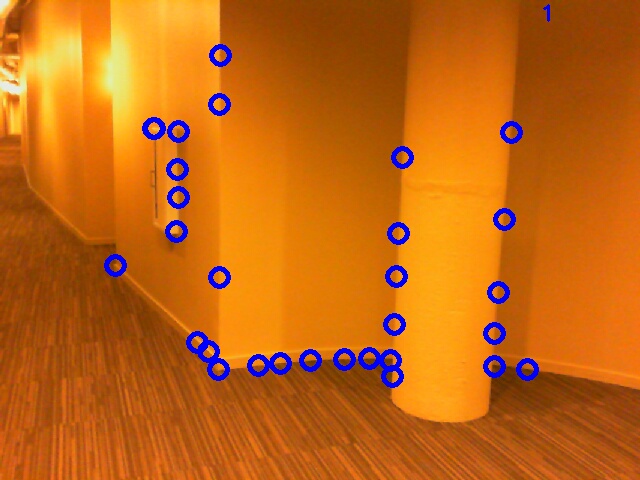}
} &
\subfloat[Frame 20]{
	\includegraphics[width=0.23\linewidth, clip=true, trim=0mm 0mm 0mm 0mm]{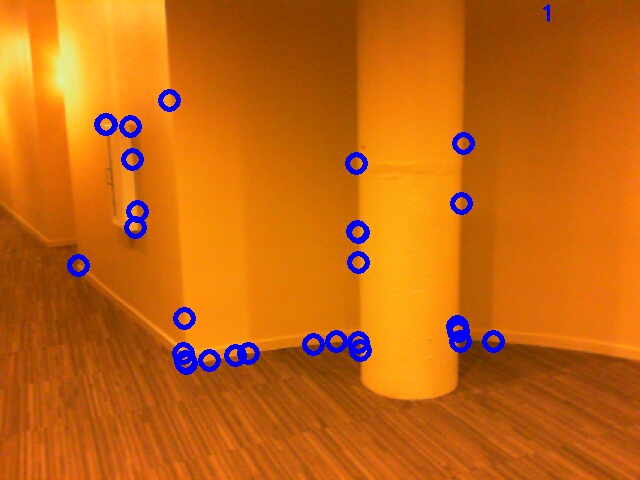}
} &
\subfloat[Frame 40]{
	\includegraphics[width=0.23\linewidth, clip=true, trim=0mm 0mm 0mm 0mm]{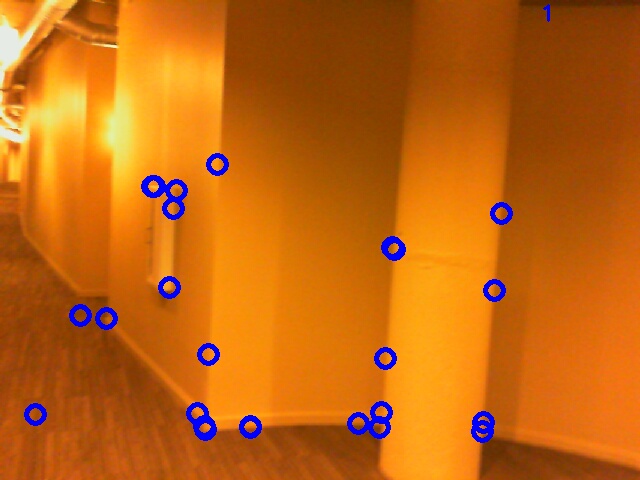}
} &
\subfloat[Frame 60]{
	\includegraphics[width=0.23\linewidth, clip=true, trim=0mm 0mm 0mm 0mm]{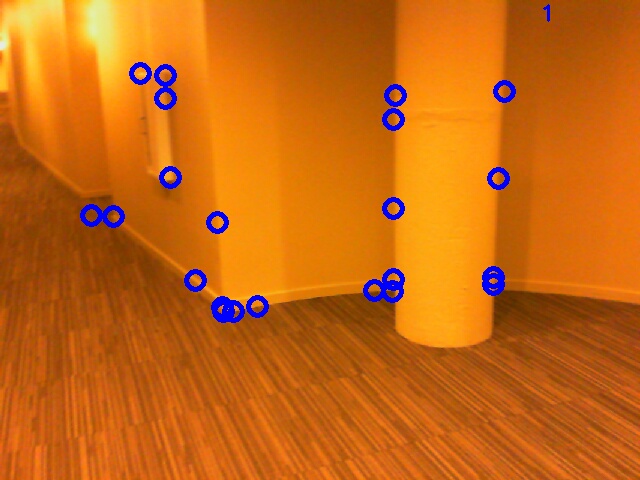}
}\\
\subfloat{
\begin{sideways}
{\setstretch{1.0}{\scriptsize \;\; Our Method}}
\end{sideways}
} &
\subfloat[Frame 0]{
	\includegraphics[width=0.23\linewidth, clip=true, trim=0mm 0mm 0mm 0mm]{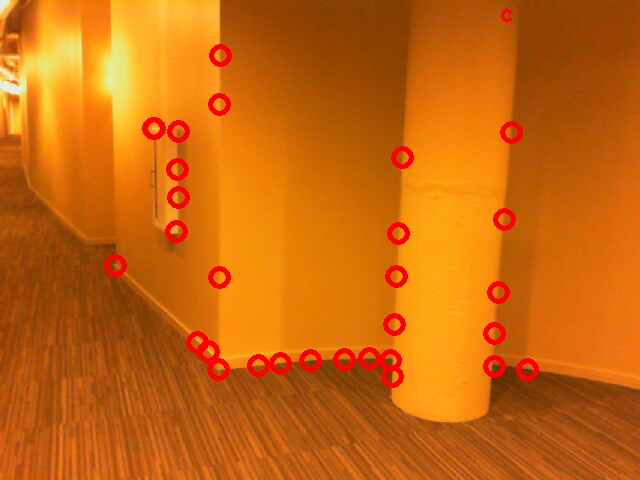}
} &
\subfloat[Frame 20]{
	\includegraphics[width=0.23\linewidth, clip=true, trim=0mm 0mm 0mm 0mm]{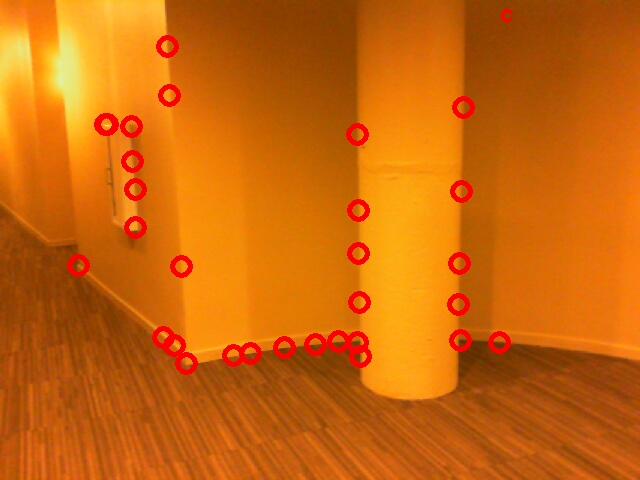}
} &
\subfloat[Frame 40]{
	\includegraphics[width=0.23\linewidth, clip=true, trim=0mm 0mm 0mm 0mm]{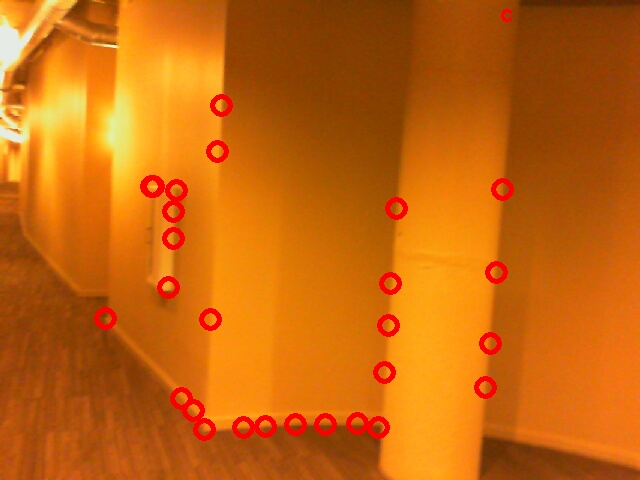}
} &
\subfloat[Frame 60]{
	\includegraphics[width=0.23\linewidth, clip=true, trim=0mm 0mm 0mm 0mm]{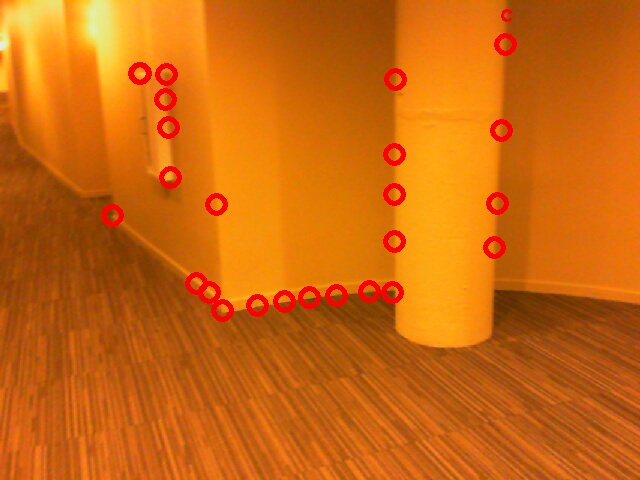}
}
\end{tabular}
\caption{\label{fig:Real Video Output 3}Sample frames showing the output of KLT + Gyro Initialization (blue circles) and our method (red circles) on a real-world low-light video (``Low Quality 3'') with many ambiguous features. Notice the ambiguous features wandering and accumulating in corner areas with KLT + Gyro Init.}
\end{figure}

\begin{figure}[h]
\captionsetup{margin=0pt,labelfont=bf}
\centering
\tabcolsep=0.0cm
\begin{tabular}{ccccc}
\subfloat{
\begin{sideways}
{\setstretch{1.0}{\scriptsize KLT + Gyro Init}}
\end{sideways}
} &
\subfloat[Frame 0]{
	\includegraphics[width=0.23\linewidth, clip=true, trim=0mm 0mm 0mm 0mm]{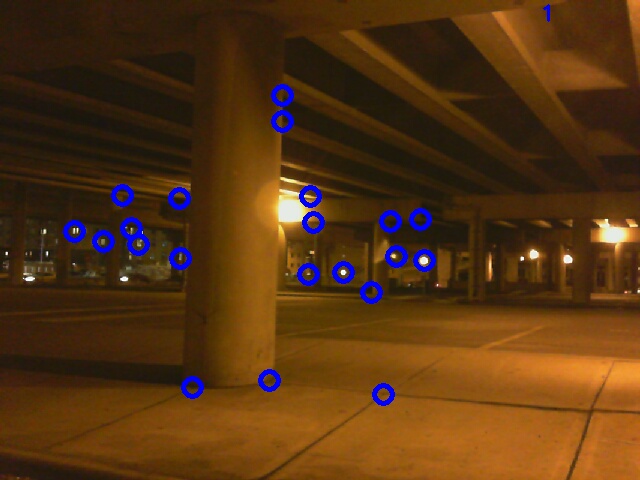}
} &
\subfloat[Frame 20]{
	\includegraphics[width=0.23\linewidth, clip=true, trim=0mm 0mm 0mm 0mm]{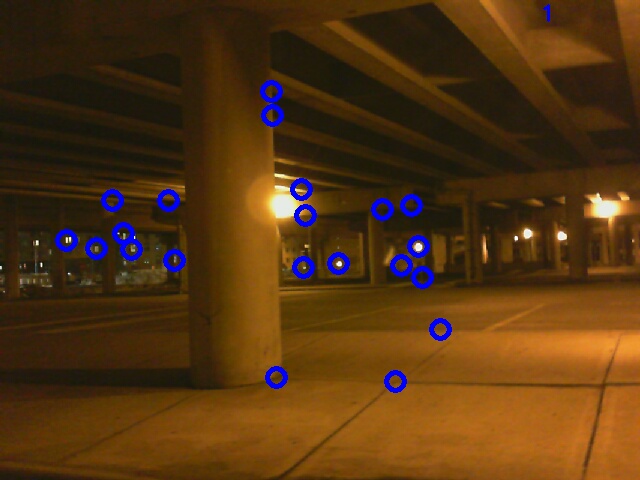}
} &
\subfloat[Frame 40]{
	\includegraphics[width=0.23\linewidth, clip=true, trim=0mm 0mm 0mm 0mm]{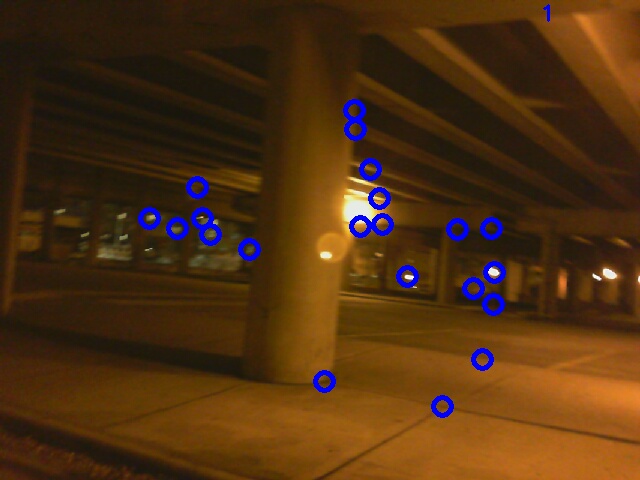}
} &
\subfloat[Frame 60]{
	\includegraphics[width=0.23\linewidth, clip=true, trim=0mm 0mm 0mm 0mm]{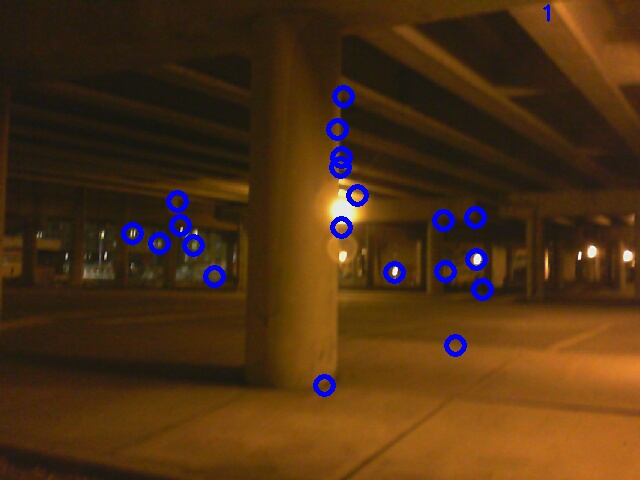}
}\\
\subfloat{
\begin{sideways}
{\setstretch{1.0}{\scriptsize \;\; Our Method}}
\end{sideways}
} &
\subfloat[Frame 0]{
	\includegraphics[width=0.23\linewidth, clip=true, trim=0mm 0mm 0mm 0mm]{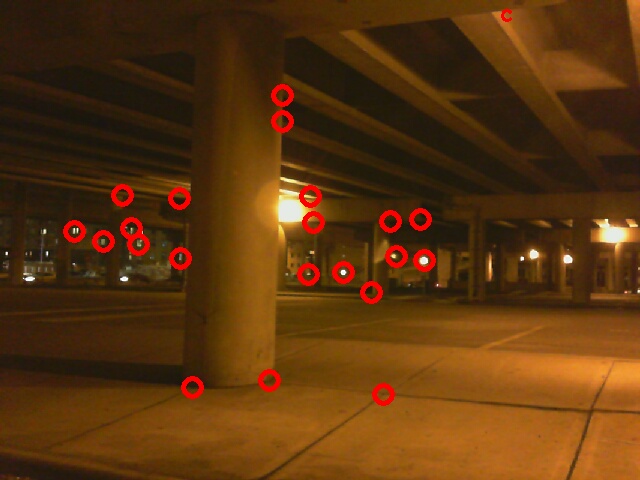}
} &
\subfloat[Frame 20]{
	\includegraphics[width=0.23\linewidth, clip=true, trim=0mm 0mm 0mm 0mm]{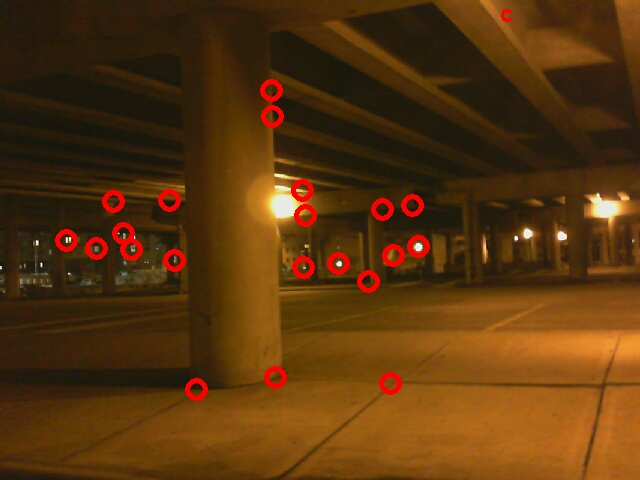}
} &
\subfloat[Frame 40]{
	\includegraphics[width=0.23\linewidth, clip=true, trim=0mm 0mm 0mm 0mm]{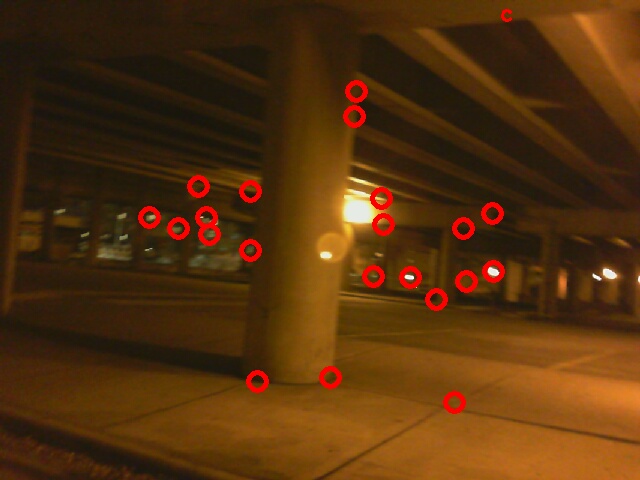}
} &
\subfloat[Frame 60]{
	\includegraphics[width=0.23\linewidth, clip=true, trim=0mm 0mm 0mm 0mm]{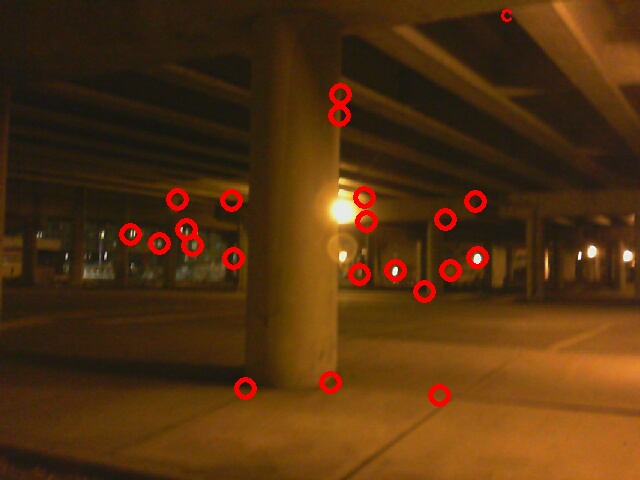}
}
\end{tabular}
\caption{\label{fig:Real Video Output 4}Sample frames showing the output of KLT + Gyro Initialization (blue circles) and our method (red circles) on a real-world low-light video (``Low Quality 4'') with many ambiguous features. Notice that several features ``jump'' with KLT + Gyro Init (lower left of pillar and along the middle section of the image frame).}
\end{figure}

To evaluate our method, we collected multiple video sequences using a custom-built gyro-optical data collection system. This system simultaneously collects imagery from a standard webcam and gyro data from an ST L3GD20 3-axis MEMS gyroscope (see \ref{Appendix: Data Collection Hardware} for more hardware information). A quantitative evaluation of feature trackers requires trustworthy ground-truth trajectories to be known for a large set of features. Manual feature registration or manual correction of feature trajectories in real-world, low-quality video is both challenging and somewhat subjective. Thus, we collected many sequences under mostly favorable conditions and synthetically degraded their quality. We used a standard KLT tracker on the non-degraded videos to generate ground-truth (the output was human-verified and corrected). To mimic different levels of video quality we present quantitative results for two different levels of synthetic degradation, which we refer to as ``low'' and ``high'' degradation. \ref{Appendix: Degradation Process for Experiments} contains precise descriptions of how the videos were degraded, along with an example frame with the two different levels of degradation. In addition to these two experiments, we tested our method on several genuine low-quality videos (not synthetically degraded). Ground-truth is not known for these videos, but we overlay output trajectories so the tracking quality can be inspected visually (all trackers are initialized on a common set of features in frame $0$). We refer to this as our qualitative experiment, and the results are included in the supplementary material. The qualitative experiment also includes some higher-quality videos to show that our method does not introduce problems in that case. Table \ref{table:Video Rotation Summary} summarizes the maximum and typical rotation rates experienced in each video sequence in both sets of test videos. Figures \ref{fig:Rotation Rates in Quantitative Test Videos} and \ref{fig:Rotation Rates in Qualitative Test Videos} show complete rotation rate profiles (rotation rates as functions of time) for each video in the two test sets. Figure \ref{fig:Example Output} shows some characteristic tracker results on videos from our quantitative high-degradation experiment. Sample output frames for several videos from our qualitative experiment are presented in Figures \ref{fig:Real Video Output 1}, \ref{fig:Real Video Output 2}, \ref{fig:Real Video Output 3}, and \ref{fig:Real Video Output 4}.

Our test set for our quantitative experiments consists of $8$ video clips, ranging in length from $132$ to $289$ frames. For generating ground truth new features are detected throughout each sequence, and features are terminated when they leave the field of view or when they could no longer be localized by hand. In total, there are $50,754$ feature-frames in our set of test videos. This is the sum of the lifespan (in frames) of each feature in each video. We have both indoor and outdoor sequences in our collection of test videos. It should be noted that the test videos are biased in the sense that they focus on situations which are difficult for un-aided feature trackers. In particular, all of the test videos are degraded, which makes features less distinctive and corner-like than ideal features in high-quality video. We focus on these situations because this is where there is room for improvement over conventional feature tracking, but one must remember the context of these experiments. There is little or no benefit to using our method when tracking ideal, corner-like features; existing methods like KLT can handle such features perfectly well. In our test videos there is significant flow due to camera translation through the environment, and in some instances due to non-rigidity of the subject matter (i.e. effects other than camera rotation). This is important to show that the proposed method  can resolve conflicts between predicted and observed flow, and that it is actually using both the gyros and the imagery for tracking, as opposed to just predicting feature locations in each frame using the gyros.

\begin{table}[H]
\captionsetup{margin=10pt,font=small,labelfont=bf}
\caption{Rotation rate summary for quantitative and qualitative test videos.}
\centering
\tabcolsep=0.11cm
\begin{tabular}{c c | c | c | c | c |}
\cline{3-6}
& & \multirow{2}{*}{Video} & Max Rotation & Mean Rotation & Median Rotation \\ 
& & & Rate (deg/s) & Rate (deg/s) & Rate (deg/s) \\
\hline
\multicolumn{1}{|c}{\multirow{8}{*}{\begin{sideways}\textbf{Quantitative}\end{sideways}}} & \multicolumn{1}{c|}{\multirow{8}{*}{\begin{sideways}\textbf{Tests}\end{sideways}}}
                        & Video 1 & 46.554 & 21.926 & 21.812 \\
\multicolumn{1}{|c}{} & & Video 2 & 56.925 & 22.983 & 21.776 \\
\multicolumn{1}{|c}{} & & Video 3 & 47.702 & 23.849 & 21.537 \\
\multicolumn{1}{|c}{} & & Video 4 & 52.408 & 26.602 & 26.455 \\
\multicolumn{1}{|c}{} & & Video 5 & 46.347 & 19.893 & 18.956 \\
\multicolumn{1}{|c}{} & & Video 6 & 44.285 & 21.538 & 21.532 \\
\multicolumn{1}{|c}{} & & Video 7 & 18.397 & 9.4527 & 8.9123 \\
\multicolumn{1}{|c}{} & & Video 8 & 0.79148 & 0.36397 & 0.35447 \\ \hline
\multicolumn{1}{|c}{\multirow{8}{*}{\begin{sideways}\textbf{Qualitative}\end{sideways}}} & \multicolumn{1}{c|}{\multirow{8}{*}{\begin{sideways}\textbf{Tests}\end{sideways}}}
                        & Low-Quality 1 & 107.462 & 36.169 & 36.641 \\
\multicolumn{1}{|c}{} & & Low-Quality 2 & 35.282 & 17.644 & 17.598 \\
\multicolumn{1}{|c}{} & & Low-Quality 3 & 33.25 & 15.539 & 15.864 \\
\multicolumn{1}{|c}{} & & Low-Quality 4 & 36.676 & 13.49 & 14.168 \\
\multicolumn{1}{|c}{} & & High-Quality 1 & 30.804 & 8.2969 & 9.2238 \\
\multicolumn{1}{|c}{} & & High-Quality 2 & 14.214 & 7.8396 & 7.9647 \\
\multicolumn{1}{|c}{} & & High-Quality 3 & 10.486 & 3.5625 & 3.8939 \\
\hline
\end{tabular}
\label{table:Video Rotation Summary}
\end{table}

\pagebreak

\pagebreak

\pagebreak

\begin{figure}[H]
\centering
\subfloat[KLT + gyro init.]{
	\includegraphics[width=0.49\linewidth, clip=true, trim=0mm 0mm 0mm 0mm]{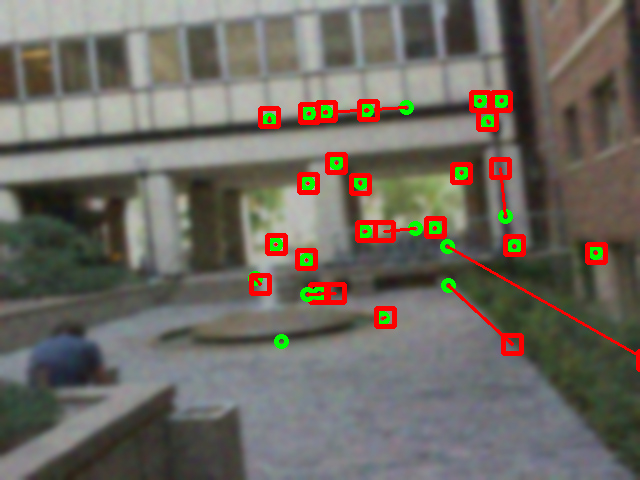}
}
\subfloat[Our method]{
	\includegraphics[width=0.49\linewidth, clip=true, trim=0mm 0mm 0mm 0mm]{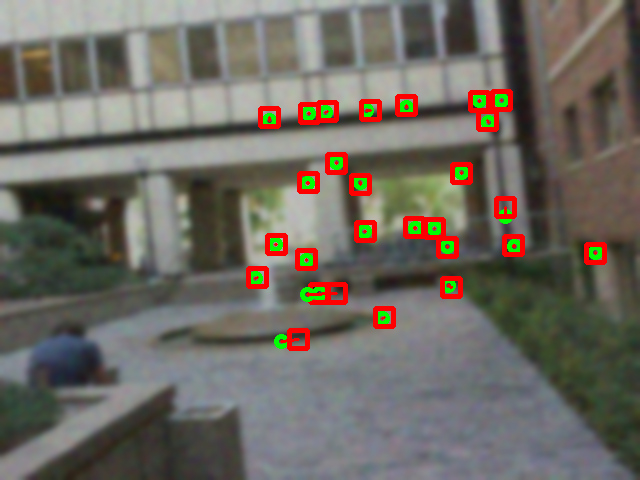}
}\\
\subfloat[KLT + gyro init.]{
	\includegraphics[width=0.49\linewidth, clip=true, trim=0mm 0mm 0mm 0mm]{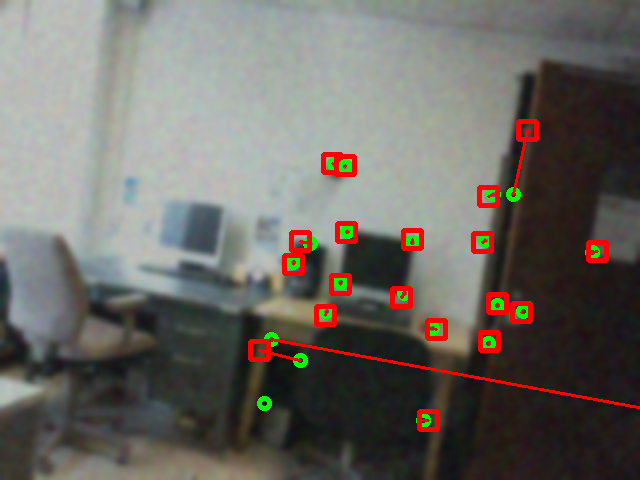}
}
\subfloat[Our method]{
	\includegraphics[width=0.49\linewidth, clip=true, trim=0mm 0mm 0mm 0mm]{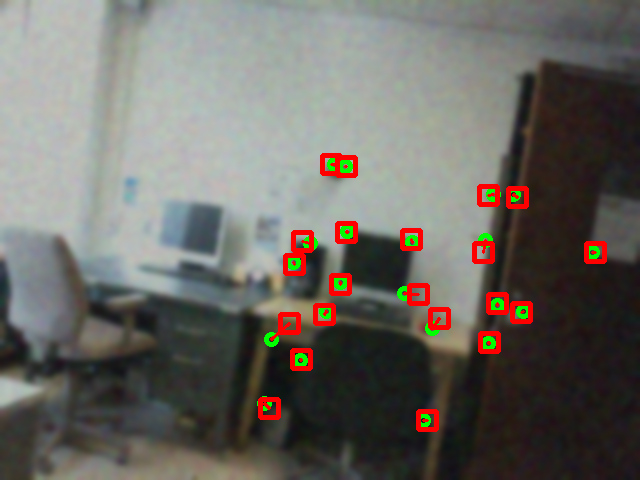}
}\\
\subfloat[KLT + gyro init.]{
	\includegraphics[width=0.49\linewidth, clip=true, trim=0mm 0mm 0mm 0mm]{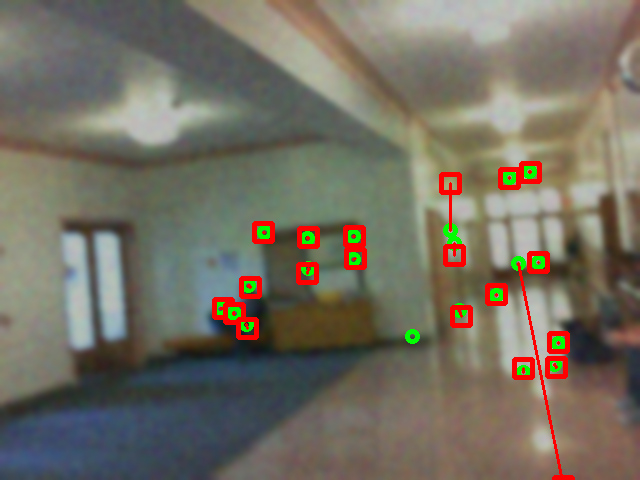}
}
\subfloat[Our method]{
	\includegraphics[width=0.49\linewidth, clip=true, trim=0mm 0mm 0mm 0mm]{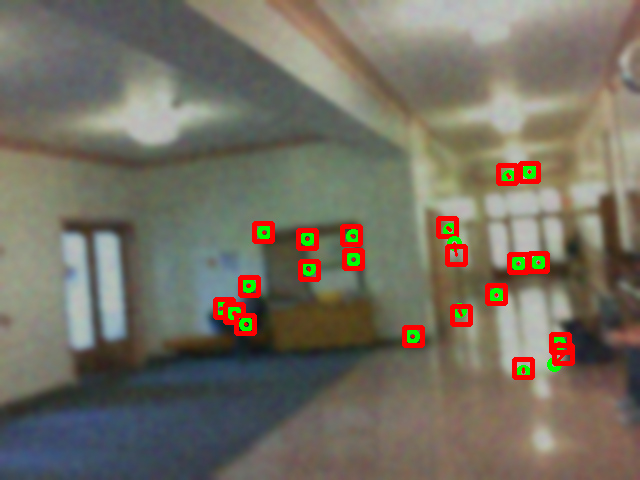}
}
\end{figure}

\begin{figure}[H]
\centering
\subfloat[KLT + gyro init.]{
	\includegraphics[width=0.49\linewidth, clip=true, trim=0mm 0mm 0mm 0mm]{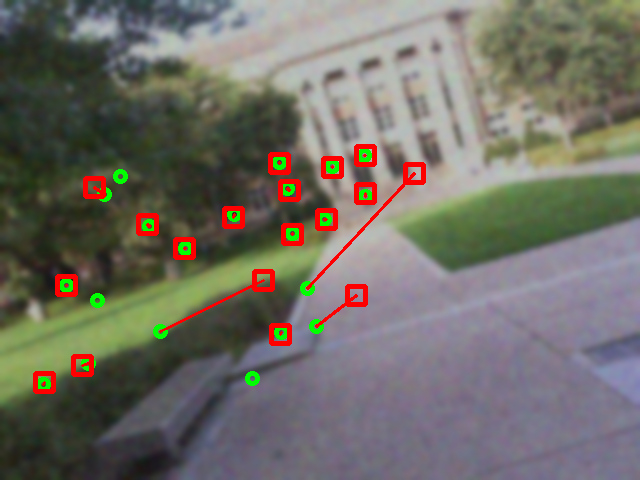}
}
\subfloat[Our method]{
	\includegraphics[width=0.49\linewidth, clip=true, trim=0mm 0mm 0mm 0mm]{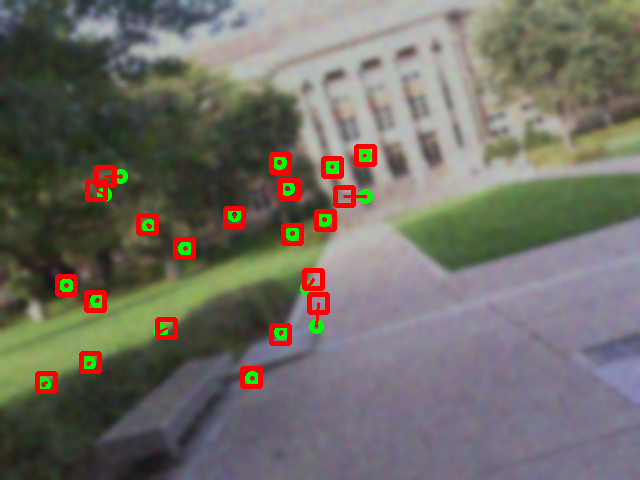}
}\\
\subfloat[KLT + gyro init.]{
	\includegraphics[width=0.49\linewidth, clip=true, trim=0mm 0mm 0mm 0mm]{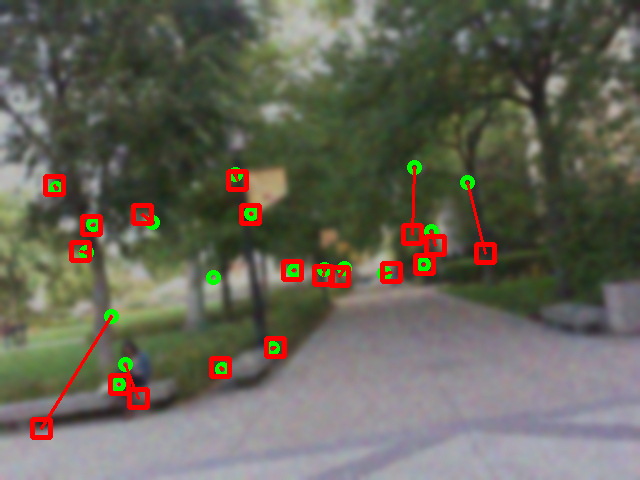}
}
\subfloat[Our method]{
	\includegraphics[width=0.49\linewidth, clip=true, trim=0mm 0mm 0mm 0mm]{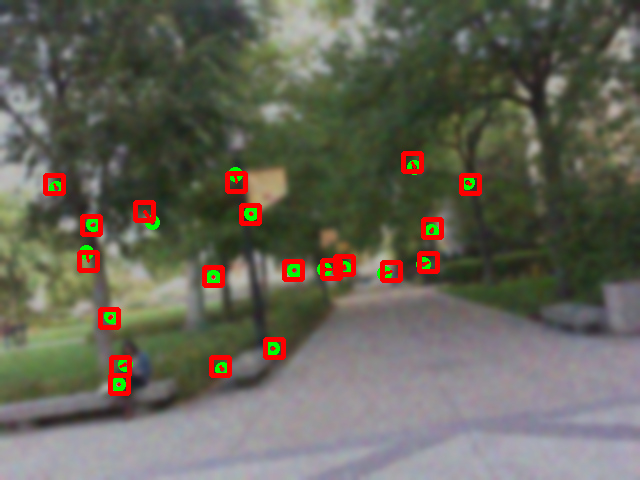}
}\\
\subfloat[KLT + gyro init.]{
	\includegraphics[width=0.49\linewidth, clip=true, trim=0mm 0mm 0mm 0mm]{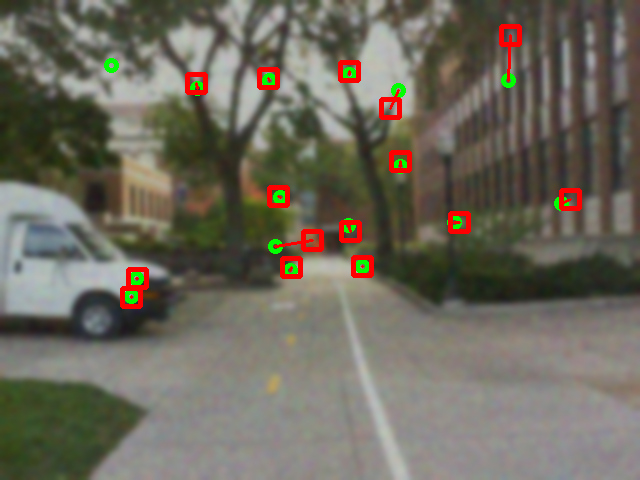}
}
\subfloat[Our method]{
	\includegraphics[width=0.49\linewidth, clip=true, trim=0mm 0mm 0mm 0mm]{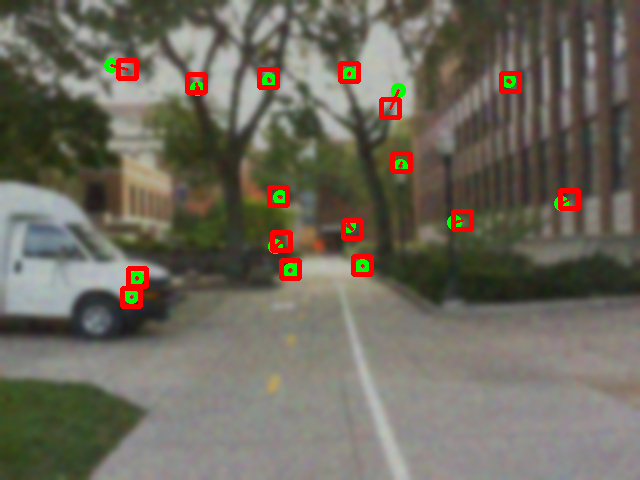}
}
\end{figure}

To evaluate a given tracker on a test video in our quantitative experiment, we initialize the set of tracked features using ground-truth feature positions. When the tracker ``loses'' a feature (which we define as wandering by at least $10$ pixels from ground truth), the feature is re-initialized using its current ground truth position. The mean number of frames between re-initializations (a.k.a.~mean track length) is used as our performance measure. This is very similar to the performance measure used in \cite{Buchanan2007,PolingLermanSzlam2014} to evaluate rank-constrained feature trackers. An alternative performance measure would be to initialize all trackers on a common set of features, let them run un-aided for some number of frames and then compute mean drift or the average $L^1$ or $L^2$ deviation of the output trajectories from ground-truth. Unfortunately, these simple measures are rather unstable because once a feature is lost trackers can behave very unpredictably. Some will drift around slowly while others may quickly wander off screen or snap to the closest corner-like object. While these differences have a large effect on trajectory error, they are not practically very important. What is of interest is how long a feature tracker can hold features and how well it tracks them during that period. Thus, we select mean track length for our performance measure.

In this experiment we compared a pyramidal KLT tracker (OpenCV 2.4.6 implementation), the same OpenCV KLT tracker but with initial flow estimates computed using the gyroscope, which we refer to as ``KLT + Gyro init'' (similar to \cite{hwangbo2011gyro}), a standard gradient-descent single-feature tracker (``$1^\text{st}$-Order Descent''), a gradient descent tracker with gyro initialization (``$1^\text{st}$-Order Descent + Gyro init''), our proposed method with the single-feature tracking implementation (``$1^\text{st}$-Order Descent + Gyro Prior''), a modern rank-penalized multi-feature tracker from \cite{PolingLermanSzlam2014}, which we refer to as ``Multi-Tracker + Rank'' (which uses ``empirical dimension'' for rank estimation and a centered trackpoint matrix for generating the rank penalty), our proposed method with the multi-feature implementation (``Multi-Tracker + Gyro Prior''), and finally, our multi-feature implementation integrated with the rank penalty of \cite{PolingLermanSzlam2014} (``Multi-Tracker + Gyro Prior + Rank''); the rank penalty here is also based on ``empirical dimension'' and a centered trackpoint matrix. All trackers were implemented pyramidally with $4$ resolution levels. For those trackers that do not use the gyro for initialization (``KLT'', ``$1^\text{st}$-Order Descent'', and ``Multi-Tracker + Rank''), features are initialized using ``average flow initialization''. With this scheme, a given frame is registered against the previous at $1/4$'th resolution to get the displacement, $\ba$, of the new frame. Then, each feature is initialized in the new frame with position $\bx$ given by:
\begin{equation}
\label{eqn:Average flow initialization}
\bx = \bx_{\text{prev}} + \ba,
\end{equation}
where $\bx_{\text{prev}}$ is the position of the same feature in the previous frame.

Some of the trackers in this comparison have tuning parameters. In our method $\lambda$ controls the relative strength of the gyro prior term, and in the rank-penalized multi-feature tracker there is a similar coefficient to control the strength of the rank penalty. The ``Multi-Tracker + Gyro Prior'' has both of these parameters. For learning these parameters we collected an additional $4$ videos using our data collection system and generated ground-truth for them. We made a set of training videos by including these $4$ videos as well as $2$ degraded copies of each (``low'' and ``high'' degradation). The parameters were learned by exhaustively searching for the values which gave the best average performance across all training videos. The learned parameters for each tracker were used in both the low-degradation and high-degradation experiments. No tracker had access to any of the $8$ test videos during training. See \ref{Appendix: Degradation Process for Experiments} for the learned parameter values for each tracker.

\begin{figure}[h!]
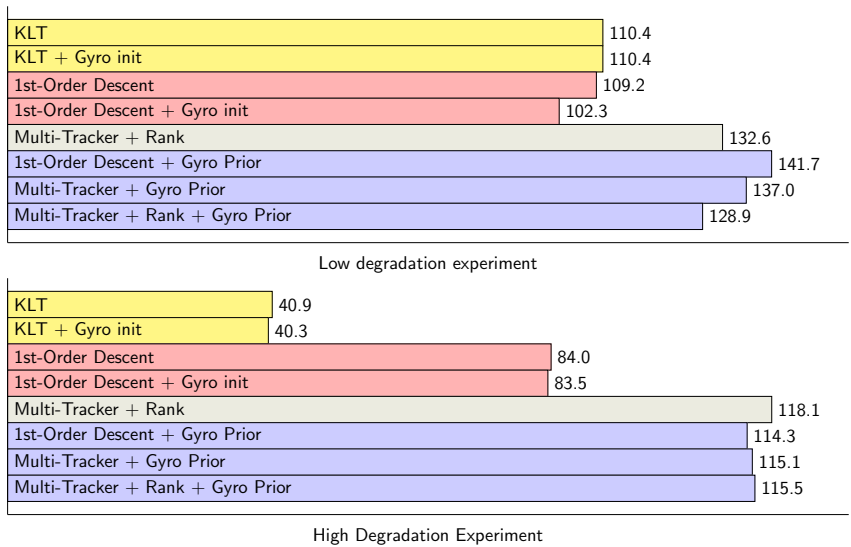

\captionsetup{margin=10pt,labelfont=bf}
\centering
\begin{adjustbox}{max width={.95\linewidth}}
{\centering
\begin{bchart}[step=2,max=156.0,plain,width=16cm]
\bcbar[text=KLT,color=yellow!60]{110.4}
\bcbar[text=KLT + Gyro init,color=yellow!60]{110.4}
\bcbar[text=1st-Order Descent,color=red!30]{109.2}
\bcbar[text=1st-Order Descent + Gyro init,color=red!30]{102.3}
\bcbar[text=Multi-Tracker + Rank,color=blue!40!yellow!20]{132.6}
\bcbar[text=1st-Order Descent + Gyro Prior,color=blue!20]{141.7}
\bcbar[text=Multi-Tracker + Gyro Prior,color=blue!20!]{137.0}
\bcbar[text=Multi-Tracker + Rank + Gyro Prior,color=blue!20!]{128.9}
\bcxlabel{Low degradation experiment}
\end{bchart}
}
\end{adjustbox}\\
\begin{adjustbox}{max width={.95\linewidth}}
{\centering
\begin{bchart}[step=2,max=130.0,plain,width=16cm]
\bcbar[text=KLT,color=yellow!60]{40.9}
\bcbar[text=KLT + Gyro init,color=yellow!60]{40.3}
\bcbar[text=1st-Order Descent,color=red!30]{84.0}
\bcbar[text=1st-Order Descent + Gyro init,color=red!30]{83.5}
\bcbar[text=Multi-Tracker + Rank,color=blue!40!yellow!20]{118.1}
\bcbar[text=1st-Order Descent + Gyro Prior,color=blue!20]{114.3}
\bcbar[text=Multi-Tracker + Gyro Prior,color=blue!20!]{115.1}
\bcbar[text=Multi-Tracker + Rank + Gyro Prior,color=blue!20!]{115.5}
\bcxlabel{High Degradation Experiment}
\end{bchart}
}
\end{adjustbox}
\caption{\label{fig:Bar Graphs}Mean track length (frames) for low and high degradation experiments. Higher is better.}
\end{figure}

\subsection{Analysis of Results}
\label{sec:Analysis of Results}
In Figure \ref{fig:Bar Graphs} we present average track lengths for each tracker for both the low degradation and high degradation experiments. These results are averaged across all of our test videos. Tables \ref{table:Low Degradation Results} and \ref{table:Medium Degradation Results} show per-video comparisons for each experiment. Table \ref{table:Average Processing Frame Rate} shows the average processing rate (measured in frames per second) of each tracker on our testing computer ($3^\text{rd}$-generation Intel Core i5-based laptop). These processing rates only take into account time spent inside the actual tracking routines and do not include common processing tasks like loading frames from the disk into memory. It should be noted that each method was configured (where possible) for the best tracking performance, and no compromises were made to reduce computational cost. The rank-penalized multi-feature tracker was able to run closer to $15$ frames per second with different settings, although this resulted in slightly degraded performance.

\begin{table*}[h]
\captionsetup{margin=10pt,font=small,labelfont=bf}
\caption{Mean track length (frames) - Low degradation. Methods with integrated gyro prior are highlighted in gray. Higher is better. Winning entries are bold.}
\centering
\scriptsize{
\tabcolsep=0.06cm
\begin{tabular}{l l | c c c c c c c c | c |}
\cline{3-11}
& & \multicolumn{8}{|c|}{\textbf{Video Number}} & \multicolumn{1}{|c|}{Average} \\
\cline{3-10}
& & \multicolumn{1}{|c|}{1} & \multicolumn{1}{|c|}{2} & \multicolumn{1}{|c|}{3} & \multicolumn{1}{|c|}{4} & \multicolumn{1}{|c|}{5} & \multicolumn{1}{|c|}{6} & \multicolumn{1}{|c|}{7} & \multicolumn{1}{|c|}{8} & \\
\hline
\multicolumn{1}{|c|}{\multirow{8}{*}{\begin{sideways}\textbf{Tracker}\end{sideways}}}
                       & \multicolumn{1}{|l|}{                           KLT}                                             &                                    81      &                                    128     &                                    19      &                            \textbf{113   } &                            \textbf{290   } &                                    93      &                                    150     &                                    9       &                                    110     \\ \cline{2-11}
\multicolumn{1}{|c|}{} & \multicolumn{1}{|l|}{                           KLT + Gyro init}                                 &                                    79      &                                    126     &                                    19      &                            \textbf{113   } &                            \textbf{290   } &                                    92      &                                    156     &                                    9       &                                    110     \\ \cline{2-11}
\multicolumn{1}{|c|}{} & \multicolumn{1}{|l|}{                           1st-Order Descent}                               &                                    157     &                                    103     &                                    100     &                                    47      &                                    124     &                                    176     &                                    156     &                                    12      &                                    109     \\ \cline{2-11}
\multicolumn{1}{|c|}{} & \multicolumn{1}{|l|}{                           1st-Order Descent + Gyro init}                   &                                    154     &                                    103     &                                    75      &                                    48      &                                    108     &                                    169     &                                    150     &                                    12      &                                    102     \\ \cline{2-11}
\multicolumn{1}{|c|}{} & \multicolumn{1}{|l|}{                           Multi-Tracker + Rank}                            &                                    188     &                                    103     &                            \textbf{134   } &                                    64      &                                    193     &                            \textbf{223   } &                                    133     &                                    23      &                                    133     \\ \cline{2-11}
\multicolumn{1}{|c|}{} & \multicolumn{1}{|l|}{\cellcolor[rgb]{0.9,.9,.9} 1st-Order Descent + Gyro Prior}                  & \cellcolor[rgb]{0.9,.9,.9} \textbf{214   } & \cellcolor[rgb]{0.9,.9,.9} \textbf{165   } & \cellcolor[rgb]{0.9,.9,.9}         82      & \cellcolor[rgb]{0.9,.9,.9}         73      & \cellcolor[rgb]{0.9,.9,.9}         130     & \cellcolor[rgb]{0.9,.9,.9}         166     & \cellcolor[rgb]{0.9,.9,.9} \textbf{282   } & \cellcolor[rgb]{0.9,.9,.9}         20      & \cellcolor[rgb]{0.9,.9,.9} \textbf{142   } \\ \cline{2-11}
\multicolumn{1}{|c|}{} & \multicolumn{1}{|l|}{\cellcolor[rgb]{0.9,.9,.9} Multi-Tracker + Gyro Prior}                      & \cellcolor[rgb]{0.9,.9,.9}         179     & \cellcolor[rgb]{0.9,.9,.9}         123     & \cellcolor[rgb]{0.9,.9,.9}         130     & \cellcolor[rgb]{0.9,.9,.9}         65      & \cellcolor[rgb]{0.9,.9,.9}         198     & \cellcolor[rgb]{0.9,.9,.9}         208     & \cellcolor[rgb]{0.9,.9,.9}         167     & \cellcolor[rgb]{0.9,.9,.9} \textbf{26    } & \cellcolor[rgb]{0.9,.9,.9}         137     \\ \cline{2-11}
\multicolumn{1}{|c|}{} & \multicolumn{1}{|l|}{\cellcolor[rgb]{0.9,.9,.9} Multi-Tracker + Rank + Gyro Prior}               & \cellcolor[rgb]{0.9,.9,.9}         175     & \cellcolor[rgb]{0.9,.9,.9}         104     & \cellcolor[rgb]{0.9,.9,.9} \textbf{134   } & \cellcolor[rgb]{0.9,.9,.9}         58      & \cellcolor[rgb]{0.9,.9,.9}         184     & \cellcolor[rgb]{0.9,.9,.9} \textbf{223   } & \cellcolor[rgb]{0.9,.9,.9}         129     & \cellcolor[rgb]{0.9,.9,.9}         24      & \cellcolor[rgb]{0.9,.9,.9}         129     \\ \hline
\end{tabular}
}
\label{table:Low Degradation Results}
\end{table*}
\begin{table*}[h]
\captionsetup{margin=10pt,font=small,labelfont=bf}
\caption{Mean track length (frames) - High degradation. Methods with integrated gyro prior are highlighted in gray. Higher is better. Winning entries are bold.}
\centering
\scriptsize{
\tabcolsep=0.06cm
\begin{tabular}{l l | c c c c c c c c | c |}
\cline{3-11}
& & \multicolumn{8}{|c|}{\textbf{Video Number}} & \multicolumn{1}{|c|}{Average} \\
\cline{3-10}
& & \multicolumn{1}{|c|}{1} & \multicolumn{1}{|c|}{2} & \multicolumn{1}{|c|}{3} & \multicolumn{1}{|c|}{4} & \multicolumn{1}{|c|}{5} & \multicolumn{1}{|c|}{6} & \multicolumn{1}{|c|}{7} & \multicolumn{1}{|c|}{8} & \\
\hline
\multicolumn{1}{|c|}{\multirow{8}{*}{\begin{sideways}\textbf{Tracker}\end{sideways}}}
                       & \multicolumn{1}{|l|}{                           KLT}                                             &                                    52      &                                    68      &                                    17      &                                    29      &                                    52      &                                    60      &                                    45      &                                    4       &                                    41      \\ \cline{2-11}
\multicolumn{1}{|c|}{} & \multicolumn{1}{|l|}{                           KLT + Gyro init}                                 &                                    52      &                                    61      &                                    17      &                                    29      &                                    54      &                                    62      &                                    44      &                                    4       &                                    40      \\ \cline{2-11}
\multicolumn{1}{|c|}{} & \multicolumn{1}{|l|}{                           1st-Order Descent}                               &                                    135     &                                    81      &                                    75      &                                    40      &                                    81      &                                    125     &                                    125     &                                    9       &                                    84      \\ \cline{2-11}
\multicolumn{1}{|c|}{} & \multicolumn{1}{|l|}{                           1st-Order Descent + Gyro init}                   &                                    135     &                                    79      &                                    66      &                                    39      &                                    79      &                                    125     &                                    137     &                                    9       &                                    84      \\ \cline{2-11}
\multicolumn{1}{|c|}{} & \multicolumn{1}{|l|}{                           Multi-Tracker + Rank}                            &                                    183     &                                    101     &                                    104     &                            \textbf{56    } &                            \textbf{160   } &                            \textbf{208   } &                                    110     &                                    22      &                            \textbf{118   } \\ \cline{2-11}
\multicolumn{1}{|c|}{} & \multicolumn{1}{|l|}{\cellcolor[rgb]{0.9,.9,.9} 1st-Order Descent + Gyro Prior}                  & \cellcolor[rgb]{0.9,.9,.9} \textbf{208   } & \cellcolor[rgb]{0.9,.9,.9} \textbf{130   } & \cellcolor[rgb]{0.9,.9,.9}         74      & \cellcolor[rgb]{0.9,.9,.9}         53      & \cellcolor[rgb]{0.9,.9,.9}         88      & \cellcolor[rgb]{0.9,.9,.9}         120     & \cellcolor[rgb]{0.9,.9,.9} \textbf{226   } & \cellcolor[rgb]{0.9,.9,.9}         15      & \cellcolor[rgb]{0.9,.9,.9}         114     \\ \cline{2-11}
\multicolumn{1}{|c|}{} & \multicolumn{1}{|l|}{\cellcolor[rgb]{0.9,.9,.9} Multi-Tracker + Gyro Prior}                      & \cellcolor[rgb]{0.9,.9,.9}         171     & \cellcolor[rgb]{0.9,.9,.9}         103     & \cellcolor[rgb]{0.9,.9,.9}         110     & \cellcolor[rgb]{0.9,.9,.9} \textbf{56    } & \cellcolor[rgb]{0.9,.9,.9}         137     & \cellcolor[rgb]{0.9,.9,.9}         176     & \cellcolor[rgb]{0.9,.9,.9}         146     & \cellcolor[rgb]{0.9,.9,.9} \textbf{23    } & \cellcolor[rgb]{0.9,.9,.9}         115     \\ \hline
\multicolumn{1}{|c|}{} & \multicolumn{1}{|l|}{\cellcolor[rgb]{0.9,.9,.9} Multi-Tracker + Rank + Gyro Prior}               & \cellcolor[rgb]{0.9,.9,.9}         171     & \cellcolor[rgb]{0.9,.9,.9}         109     & \cellcolor[rgb]{0.9,.9,.9} \textbf{119   } & \cellcolor[rgb]{0.9,.9,.9}         52      & \cellcolor[rgb]{0.9,.9,.9}         145     & \cellcolor[rgb]{0.9,.9,.9}         194     & \cellcolor[rgb]{0.9,.9,.9}         113     & \cellcolor[rgb]{0.9,.9,.9}         21      & \cellcolor[rgb]{0.9,.9,.9}         116     \\ \hline
\end{tabular}
}
\label{table:Medium Degradation Results}
\end{table*}
\begin{table*}[htb]
\captionsetup{margin=10pt,font=small,labelfont=bf}
\caption{Average processing frame rate (frames per second). Methods with integrated gyro prior are highlighted in gray. Higher is better.}
\centering
\scriptsize{
\tabcolsep=0.20cm
\begin{tabular}{|l|c|}
\hline
Tracker & FPS \\ \hline \hline
KLT & 101.2\\ \hline
KLT + Gyro init & 100.9\\ \hline
1st-Order Descent & 44.9\\ \hline
1st-Order Descent + Gyro init & 49.8\\ \hline
Multi-Tracker + Rank & 5.1\\ \hline
\cellcolor[rgb]{0.9,.9,.9} 1st-Order Descent + Gyro Prior & \cellcolor[rgb]{0.9,.9,.9} 41.8\\ \hline
\cellcolor[rgb]{0.9,.9,.9} Multi-Tracker + Gyro Prior & \cellcolor[rgb]{0.9,.9,.9} 37.3\\ \hline
\cellcolor[rgb]{0.9,.9,.9} Multi-Tracker + Rank + Gyro Prior & \cellcolor[rgb]{0.9,.9,.9} 4.9\\ \hline
\end{tabular}
}
\label{table:Average Processing Frame Rate}
\end{table*}

The first thing to notice is that we seldom see any advantage from only initializing KLT or the gradient descent tracker using the gyroscope. This is seen in both the low and high-degradation experiments, where the performance of both KLT and 1st-order descent appears independent of the initialization scheme which was used. The only problems that can be fixed by better tracker initialization are convergence problems and it appears that the 4-level pyramidal tracking scheme, combined with careful optical-only initialization, appears to be good enough to ensure minimization of the respective energy functions in our experiments. 

This is contrary to the findings of some other authors, for instance, \cite{hwangbo2011gyro}, where significant improvements in tracking performance were attributed to gyro-based initialization. However, it is suggested in \cite{hwangbo2011gyro} that for their un-aided tracker they simply initialize each feature using its location from the previous frame. This is certainly inferior to estimating ``average flow'' between frames, as we did for the non-gyro-initialized trackers in our experiments. Estimating average flow can at least compensate for the common components of flow due to large camera rotations about axes orthogonal to the optical axis. Our combined results would suggest that gyro-based initialization is indeed superior to naive initialization, but with a little effort (and no additional hardware) one can achieve the same results with careful, optical-only initialization. The consequence of this is that to gain true performance advantages from gyroscopes one must employ some other mechanism for exploiting them, beyond initialization (In \cite{hwangbo2011gyro} they also pre-warp template images, which is another exploitation strategy which we do not explore in this work).

Another observation is that both regularization with a gyro-derived prior estimate of flow and low-rank regularization offer measurable performance advantages over un-regularized trackers. In the low-degradation experiment the improvement is somewhat modest, while it is much larger in the high-degradation experiment. This makes sense because when the video is higher quality, features are frequently distinctive enough to enable tracking without needing additional information. In all videos in both sets of experiments, both forms of regularization result in approximately the same or better performance than 1st-order descent (with or without gyro initialization). There are two videos ($\#4$ and $\#5$), however, in the low-degradation experiment where KLT outperforms all other methods (again, regardless of initialization). The regularized trackers still offer better performance than un-regularized 1st-order descent, however, which suggests that KLT's advantage in these videos is due to the Gauss-Newton optimization scheme, rather than its lack of regularization. When the energy function is nice enough, Gauss-Newton optimization can cover large distances and converge in a very small number of iterations compared to first-order methods. This may be the source of KLT's advantage in these instances. In the high-degradation experiment both forms of regularization offer clear advantages to both KLT and 1st-order descent. It is also clear from these experiments that even without regularization 1st-order descent offers greater reliability than the Gauss-Newton scheme used by KLT. This is in line with intuition, since taking relatively large steps based on local information can be risky when the local information is poor-quality. The fast line search used by our first-order methods is a safer (albeit slower) approach. It should also be noted that while both regularization with a gyro-derived prior estimate of flow and low-rank regularization tend to offer better performance than un-regularized trackers, the improvements offered from these two techniques are not complimentary. That is, combining these two techniques does not offer a significant advantage over using just one of them. Thus, it would not be advisable for one to use both techniques together (as in the ``Multi-Tracker + Rank + Gyro Prior'').

Finally, the performance of the rank-penalized multi-feature tracker and the trackers which use gyro-based regularization are very similar in both experiments. The advantage that the gyro-regularized tracker has is speed. As can be seen from Table \ref{table:Average Processing Frame Rate}, the gyro prior term adds very little to the computational expense of a tracker. On the other hand, rank-penalized multi-feature tracking is quite expensive. Even when configured for speed (where the tracker can run at approximately $15$ fps in exchange for slightly degraded performance), the rank-penalized multi-feature tracker is several times slower than the single-feature $1^\text{st}$-order descent tracker with gyro prior.

\section{Future Work}
\label{sec:Future Work}
We see three major avenues for continuing this work. The first is to use the regularization technique that we propose with more modern, faster optimization algorithms. The 1st-order descent with line search used in this work is reliable and we chose to use it so that we could evaluate the fundamental ideas of this work without confusing core issues with the challenges of massaging more complicated nonlinear optimization algorithms. However, there is a gap in speed between the method we present and KLT, and the gap will only be closed by using a more sophisticated optimization algorithm.

The second area for farther work is to characterize the types of features that can be reliably tracked when exploiting a prior estimate of flow. It is well-known how to identify whether a given feature is distinctive enough for KLT to track it (see \cite{Tomasi91detectionand}, for instance). We have shown in this work that when you have a prior estimate of flow and you exploit it by regularizing the tracking energy function, it is possible to track features that are not trackable on their own. It should therefore be possible to relax the requirements that are used in real-world applications to decide when to drop and replace bad features. This is an important aspect of any practical application of this work.

The third avenue for future work is to develop a framework for estimating the sensor calibration on-line. In order to exploit gyroscopes to predict flow, we need to have estimates of certain quantities, including gyro biases, relative sensor latencies, and $\bKt$. In this work we estimate these quantities off-line (see \S\ref{sec:Using Gyroscopes to Predict Optical Flow} and \ref{Appendix: Data Collection Hardware}). This is not fundamentally problematic, except that some of the quantities that are needed can change with time. For instance, sensor latency can change whenever camera settings are changed, and gyro biases can drift slowly with time. It would make the techniques of this paper more accessible if these items could be estimated on-line by, for instance, adjusting the calibration constants to reduce the distances between flow predictions and measured optical flow in a handful of ``nice'' features. This would make it possible to exploit gyro-derived optical flow estimates without needing to worry about many of the practical details of the sensors involved.

\section{Conclusion}
\label{sec:Conclusions}
We presented a deeply integrated method of exploiting low-cost gyroscopes to improve general purpose feature tracking. Beyond initializing the search for features using a gyro-derived optical flow prediction, we built on previous work in the area by also using the sensors to regularize the tracking energy function to directly assist in the tracking of ambiguous and poor-quality features. We demonstrated that our technique offers significant improvements in tracking performance over conventional template-based tracking methods, and is in fact competitive with more complex and computationally expensive state-of-the-art trackers, but at a fraction of the computational cost. Additionally, we showed that the practice of initializing a template-based feature tracker using gyro-predicted optical flow does not outperform careful optical-only initialization. This suggests that a more tightly integrated solution, like the one proposed here, is needed to achieve genuine gains in tracking performance from these inertial sensors.

\section*{Acknowledgements}
This work was supported by NSF awards DMS-09-56072 and DMS-14-18386, the University of Minnesota Doctoral Dissertation Fellowship Program, and the Feinberg Foundation Visiting Faculty Program Fellowship of the Weizmann Institute of Science.


\bibliography{refs_3_07_14}

\appendix
\section{Data Collection Hardware}
\label{Appendix: Data Collection Hardware}
\begin{figure}[h!]
\captionsetup{margin=10pt,labelfont=bf}
\centering
\subfloat[PCB Top]{
	\includegraphics[width=0.31\linewidth, clip=true, trim=0mm 0mm 0mm 0mm]{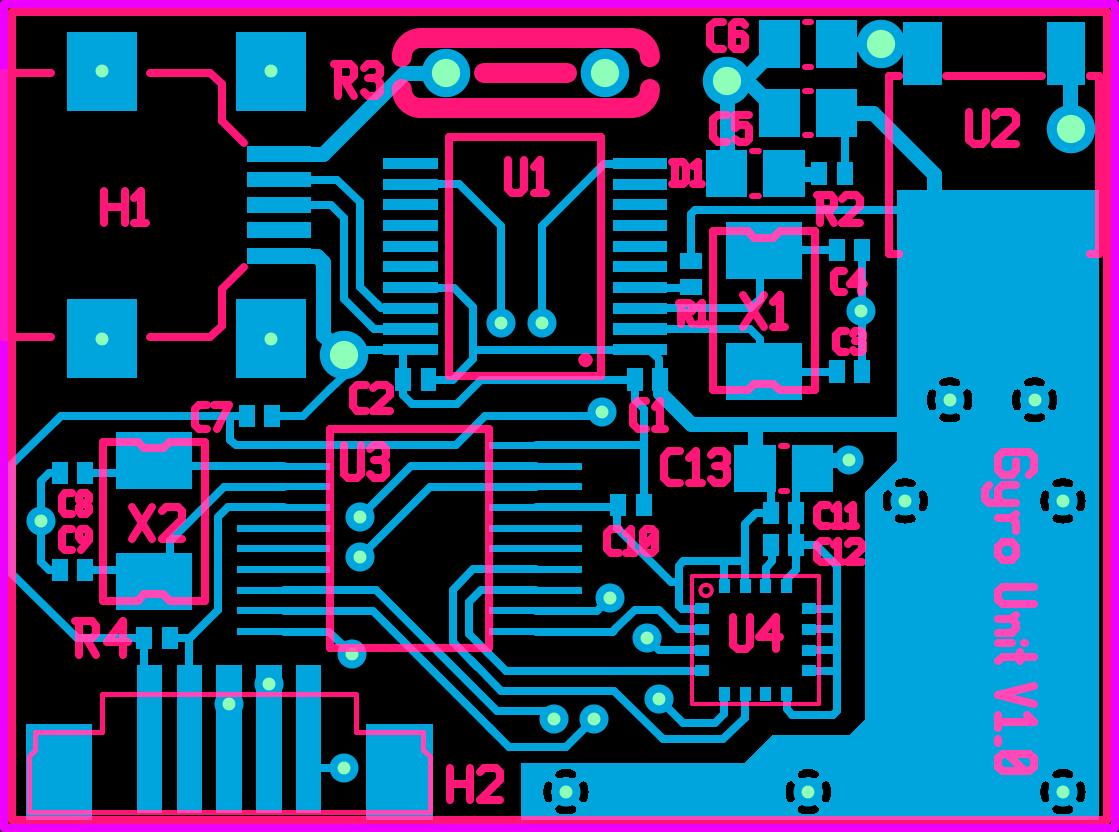}
}
\subfloat[PCB Bottom]{
	\includegraphics[width=0.31\linewidth, clip=true, trim=0mm 0mm 0mm 0mm]{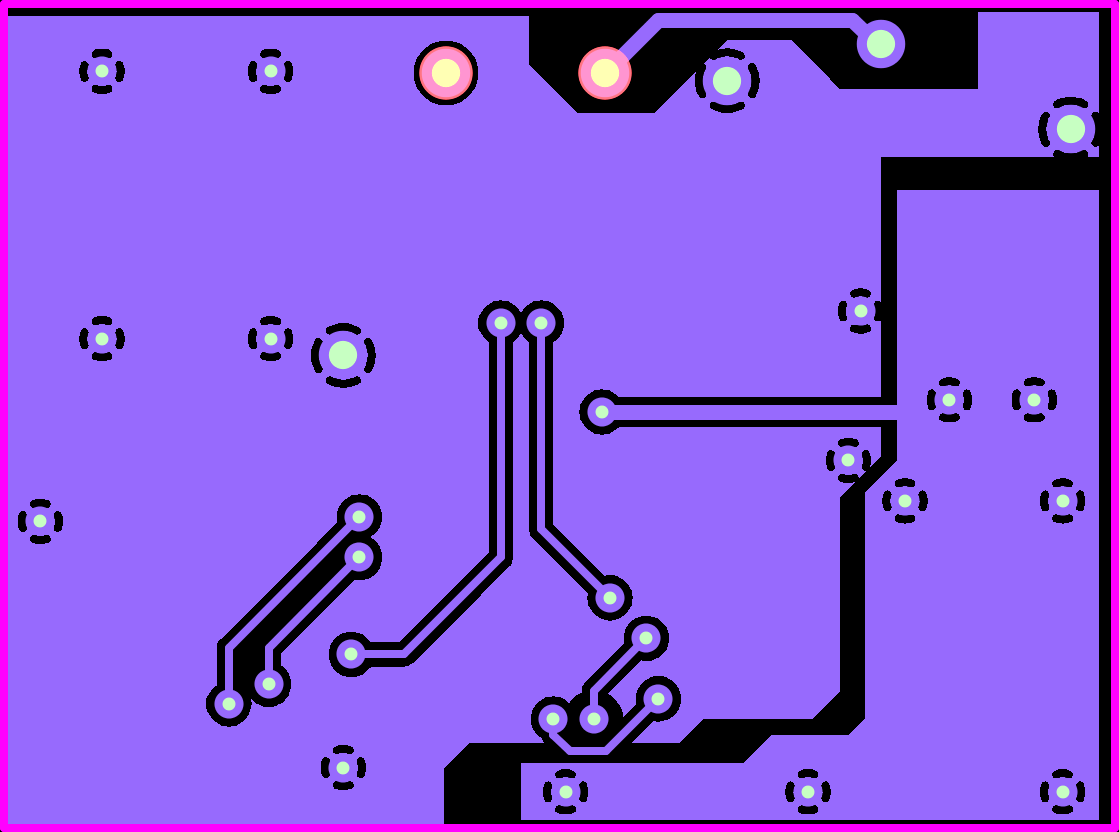}
}
\subfloat[Assembled System]{
	\includegraphics[width=0.31\linewidth, clip=true, trim=0mm 10mm 0mm 70mm]{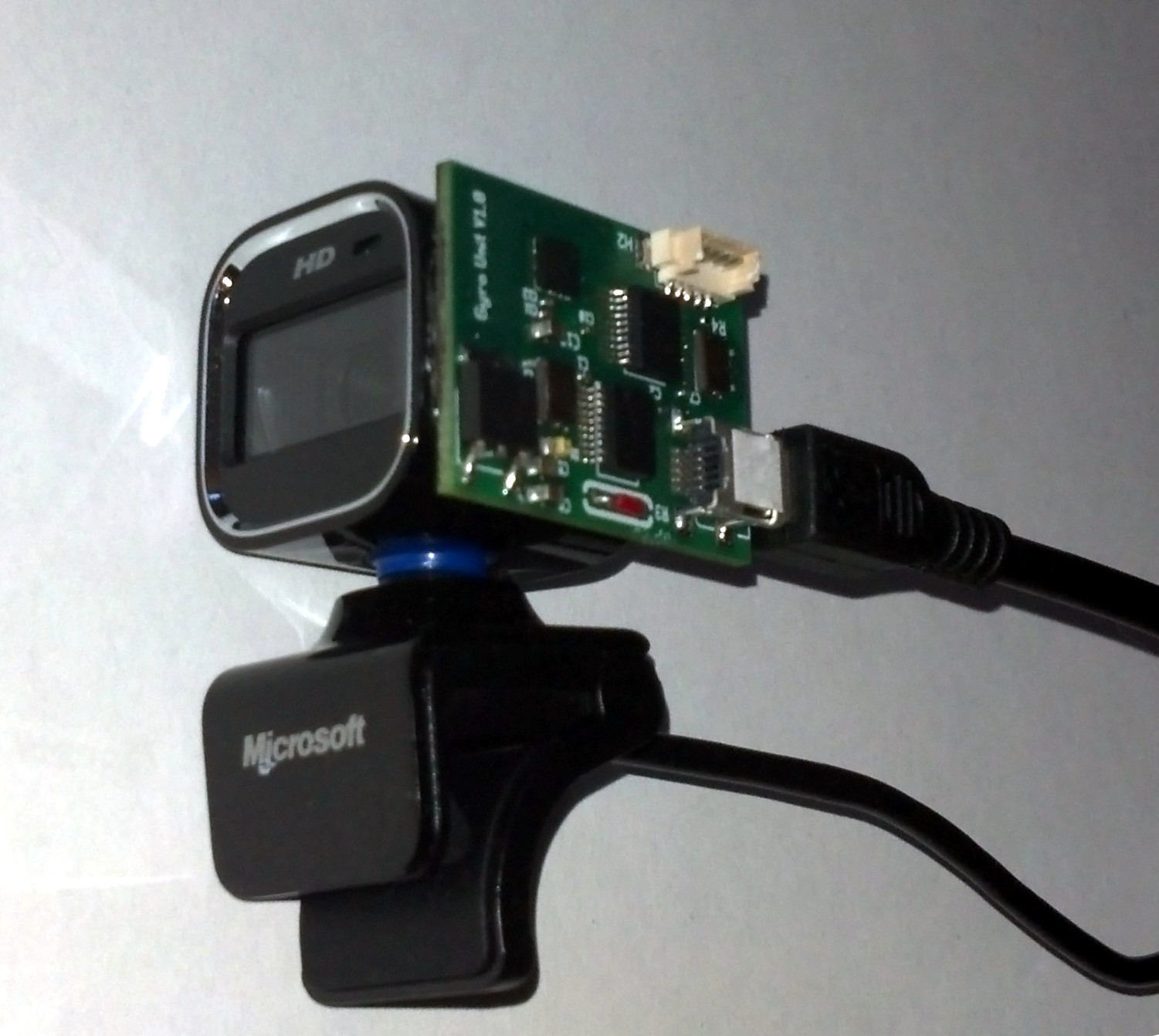}
}
\caption{\label{fig:Hardware Pics}Pictures of the data collection hardware. A small printed circuit board (PCB) with a gyroscope and USB interface was attached to a standard webcam.}
\end{figure}

For our experiments we collected video and gyro data from a custom-built data collection system. This consists of a standard webcam (Microsoft LifeCam HD-6000) and a small custom circuit board with a 3-axis MEMS (Microelectromechanical system) gyroscope. This circuit board was physically attached to the webcam casing with glue to ensure that the camera and the gyro experienced the same rotations. The gyro is an ST L3GD20. It is controlled by a small 8-bit Microchip microcontroller (Pic 18F13K50), which uses a USB-serial adapter chip (Microchip MCP2200) to provide a USB interface. We wrote software for data collection that simultaneously collects imagery from the webcam and gyro data from the L3GD20 and saves the data to a computer running Linux. Images of the gyro circuit and the full data collection system are shown in Figure \ref{fig:Hardware Pics}. All files necessary to re-produce the gyro circuit, along with our data collection software will be available on our supplemental web page.

Gyros suffer from various error sources. Because our proposed method only uses the gyros to propagate attitude for the short periods of time between consecutive frames, we do not need to worry about some of the smaller error sources. However, one source that must be accounted for is sensor bias. This is a constant (or very slowly changing) offset that gets added to each measurement. The biases are generally different on each sensor axis and they can change with temperature, humidity, and sensor age. Fortunately it is easy to measure and compensate for biases. You simply record and average stationary data for a few seconds and then subtract this value from all subsequent measurements. We call this de-biasing. We de-biased our gyros prior to recording each data set. It is also possible to estimate biases ``on-line'' by measuring typical deviation from gyro-based prior flow estimates and final values of flow. This would effectively eliminate the need for collecting stationary data before recording, but we did not attempt that in this work.

The imagery from the webcam and the gyro data from the custom circuit board are synchronized in software on the data collection computer. In order to use the gyro to predict optical flow it is important that the relative latency between the camera and gyro systems be known with high accuracy. An error as small as $0.01$ seconds in relative latency will result in a measurable drop in optical flow estimation accuracy. We have a program for estimating this latency (which will be available on our supplemental web page). One simply rotates the camera back and forth several times and the cross-correlation between the mean optical flow and the gyro rotation rate (as functions of time) is computed. The maximum of the cross-correlation corresponds to the relative latency of the two systems. This process works well and is rather simple. However, we found that the latency of our camera system depends on the exposure setting (which needs to be adjusted for different lighting conditions), and it can also change with CPU load. Additionally, we found the latency of the camera system to be less stable in low-light (high exposure time) settings. Care must be taken to ensure that this value is estimated correctly when recording test data or the gyro-predicted optical flow will be unusable. It is important to note that this is not a deficiency of our proposed method of integrating gyro-derived optical flow with feature tracking, but is instead a limitation of our data collection system. If one used a scientific camera (preferably with a global shutter) with strobe or trigger capability and synchronized the data in hardware, then relative latency would not even need to be estimated.

\section{Degradation Process for Experiments}
\label{Appendix: Degradation Process for Experiments}
The synthetic degradation process for our experiments was a multi-step process. Each frame was degraded separately by first multiplying each pixel by a constant $m$, and then adding per-pixel, Gaussian, i.i.d noise with mean $\mu_1$ and standard deviation $\sigma_1$. Frames were then blurred using a Gaussian mask with standard deviations $\sigma_x$ and $\sigma_y$ in the $x$ and $y$ directions, respectively. Finally, we added additional per-pixel, Gaussian, i.i.d noise with mean $\mu_2$ and standard deviation $\sigma_2$.

\begin{figure}[h!]
\captionsetup{margin=10pt,labelfont=bf}
\centering
\subfloat[Original Frame]{
	\includegraphics[width=0.31\linewidth, clip=true, trim=0mm 0mm 0mm 0mm]{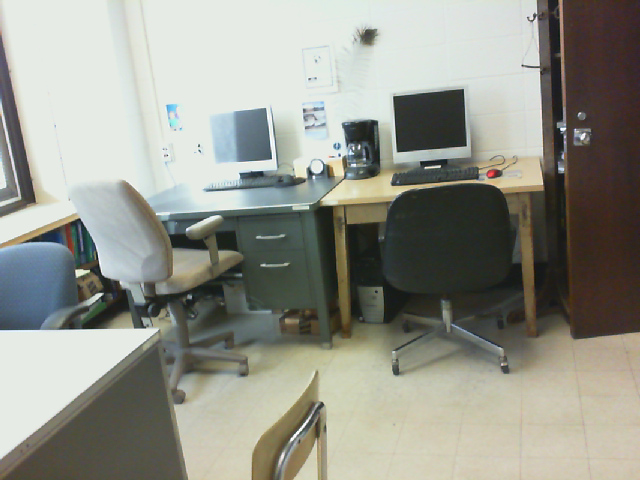}
}
\subfloat[Low Degradation]{
	\includegraphics[width=0.31\linewidth, clip=true, trim=0mm 0mm 0mm 0mm]{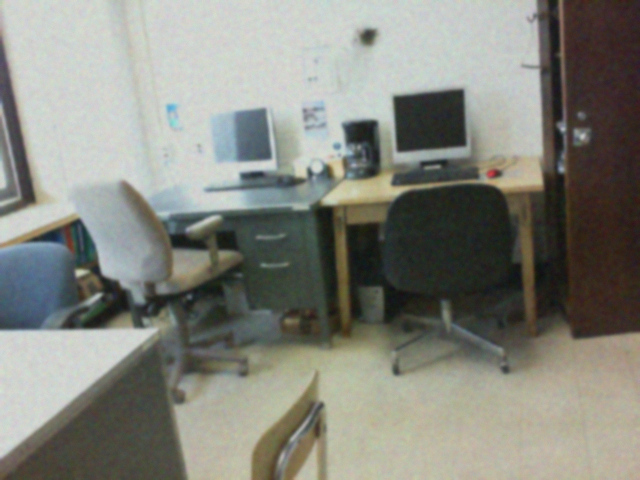}
}
\subfloat[High Degradation]{
	\includegraphics[width=0.31\linewidth, clip=true, trim=0mm 0mm 0mm 0mm]{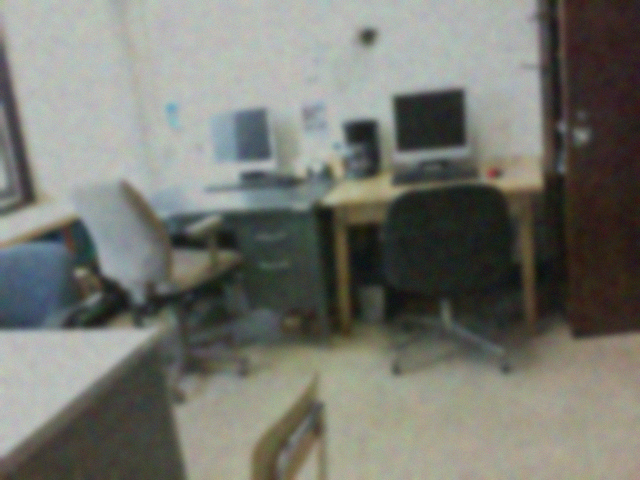}
}
\caption{\label{fig:Degradation Examples}Sample frame and synthetically degraded copies.}
\end{figure}

The degradation adds noise before and after blurring to ensure the noise has strong high and low frequency components. The parameters for the low-degradation experiments were: $m = 0.9$, $\mu_1 = 0.0$, $\sigma_1 = 15.0$, $\sigma_x = 1.5$, $\sigma_y = 1.5$, $\mu_2 = 0.0$, $\sigma_2 = 1.5$. The parameters for the high-degradation experiments were: $m = 0.8$, $\mu_1 = 0.0$, $\sigma_1 = 30.0$, $\sigma_x = 3.0$, $\sigma_y = 3.0$, $\mu_2 = 0.0$, $\sigma_2 = 3.0$. A sample frame is shown in Figure \ref{fig:Degradation Examples} next to degraded copies of the image using both the high and low degradation profiles. Of course, the source code for degrading the videos will be available on our supplemental web page. We note that we originally used lower multipliers for both degradation profiles to farther darken the videos, as done in [19]. However, we found that the performance of the KLT implementation we were using (OpenCV) began to fall off sharply when using multipliers below $0.8$. Since this was synthetic degradation we felt it was unfair to select values that seemed to cause unreasonable harm to one of the trackers.

\subsection{Learned Parameters}
For each tracker with parameters, the parameters were learned via the process described in \S5 of the paper. The learned parameters for each tracker are given in table \ref{table:Learned Parameters}.
\begin{table*}[htb]
\captionsetup{margin=10pt,font=small,labelfont=bf}
\caption{\label{table:Learned Parameters}Learned parameters for each feature tracker (trackers not listed in this table did not have tuning parameters)}
\centering
\tabcolsep=0.5cm
\begin{tabular}{|l|c|}
\hline
Tracker & Learned Parameters \\ \hline \hline
Multi-tracker + Rank & rank coeff. $= 0.45$ \\ \hline
$1^\text{st}$-order descent + Gyro prior & $\lambda = 0.0125$ \\ \hline
Multi-tracker + Gyro prior & $\lambda = 0.005$ \\ \hline
Multi-tracker + Rank + Gyro prior & rank coeff. $= 0.4$, $\lambda = 0.00003$ \\ \hline
\end{tabular}
\end{table*}

\end{document}